\documentclass{article} 
\usepackage{iclr2026_conference,times}
\usepackage{makecell}

\usepackage{amsmath,amsfonts,bm}

\def\eqref#1{equation~\ref{#1}}

\def\1{\bm{1}}

\DeclareMathAlphabet{\mathsfit}{\encodingdefault}{\sfdefault}{m}{sl}
\SetMathAlphabet{\mathsfit}{bold}{\encodingdefault}{\sfdefault}{bx}{n}

\definecolor{denim}{rgb}{0.08, 0.38, 0.74}
\definecolor{darkgreen}{RGB}{0,100,0}
\usepackage{hyperref}
\usepackage{float}
\usepackage{url}
\usepackage[table,xcdraw,dvipsnames]{xcolor}
\usepackage{enumitem}
\definecolor{rowgray}{gray}{0.96}

\hypersetup{
  colorlinks   = true,
  urlcolor     = denim,
  linkcolor    = denim, 
  citecolor    = denim,
}
\usepackage[most]{tcolorbox}
\usepackage{inconsolata}
\usepackage{booktabs}
\usepackage{wrapfig}
\usepackage{xspace}
\usepackage{tabularx}
\newtcolorbox{prompt}[2][]{%
    enhanced,
    left=4mm,
    right=4mm,
    top=2mm,
    bottom=2mm,
    boxsep=0.7mm,
    rounded corners,
    title=#2,
    fontupper=\footnotesize\linespread{0.9}\ttfamily,
    #1
}
\usepackage{multirow}
\newcommand\dataset{\textsc{Menlo}\xspace}

\usepackage{amssymb}

\ifx\XeTeXrevision\undefined
\else
  \usepackage{xunicode}
\fi

\newcommand\greencheckmark{$\color{Green}{\checkmark}$}
\newcommand\redcross{$\color{red}{\pmb\times}$}
\newcommand\redquestion{$\color{red}{?}$}
\newcommand\ok{\tikz[baseline=(char.base)]{
  \node[shape=circle,draw,inner sep=0.3pt] (char) {\tiny OK};
}}
\newcommand{\huggingface}{\raisebox{-1.5pt}{\includegraphics[height=1.05em]{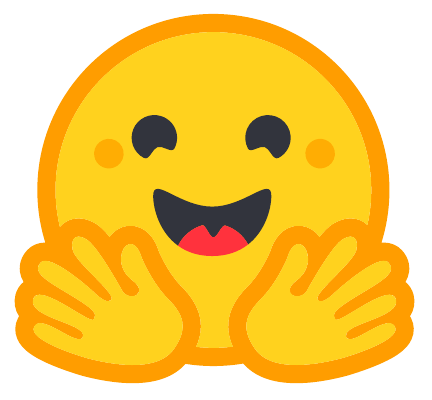}}\xspace}
\newcommand{\menlo}{\raisebox{-1.5pt}{\includegraphics[height=1.4em]{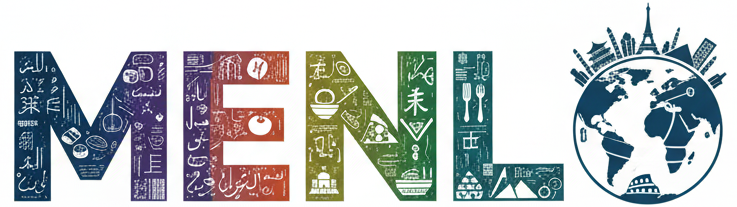}}\xspace}
\title{\menlo: From Preferences to Proficiency --\\Evaluating and Modeling Native-like Quality Across 47 languages}

\author{\textbf{Chenxi Whitehouse\thanks{Equal Contribution.} \qquad
Sebastian Ruder\footnotemark[2] \qquad 
Tony Zhiyang Lin \qquad  
Oksana Kurylo} \\
\textbf{Haruka Takagi \hspace{1em}
Janice Lam \hspace{1em} 
Nicolò Busetto \hspace{1em}
Denise Diaz \hspace{1em} 
Francisco Guzmán\thanks{Handshake AI. Work conducted while at Meta.} 
}
    \\ \\
    Meta Superintelligence Labs  \\
    \texttt{chenxwh@meta.com \quad  ruder@meta.com} 
}

\iclrfinalcopy
\begin{document}

\maketitle

\begin{abstract}

Ensuring native-like quality of large language model (LLM) responses across many languages is challenging. To address this, we introduce \dataset, a framework that operationalizes the evaluation of native-like response quality based on audience design-inspired mechanisms. Using \dataset, we create a dataset of 6,423 human-annotated prompt–response preference pairs covering four quality dimensions with high inter-annotator agreement in 47 language varieties. 
Our evaluation reveals that zero-shot LLM judges benefit significantly from pairwise evaluation and our structured annotation rubrics, yet they still underperform human annotators on our dataset. We demonstrate substantial improvements through fine-tuning with reinforcement learning, reward shaping, and multi-task learning approaches. Additionally, we show that RL-trained judges can serve as generative reward models to enhance LLMs' multilingual proficiency, though discrepancies with human judgment remain. Our findings suggest promising directions for scalable multilingual evaluation and preference alignment. We release our dataset and evaluation framework to support further research in multilingual LLM evaluation.

\begin{center}
         \huggingface \hspace{0.1em} \textbf{Dataset}  \hspace{0.1em} \url{https://huggingface.co/datasets/facebook/menlo}
    \end{center}
\end{abstract}
\section{Introduction}

In order for LLMs to be most useful across the globe, they need to be able to provide high-quality responses in many languages. Responses should be relevant \citep{zhuang2024hydra}, factually accurate \citep{jacovi2025facts}, and natural \citep{marchisio-etal-2024-understanding,guo-etal-2025-large}, among other considerations. Ultimately, for interaction in any language to be seamless, responses need to be indistinguishable from those of a native speaker \citep{novikova-etal-2016-crowd,liu-etal-2021-naturalness}.
Language proficiency in humans has traditionally been evaluated via standardized tests \citep{jamieson2000toefl}. While such tests have been applied to evaluating LLMs \citep{anil2023palm,mayor2024evaluating,lothritz2025testing}, they are difficult to scale and do not readily correspond to real-world conversations. 
What is considered a \textit{native-like} response 
largely depends on speakers' and listeners' interpretations of whom they are speaking to \citep{bell1984language}. 

\begin{figure}
    \centering    \includegraphics[width=0.95\textwidth]{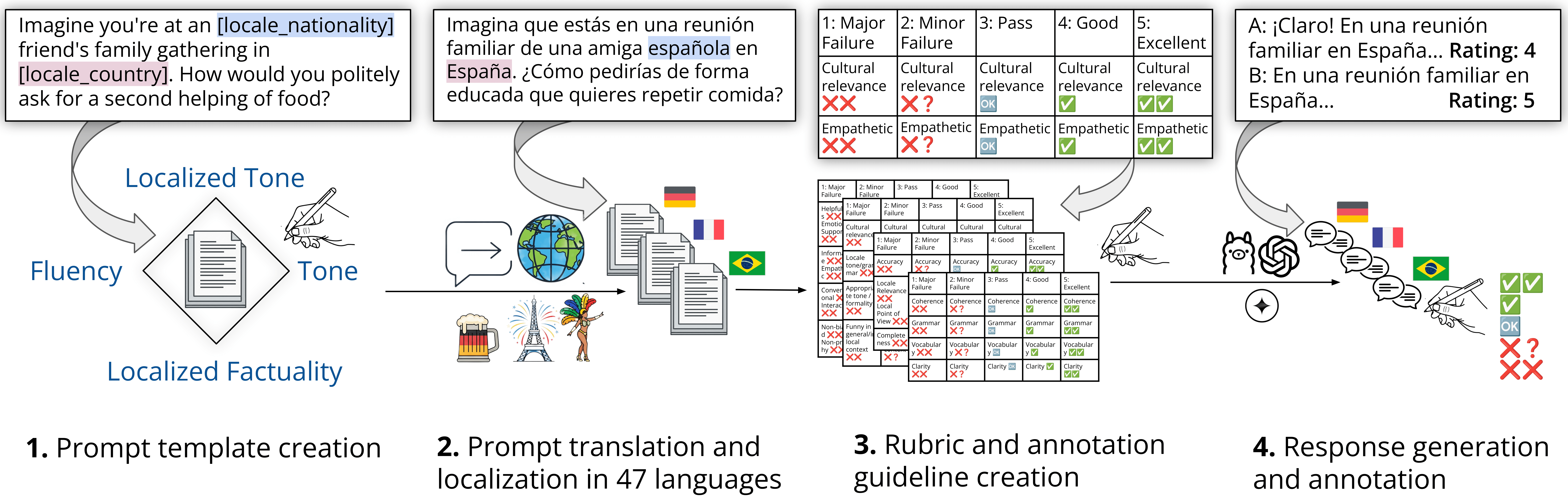}
      \caption{\dataset framework and annotation process. 1) Human-written prompt templates evoking local contexts are created in English for the four dimensions. 2) Prompt templates are translated and localized into 47 language varieties. 3) Annotation guidelines are created that break down each dimension into easy-to-follow rubrics. 4) LLMs are used to generate response pairs for each prompt, which are annotated with Likert-scale ratings and preferences.}
\label{fig:menlo_framework}
\end{figure}

To operationalize the evaluation of native-like response quality across languages, we propose \textbf{M}ultilingual \textbf{E}valuation of \textbf{N}ative-\textbf{L}ike \textbf{O}utput (\dataset); see \autoref{fig:menlo_framework} for an overview. \dataset  breaks down native-like response quality into four key dimensions: i) language quality and coherence; ii) alignment with cultural and linguistic nuances of a specific language variety/locale; iii) factual correctness and grounding in the local context; and iv) overall writing style and helpfulness.
 
Building on mechanisms from audience design \citep{bell1984language}, we propose creating tailored prompts that effectively evoke local contexts by defining the target audience (e.g., an addressee or reference group), thereby guiding the generated language to converge to contextually appropriate ``native'' styles. 
We develop instructions that reduce annotation subjectivity and improve inter-annotator agreement. Responses are generated using state-of-the-art LLMs and annotated with ratings on a 1--5 Likert scale, with an average Krippendorff's $\alpha = 0.84$.
Overall, the \dataset dataset consists of 6,423 annotated prompt-response preference pairs, and 81,014 annotations, in 47 language varieties.

Human evaluation, particularly at a massively multilingual scale is expensive. We thus evaluate the ability of LLMs to serve as judges of native-like quality responses. We find that in zero-shot setting, \textbf{pairwise evaluation}---where models predict scores for two responses simultaneously (without explicitly predicting preference)---significantly outperforms its pointwise counterpart.
The advantage of evaluating two responses side-by-side is even bigger than in-context examples with labels. In addition, we observe significant improvements with judges using our annotation rubrics compared to judges without rubrics, highlighting the generality of the \dataset framework.

As zero-shot judges remain below human annotation quality, using the pairwise evaluation setup, we fine-tune \texttt{Qwen3-4B} and \texttt{Llama4-Scout} as LLM judges on the \dataset training data, finding that RL-trained models outperform their SFT counterparts. In particular, a multi-task \texttt{Llama4-Scout} model trained with shaped rewards surpasses frontier API models with the strongest overall performance across 47 language varieties, reaching agreement levels comparable to human annotators.

Finally, we demonstrate that these judges can be used as generative reward models (RMs) to directly improve a policy model's proficiency. By using our pairwise RL-trained \texttt{Qwen3-4B} judge to post-train the base \texttt{Qwen3-4B} model, we observe quality gains as measured by both LLM evaluators and human raters. However, LLM evaluators tend to be overconfident in assessing improvements compared to human judgments ($+0.6$ higher gain). This finding shows that while judges trained with our framework can successfully drive model improvements, the gap between LLM and human raters highlights the remaining challenges in reliably modeling native-like quality across languages.

Our contributions are the following: \textbf{1)} We develop \dataset, a framework for the evaluation of native-like response quality in four dimensions, based on principles from audience design, employing parametric templates and carefully crafted annotation guidelines. \textbf{2)} We create the \dataset dataset, consisting of 6,423 annotated prompt-response preference pairs in 47 language varieties. \textbf{3)} We evaluate zero-shot judges on the annotated data, demonstrating the benefits of pairwise evaluation and rubrics. \textbf{4)} We show that multi-task RL and reward shaping enables fine-tuning a judge that is on par with human annotators in 47 language varieties. \textbf{5)} We demonstrate that pairwise fine-tuned judges can be used as generative RMs to improve policy model language proficiency, while we find that LLM evaluations tend to overestimate improvements compared to human raters.

Our framework unifies the \dataset dataset, RL-trained pairwise judging, and generative reward modeling, offering a practical and scalable approach to both assess and improve native-like quality.

\section{The \dataset Dataset} 
\label{sec:data}

\begin{figure}[t]
\vspace{-0em}
    \centering    \includegraphics[width=\textwidth]{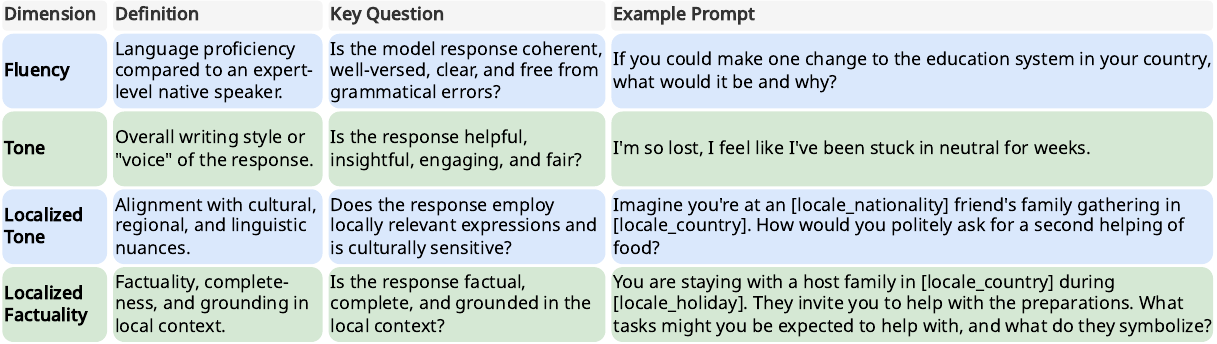}
          \vspace{-1em}
      \caption{Dimensions of native-like response quality in \dataset and example prompt (template).}
\label{fig:menlo_dimensions}
\end{figure}

\dataset characterizes native-like conversational response quality in a language along four key dimensions: fluency, tone, localized tone, and localized factuality. These dimensions go beyond prior work that focused mainly on naturalness \citep{novikova-etal-2016-crowd,liu-etal-2021-naturalness,guo-etal-2025-large} and are motivated by work on language proficiency assessment \citep{ke2019automated}, cross-cultural variation \citep{hershcovich-etal-2022-challenges,myung2024blend}, local knowledge grounding \citep{hupkes2025multiloko}. We provide further context on our definition of these dimensions in \autoref{fig:menlo_dimensions}.

From a sociolinguistic perspective, the Style Axiom \citep{bell1984language} states that intraspeaker variation (style) reflects interspeaker variation (social). Native-like quality is therefore not a single fixed target but a socially conditioned range of stylistic choices that depend on interlocutors. Key mechanisms include accommodation, where speakers adapt their style to the addressee, and referee design, where speakers align with an absent reference group they wish to identify with. These mechanisms motivate our focus on tone and localized tone as central to native-like quality.
To operationalize these ideas, we design human-written parametric English prompt templates for each dimension with placeholders such as \texttt{[locale\_nationality]}, \texttt{[locale\_country]}, \texttt{[locale\_holiday]}, etc. By defining the addressee or reference group, these prompts evoke local contexts and guide models toward contextually appropriate ``native'' styles. We provide an overview of the \dataset framework and annotation process in \autoref{fig:menlo_framework}.

We select 47 language varieties representing a typologically diverse set of widely used languages and their major variants, including, e.g., South American and European varieties of Spanish and Portuguese, several varieties of English, and romanized versions of non-Latin script languages (see Appendix \ref{app:menlo_languages}). Native speakers are recruited to professionally translate these prompt templates, with placeholders instantiated using locally relevant entities. As native quality is tied to the local context, we ensure that native speakers are from the specific regions where the corresponding language varieties are spoken. Similar criteria are used to select annotators for each language variety. Each language variety has approximately the same number of examples in \dataset.

\begin{table}[t]
\caption{\label{tab:annotation_stats}
Annotation and statistics of \dataset across evaluation dimensions. IAA presents \textit{Krippendorff's}~$\alpha$ measuring inter-annotator agreement. Average token counts are computed using \texttt{Qwen3-4B}.}
\vspace{0.5em}
\centering
\small
\resizebox{\textwidth}{!}{
\begin{tabular}{lcccccccc}
\toprule
\multirow{2}{*}{\textbf{Dimension}} & \multirow{2}{*}{\# Annotations} & \multirow{2}{*}{\# Annotators} & \multirow{2}{*}{\# Prompts} & \multicolumn{2}{c}{Avg \# Tokens} & 
\multicolumn{3}{c}{Rating (1--5 Scale)} \\

& & & & Prompt & Response & IAA & Mean & Std. \\
\cmidrule(r){1-1} \cmidrule(lr){2-4} \cmidrule(lr){5-6} \cmidrule(l){7-9}
Fluency              & 23,556 & 450 & 1,820 & 81.6  & 804.3 & 0.82 & 4.01 & 1.11 \\
Tone                 & 18,712 & 429 & 1,410 & 27.8  & 575.3 & 0.86 & 3.48 & 1.35 \\
Localized Tone       & 22,324 & 530 & 1,815 & 71.7  & 559.2 & 0.83 & 3.89 & 1.17 \\
Localized Factuality & 16,422 & 525 & 1,378 & 121.8 & 839.1 & 0.84 & 3.82 & 1.16 \\
\midrule
\textbf{Overall}       & 81,014 & 1,934 & 6,423 &     75.6  & 692.2 & 0.84 & 3.80 & 1.20  \\
\bottomrule
\end{tabular}}
\end{table}

To ensure consistency in evaluation, we develop instructions that reduce the subjectivity of the annotation and break down the four broad dimensions into easy-to-follow rubrics and self-explanatory signals (human-written). Annotators receive guidelines with examples for each dimension. We additionally develop a customized annotation tool and annotator screening tests to filter out unreliable annotators. Furthermore, we train 1--2 expert annotators per language who provide language-specific feedback to annotators and provide gold annotations on a subset of examples.

We generate two responses for each prompt with state-of-the-art LLMs including GPT-4o, Llama4-Maverick, Llama4-Maverick with Search, and Gemini 1.5 with Search. We present both responses in randomized order to human annotators and ask them to provide 1–5 Likert ratings per response, allowing ties. Each response pair is annotated by at least 3 annotators, with final scores aggregated via majority vote. Annotators achieve high reliability, with an average Krippendorff's $\alpha = 0.84$. 

\begin{table}[t]

\centering
\caption{Comparison of multilingual response quality datasets: \textsc{Recon} \citep{doddapaneni-etal-2025-cross}, 
\textsc{Pariksha} \citep{watts-etal-2024-pariksha}, \textsc{BigGen Bench} \citep{kim-etal-2025-biggen}, \textsc{M-RewardBench} \citep{gureja-etal-2025-rewardbench}, \textsc{MM-Eval} \citep{son2024mm}. 
$|\mathcal{L}|$: \# of languages, $|\mathcal{D}|$: \# of prompts, IAA: inter-annotator agreement, IF: instruction following. $^*$: 81,014 annotations; 1,776 test examples.}
\label{tab:dataset_comparison}
\vspace{0.5em}

\renewcommand{\arraystretch}{1.5}
\rowcolors{0}{white}{rowgray}
\resizebox{\textwidth}{!}{
\begin{tabular}{%
    p{0.14\textwidth}  
    p{0.06\textwidth}  
    p{0.08\textwidth}  
    p{0.06\textwidth}  
    p{0.28\textwidth}  
    p{0.22\textwidth}  
    p{0.16\textwidth}  
    p{0.10\textwidth}
}
\toprule
\textbf{Dataset} & 
\textbf{$| \mathcal{L} |$} 
& \phantom{00}\textbf{$| \mathcal{D} |$} & \textbf{IAA} & \textbf{Dimensions} & \textbf{Prompts} & \textbf{Responses} & \textbf{Ratings} \\
\midrule
\dataset & 47 & \phantom{0}6,423$^*$  & 0.84 & Fluency, tone, localized tone, localized factuality & Human-written, translated \& localized &
\makecell[lt]{Annotated in \\ each language} & Preference \& 1--5 \\

\textsc{Pariksha} & 10 & \phantom{00,}200 & 0.54 & Hallucinations, task quality, linguistic acceptability & Human-written &\makecell[tl]{Annotated in \\ each language} & Preference \& 0--2  \\

\textsc{BigGen Bench} & 10 & \phantom{00,}420 & \phantom{0.}-- & Poem, reasoning, humor, translation, historical text & Human-written, LLM-augmented & Generated in each language & 1--5 \\

\textsc{Recon} & \phantom{.}6 & \phantom{0}3,000 & \phantom{0.}-- & IF, theory of mind, reasoning, safety, planning, etc. & Translated & Generated in each language & Preference \& 1--5 \\

\textsc{MM-Eval} & 18 & \phantom{0}4,981 & \phantom{0.}-- & Reasoning, chat, linguistics, hallucination, safety & Translated & Generated in each language & Preference \\

\makecell[lt]{\textsc{M-Reward} \\ \textsc{Bench}} & 23 & 66,787 & \phantom{0.}-- & Chat, safety, reasoning, translation & Translated & Translated & Preference \\

\bottomrule
\end{tabular}}
\end{table}

Overall,  \dataset consists of 6,423 annotated prompt-response preference pairs across 47 language varieties, each containing a prompt, two responses, and corresponding scores, totaling 81,014 human annotations. Summary statistics are reported in \autoref{tab:annotation_stats}. 
Example prompts per dimension are shown in \autoref{fig:menlo_dimensions}. Further details of \dataset including annotation process, language coverage, rubrics, and full examples featuring responses and their corresponding ratings are provided in \autoref{sec:data_detail}.

\autoref{tab:dataset_comparison} compares \dataset with existing multilingual preference datasets. 
\dataset provides localized prompts and responses, spans more languages, and reaches higher agreement than prior work.

\section{Evaluating LLM-Judges on \dataset}
\label{sec:judges}

We next evaluate the ability of LLMs to serve as automatic judges of native-like quality on \dataset. 
Out of the 6,423 pairs, we hold out 1,766 pairs (3,552 responses) as the test set,\footnote{Translations of the same prompt template are assigned the same set to prevent train-test leakage.} and use the remainder for training and prompt development (see \S\ref{sec:training} and \S\ref{sec:rm}). Where expert annotations are available, we use these as labels. For the remaining responses, we average the annotated ratings of each response.\footnote{Multiple annotations can be used in future work on pluralistic alignment \citep{sorensen2024roadmap}.}
Our evaluation focuses on three questions: 
(i) how pointwise and pairwise setups compare,
(ii) the effect of few-shot exemplars, and 
(iii) the role of explicit grading rubrics.  

We benchmark a range of open-source and API-based models, covering both \emph{thinking} and \emph{non-thinking} variants: 
\texttt{Qwen3-4B}, \texttt{Qwen3-32B}, \texttt{Llama-3.1-8B},\footnote{Llama models are instruction-tuned and we omit the \texttt{Instruct} suffix for brevity.}
\texttt{Llama-3.3-70B}, \texttt{Llama4-Scout}, \texttt{o3}, \texttt{gpt-4o}, and \texttt{gpt-4.1}. 
All models are used in the default setup with maximum output length $8192$.

We report two primary metrics:
(i) \textit{Macro-F1} for 5-way classification, and
(ii) \textit{Preference} accuracy over Win/Loss/Tie outcomes. Note that \textit{we do NOT directly ask for preference judgments; rather, we infer these from the assigned grades}. Additionally, we report classification accuracy, Krippendorff's $\alpha$, which measures agreement with human annotators while accounting for chance agreement and missing data, and provide detailed per-dimension and per-language breakdowns in \autoref{sec:additional}.

\subsection{Pointwise vs.~Pairwise}
\label{sec:point_pair}
\begin{table}[t]
\caption{\label{tab:benchmark_main} 
Zero-shot and few-shot results of open-source and API models on the \dataset{} test set using \textbf{\textsc{Pointwise}} (grading single responses) and \textbf{\textsc{Pairwise}} (grading response pairs) scoring (see \S\ref{sec:point_pair}). 
\textit{Macro-F1} shows 5-way classification performance and \textit{Preference} reports accuracy on Win/Loss/Tie. 
Reported gains/loss are relative to zero-shot pointwise performance.
}
\vspace{0.5em}
\centering
\small
\resizebox{\textwidth}{!}{
\begin{tabular}{lcclclc}
\toprule
\multirow{3}{*}{\sc \textbf{ Models}} & \multicolumn{3}{c}{{\textbf{Macro F1}}}                                  & \multicolumn{3}{c}{{\textbf{Preference Accuracy}}}                       \\  
\cmidrule(lr){2-4}   \cmidrule(l){5-7}  
& \sc zero-shot  &\sc  few-shot  & \sc zero-shot  & \sc zero-shot  &\sc  few-shot  & \sc zero-shot  \\
& \sc  \textbf{Pointwise} & \sc \textbf{Pointwise} &\sc  \textbf{Pairwise} & \sc  \textbf{Pointwise}& \sc \textbf{Pointwise} &  \sc \textbf{Pairwise} \\
\cmidrule(r){1-1}   \cmidrule(lr){2-4}  \cmidrule(l){5-7}
    
\texttt{Qwen3-4B} & 23.06 
& 31.18\hspace{2pt}\textsubscript{\textcolor{darkgreen}{+8.12}} 
& 35.46\hspace{2pt}\textsubscript{\textcolor{darkgreen}{+12.40}} 
& 40.54 
& 39.35\hspace{2pt}\textsubscript{{-1.19}} 
& 57.13\hspace{2pt}\textsubscript{\textcolor{darkgreen}{+16.57}} \\

\texttt{Qwen3-32B} & 28.53 
& 35.45\hspace{2pt}\textsubscript{\textcolor{darkgreen}{+6.92}} 
& 37.48\hspace{2pt}\textsubscript{\textcolor{darkgreen}{+8.95}} 
& 42.19 
& 42.87\hspace{2pt}\textsubscript{\textcolor{darkgreen}{+0.68}} 
& 59.12\hspace{2pt}\textsubscript{\textcolor{darkgreen}{+16.59}} \\

\texttt{Llama-3.1-8B} & 22.27 
& 23.29\hspace{2pt}\textsubscript{\textcolor{darkgreen}{+1.02}} 
& 29.46\hspace{2pt}\textsubscript{\textcolor{darkgreen}{+7.19}} 
& 39.92 
& 37.15\hspace{2pt}\textsubscript{{-2.77}} 
& 50.45\hspace{2pt}\textsubscript{\textcolor{darkgreen}{+10.48}} \\

\texttt{Llama-3.3-70B} & 27.93 
& 30.52\hspace{2pt}\textsubscript{\textcolor{darkgreen}{+2.59}} 
& 37.50\hspace{2pt}\textsubscript{\textcolor{darkgreen}{+9.57}} 
& 37.37 
& 38.56\hspace{2pt}\textsubscript{\textcolor{darkgreen}{+1.19}} 
& 55.32\hspace{2pt}\textsubscript{\textcolor{darkgreen}{+17.89}} \\

\texttt{Llama4-Scout} & 25.63 
& 32.84\hspace{2pt}\textsubscript{\textcolor{darkgreen}{+7.21}} 
& 36.11\hspace{2pt}\textsubscript{\textcolor{darkgreen}{+10.48}} 
& 42.19 
& 41.22\hspace{2pt}\textsubscript{{-0.97}} 
& 56.25\hspace{2pt}\textsubscript{\textcolor{darkgreen}{+14.12}} \\ \midrule

\texttt{o3} & 26.54 
& 27.92\hspace{2pt}\textsubscript{\textcolor{darkgreen}{+1.38}} 
& 35.35\hspace{2pt}\textsubscript{\textcolor{darkgreen}{+8.81}} 
& 45.07 
& 44.68\hspace{2pt}\textsubscript{{-0.39}} 
& 58.72\hspace{2pt}\textsubscript{\textcolor{darkgreen}{+13.68}} \\

\texttt{gpt-4o} & 25.99 
& 29.57\hspace{2pt}\textsubscript{\textcolor{darkgreen}{+3.58}} 
& 37.57\hspace{2pt}\textsubscript{\textcolor{darkgreen}{+11.58}} 
& 42.92 
& 45.87\hspace{2pt}\textsubscript{\textcolor{darkgreen}{+2.95}} 
& 57.98\hspace{2pt}\textsubscript{\textcolor{darkgreen}{+15.09}} \\

\texttt{gpt-4.1} & 32.23 
& 33.84\hspace{2pt}\textsubscript{\textcolor{darkgreen}{+1.61}} 
& 38.53\hspace{2pt}\textsubscript{\textcolor{darkgreen}{+6.30}} 
& 41.73 
& 44.00\hspace{2pt}\textsubscript{\textcolor{darkgreen}{+2.27}} 
& 59.23\hspace{2pt}\textsubscript{\textcolor{darkgreen}{+17.50}} \\

\bottomrule  
\end{tabular}}
\end{table}

Although \dataset\ provides paired responses for each prompt, the presence of detailed \textit{grading rubrics} means that \emph{pointwise} evaluation is in principle sufficient: a model could \textit{assign absolute scores to individual responses without needing comparisons}. 
However, pairwise setups may provide stronger relative signals by anchoring judgments against another candidate. 
We therefore compare three setups:  
\textbf{Zero-shot pointwise}: the model is given a prompt, a \textit{single} response, and a detailed 5-point grading rubric, and asked to generate evaluation reasoning (in \textit{thinking}) and assign a final grade;  
\textbf{Few-shot pointwise}: we additionally provide three graded examples: one from 1--2, one with a grade of 3, and one from 4--5;
\textbf{Zero-shot pairwise}: the model is presented with \textit{both} responses to the same prompt and asked to assign a grade to each, following the template in \autoref{fig:template} (\autoref{sec:template}), without constraints on ties. 
The order of the two responses is randomized.

\autoref{tab:benchmark_main} reports Macro-F1 and Preference results. Zero-shot pairwise consistently outperforms both zero-shot and few-shot pointwise scoring across models, with gains of up to $+12.4\%$ in Macro-F1 and $+18.0\%$ in Preference accuracy over zero-shot pointwise. Few-shot pointwise improves Macro-F1 relative to zero-shot pointwise but yields only marginal gains in Preference, still falling short of zero-shot pairwise by an \textit{average} of $-5.5\%$ in Macro-F1 and $-15.1\%$ in Preference  across models.

These results indicate that models are substantially more reliable at assigning scores when evaluating two responses side by side, even without ground-truth labels. The unexpectedly large gains over few-shot in-context examples highlight \textbf{pairwise evaluation}, which explicitly anchors outputs against a competing candidate \citep{wang2025improving}, as a promising direction for improving automated judging reliability, even when the ultimate goal is pointwise scoring. We also evaluate few-shot pairwise on \texttt{Qwen3-4B}, observing only a small gain in Macro-F1 ($+0.6$) relative to zero-shot pairwise, further supporting our findings. Future work may investigate whether extending pairwise comparisons to a listwise evaluation of multiple responses offers additional benefits.

\subsection{With and without Grading Rubrics}
\label{sec:rubrics} 
\begin{table}[t]
\caption{\label{tab:benchmark_rubrics} 
Zero-shot performance comparing without and with detailed 5-Point Grading Rubrics. 
}
\setlength{\tabcolsep}{3pt}
\vspace{0.5em}
\centering
\small
\resizebox{\textwidth}{!}{
\begin{tabular}{lclclclcl}
\toprule

\multirow{3}{*}{\sc \textbf{Models}} & \multicolumn{4}{c}{{\textbf{Macro F1}}} & \multicolumn{4}{c}{{\textbf{Preference Accuracy}}} \\
\cmidrule(lr){2-5}   \cmidrule(l){6-9}  
& \multicolumn{2}{c}{\sc  \textbf{Pointwise}} & \multicolumn{2}{c}{\sc  \textbf{Pairwise}} & \multicolumn{2}{c}{\sc  \textbf{Pointwise}} & \multicolumn{2}{c}{\sc \textbf{Pairwise}} \\
& \textit{wo/ Rubrics} & \textit{w/ Rubrics} & \textit{wo/ Rubrics} & \textit{w/ Rubrics}  
& \textit{wo/ Rubrics} & \textit{w/ Rubrics} & \textit{wo/ Rubrics} & \textit{w/ Rubrics} \\

\cmidrule(r){1-1}   \cmidrule(lr){2-3}   \cmidrule(lr){4-5}   \cmidrule(lr){6-7}  \cmidrule(l){8-9}

\texttt{Qwen3-4B} & 16.00 
& 23.06\hspace{2pt}\textsubscript{\textcolor{darkgreen}{+7.06}} 
& 32.74 
& 35.46\hspace{2pt}\textsubscript{\textcolor{darkgreen}{+2.72}} 
& 33.52 
& 40.54\hspace{2pt}\textsubscript{\textcolor{darkgreen}{+7.02}} 
& 54.08 
& 57.13\hspace{2pt}\textsubscript{\textcolor{darkgreen}{+3.05}} \\

\texttt{Qwen3-32B} & 25.59 
& 28.53\hspace{2pt}\textsubscript{\textcolor{darkgreen}{+2.94}} 
& 38.10 
& 37.48\hspace{2pt}\textsubscript{{-0.62}} 
& 43.32 
& 42.19\hspace{2pt}\textsubscript{{-1.13}} 
& 59.23 
& 59.12\hspace{2pt}\textsubscript{{-0.11}} \\

\texttt{Llama-3.1-8B} & 21.50 
& 22.27\hspace{2pt}\textsubscript{\textcolor{darkgreen}{+0.77}} 
& 30.89 
& 29.46\hspace{2pt}\textsubscript{{-1.43}} 
& 38.34 
& 39.92\hspace{2pt}\textsubscript{\textcolor{darkgreen}{+1.58}} 
& 49.55 
& 50.45\hspace{2pt}\textsubscript{\textcolor{darkgreen}{+0.90}} \\

\texttt{Llama-3.3-70B} & 22.71 
& 27.93\hspace{2pt}\textsubscript{\textcolor{darkgreen}{+5.22}} 
& 35.12 
& 37.50\hspace{2pt}\textsubscript{\textcolor{darkgreen}{+2.38}} 
& 34.54 
& 37.37\hspace{2pt}\textsubscript{\textcolor{darkgreen}{+2.83}} 
& 56.29 
& 55.32\hspace{2pt}\textsubscript{{-0.97}} \\

\texttt{Llama4-Scout} & 22.15 
& 25.63\hspace{2pt}\textsubscript{\textcolor{darkgreen}{+3.48}} 
& 35.21 
& 36.11\hspace{2pt}\textsubscript{\textcolor{darkgreen}{+0.90}} 
& 41.28 
& 42.19\hspace{2pt}\textsubscript{\textcolor{darkgreen}{+0.91}} 
& 55.10 
& 56.25\hspace{2pt}\textsubscript{\textcolor{darkgreen}{+1.15}} \\ \midrule

\texttt{o3} & 25.43 
& 26.54\hspace{2pt}\textsubscript{\textcolor{darkgreen}{+1.11}} 
& 35.60 
& 35.35\hspace{2pt}\textsubscript{-0.25} 
& 45.13 
& 45.07\hspace{2pt}\textsubscript{-0.06} 
& 57.98 
& 58.72\hspace{2pt}\textsubscript{\textcolor{darkgreen}{+0.74}} \\

\texttt{gpt-4o} & 22.45 
& 25.99\hspace{2pt}\textsubscript{\textcolor{darkgreen}{+3.54}} 
& 36.74 
& 37.57\hspace{2pt}\textsubscript{\textcolor{darkgreen}{+0.83}} 
& 37.60 
& 42.92\hspace{2pt}\textsubscript{\textcolor{darkgreen}{+5.32}} 
& 56.85 
& 57.98\hspace{2pt}\textsubscript{\textcolor{darkgreen}{+1.13}} \\

\texttt{gpt-4.1} & 22.26 
& 32.23\hspace{2pt}\textsubscript{\textcolor{darkgreen}{+9.97}} 
& 37.35 
& 38.53\hspace{2pt}\textsubscript{\textcolor{darkgreen}{+1.18}} 
& 38.67 
& 41.73\hspace{2pt}\textsubscript{\textcolor{darkgreen}{+3.06}} 
& 56.96 
& 59.23\hspace{2pt}\textsubscript{\textcolor{darkgreen}{+2.27}} \\

\bottomrule  
\end{tabular}}
\end{table}

We further examine the role of detailed grading rubrics in judge performance. All rubrics are human-written 5-point guidelines specific to the dimension and question type of each prompt. Examples of dimension-specific rubrics are shown in Appendix \ref{sec:rubrics_example}.

\autoref{tab:benchmark_rubrics} compares zero-shot pointwise and pairwise performance with and without access to rubrics. The latter shows only the five class labels, without accompanying criteria or definitions.
Results show that rubrics provide a substantial benefit, especially for pointwise evaluation, yielding average gains of $+4.3\%$ in Macro-F1 and $+2.5\%$ in Preference accuracy. In contrast, pairwise evaluation benefits more modestly, with improvements of roughly $+1\%$ on both metrics.  

These findings suggest that judges perform better when grounded, either by explicit rubrics or by comparison with another response. Since pairwise comparison itself offers a strong grounding signal, it sees limited impact from rubrics. This highlights the importance of high-quality rubrics: if judges could automatically generate and evaluate high-quality, detailed rubrics, we hypothesize that the performance gap between pairwise and pointwise evaluation would further narrow.

\section{Training LLM-Judges on \dataset}
\label{sec:training}

Having established in \S\ref{sec:judges} that pairwise evaluation yields substantial advantages over pointwise scoring, we next examine whether training LLMs as judges can further close the gap to human annotators. We train on the \dataset\ training split (total 4,675 response pairs, where 232 pairs are held out for validation) and explore different learning strategies, model families, and reward designs. Inspired by the success of recent reasoning-based judges such as J1 \citep{whitehouse2025j1}, we compare supervised fine-tuning (SFT) and reinforcement learning (RL), as well as single-task (dimension-specific) and multi-task (all dimensions) training.

We fine-tune two contrasting models: \texttt{Qwen3-4B} (dense, reasoning-oriented) and \texttt{Llama4-Scout} (Mixture-of-Experts, non-reasoning), which differ in architecture and cognitive approach.
For \textbf{SFT}, models directly predict 5-point grades using cross-entropy loss under teacher forcing, without intermediate reasoning generation. For \textbf{RL}, we use GRPO \citep{shao2024deepseekmath} with the template from \autoref{fig:template}, encouraging step-by-step reasoning before score assignment. Following \citet{whitehouse2025j1}, we augment training data by including both response orders (A,B) and (B,A) to mitigate positional bias. Training details are provided in Appendix \ref{sec:appendix_judge_ft}.

\begin{table}[t]
\centering
\begin{minipage}{0.51\textwidth}
\caption{\label{tab:sft_rl} 
\textsc{Pairwise} SFT, RL, and SFT+RL-trained \texttt{Qwen3-4B} and \texttt{Llama4-Scout} results.
RL-trained models perform best overall.
}
\centering
\small
\setlength{\tabcolsep}{3pt}
\vspace{0.5em}
\resizebox{\textwidth}{!}{
\begin{tabular}{lllll}
\toprule
  {\multirow{2}{*}{{\sc Pairwise }}}  
& \multicolumn{2}{c}{\texttt{\textbf{Qwen3-4B}} }                              & \multicolumn{2}{c}{\texttt{\textbf{Llama4-Scout}}  }                                \\

& {Marco-F1}        & {Preference}        & {Marco-F1}           & {Preference}  \\
        
\cmidrule(r){1-1} \cmidrule(lr){2-3} \cmidrule(l){4-5}
\sc Zero-shot               & 35.46         & 57.13 &    36.11        & 56.25         \\
\sc SFT     & 33.44\hspace{1pt}\textsubscript{-2.02} 
& 53.68\hspace{1pt}\textsubscript{{-3.45}} 
& 44.17\hspace{1pt}\textsubscript{\textcolor{darkgreen}{+8.06}} 
& 60.08\hspace{1pt}\textsubscript{\textcolor{darkgreen}{+3.83}} \\

\sc RL       & \textbf{39.44}\hspace{1pt}\textsubscript{\textcolor{darkgreen}{+3.98}} 
& \textbf{60.02}\hspace{1pt}\textsubscript{\textcolor{darkgreen}{+2.89}}  
& 45.62\hspace{1pt}\textsubscript{\textcolor{darkgreen}{+9.51}} 
& \textbf{62.60}\hspace{1pt}\textsubscript{\textcolor{darkgreen}{+6.35}} \\ 

\sc SFT + RL & 39.33\hspace{1pt}\textsubscript{\textcolor{darkgreen}{+3.87}} 
& 58.78\hspace{1pt}\textsubscript{\textcolor{darkgreen}{+1.65}} 
& \textbf{45.82}\hspace{1pt}\textsubscript{\textcolor{darkgreen}{+9.71}} 
& 61.10\hspace{1pt}\textsubscript{\textcolor{darkgreen}{+4.85}} \\

\bottomrule
\end{tabular}
}

\end{minipage}
\hfill
\begin{minipage}{0.46\textwidth}
\caption{\label{tab:reward} 
Ablation of different reward designs for \textsc{Pairwise} RL-trained \texttt{Qwen3-4B}. \textit{Smooth}. and \textit{Prefer.} refer to \textit{Reward Smoothing} and \textit{Preference Bonus}. See \S\ref{sec:reward} for details.
}
\centering
\small
\vspace{0.5em}
\setlength{\tabcolsep}{6pt}
\resizebox{\textwidth}{!}{
\begin{tabular}{lll}
\toprule
\sc \textbf{Rewards} & \textbf{Marco-F1}  & \textbf{Preference}  \\
\midrule
\textit{Binary Only}  & 37.11 & 58.27  \\
\textit{Binary + Smooth.} & 37.30\hspace{2pt}\textsubscript{\textcolor{darkgreen}{+0.19}} & 51.47\hspace{2pt}\textsubscript{{-6.80}} \\
\textit{Binary + Prefer.} & 37.05\hspace{2pt}\textsubscript{{-0.06}} & \textbf{60.48}\hspace{2pt}\textsubscript{\textcolor{darkgreen}{+2.21}} \\
\textit{Binary + Smooth. + Prefer.} & \textbf{39.44}\hspace{2pt}\textsubscript{\textcolor{darkgreen}{+2.33}}  & 60.02\hspace{2pt}\textsubscript{\textcolor{darkgreen}{+1.75}} \\

\bottomrule
\end{tabular}
}
\end{minipage}
\end{table}

\subsection{Reward Designs for RL}
\label{sec:reward}

To make RL training effective, we design a composite reward signal that combines absolute accuracy with relative preference alignment and robustness to near-miss predictions:
(i) \textbf{Pointwise binary reward}: $+1$ if the predicted score matches the gold label, $0$ otherwise.  
(ii) \textbf{Reward smoothing}: partial reward ($+0.5$) if the prediction differs by exactly one grade.  
(iii) \textbf{Preference bonus}: additional $+1$ if the \emph{sign} of the difference between the two predicted scores matches the label.
(iv) \textbf{Penalties}: $-1$ for invalid or missing scores, and $-0.2$ for formatting violations, i.e. each tag must appear in the correct order and only once.

All reward components are summed to produce the final RL signal. Formally, the reward can be expressed as follows, where $s$ and $gt$ represent predicted and ground truth grades, respectively: 
{
\[
R = \sum_{i \in \{A,B\}} 
\underbrace{\max\Big( \mathbf{1}[s_i = gt_i],\ 0.5 \cdot \mathbf{1}[|s_i - gt_i| = 1] \Big)}_{\text{pointwise binary reward w/ reward smoothing}}
+ \underbrace{\mathbf{1}[\text{sign}(s_A - s_B) = \text{sign}(gt_A - gt_B)]}_{\text{preference bonus}}
\]
\[
- \underbrace{\mathbf{1}[\text{failed extraction}]}_{\text{extraction penalty}}
- \underbrace{0.2 \cdot \mathbf{1}[\text{formatting violation}]}_{\text{format penalty}}.
\]}

\subsection{Overall Performance: SFT vs. RL}

We first compare the overall performance of SFT and RL-trained models. \autoref{tab:sft_rl} shows that RL-trained \texttt{Qwen3-4B} and \texttt{Llama4-Scout} consistently outperform their SFT counterparts.
For inherently thinking models like \texttt{Qwen3-4B}, SFT without Chain-of-Thought (CoT) reasoning actually hurts performance, causing a $-2.0\%$ drop in Macro-F1 and $-3.5\%$ in Preference accuracy. In contrast, RL, which incentivizes reasoning, improves performance by $+4.0\%$ in Macro-F1 and $+2.9\%$ in Preference, surpassing the best frontier API model \texttt{gpt-4.1}.  

For non-thinking models like \texttt{Llama4-Scout}, SFT already provides substantial gains ($+8.1\%$ in Macro-F1 and $+3.8\%$ in Preference) compared to zero-shot. RL training further improves results, particularly in Preference ($+2.5\%$). This demonstrates the promise of pairwise RL training across model families, scales, and reasoning capabilities.

We also experimented with initializing RL from the best SFT checkpoint, but observed little or no improvement over starting RL from scratch. Models trained on SFT without CoT tend to copy the placeholder  ``\textit{\textless think\textgreater ~Your analysis and reasoning here.~\textless /think\textgreater}''
 from the prompt rather than generating meaningful reasoning, which limits the benefit of RL. This suggests that for tasks requiring reasoning, it is preferable to start RL directly when the SFT target lacks CoT supervision.

\label{sec:sft_rl}

\subsection{Ablation of RL Rewards}

In \autoref{tab:reward}, we ablate the RL reward design to validate the contribution of each reward component in RL training:  
(i) \textit{binary only}: reward $+1$ for exact score match, $0$ otherwise;  
(ii) \textit{binary+smooth.}: adds partial reward for near-miss scores, no preference bonus;  
(iii) \textit{binary+prefer.}: includes preference reward, no smoothing; and  
(iv) \textit{binary+smooth.+prefer.}: the default reward design in \S\ref{sec:reward}.
Results show clear benefits from combining reward smoothing and preference bonus, achieving the best overall Macro-F1 and Preference accuracy for \texttt{Qwen3-4B}, achieving $+2.3$ boost on Macro-F1 and $+1.7$ on preference accuracy over the \textit{binary only} reward.

\subsection{Per-Dimension Performance and Single vs. Multi-Task}

Next, we compare pairwise RL-trained \texttt{Qwen3-4B} models trained jointly across all dimensions versus individually per dimension.
Across the four dimensions, \textit{Tone}  achieves the strongest performance, with a Macro-F1 of 43.1 in the zero-shot setting and gains of up to $+3.8$ with multitask RL. \textit{Localized Tone} and \textit{Fluency} follow, reaching  $32.8$ and $32.2$ Macro-F1 in zero-shot, and up to $+5.7$ improvement when trained with multitask RL. In constrast, \textit{Localized Factuality} lags behind the other dimensions, achieving only $22.5$ Macro-F1 in zero-shot, a trend consistent across all models. Moreover, RL yields limited benefit ($+0.6$ in single-task RL) or even regressions in the multitask setup. These results highlight the challenge of localized factuality and suggest that alternative strategies, such as incorporating retrieval, search, or external tool use, may be necessary. Full results are provided in \autoref{tab:single_multi} in Appendix \ref{sec:additional}.  

Overall, aside from \textit{Localized Factuality}, joint multi-dimension training performs on par with single-dimension optimization while offering greater efficiency and practical benefits, such as serving as a reward model for post-training, which we explore in \S\ref{sec:rm}.

\subsection{Cross-Language Performance}

\begin{figure} [t]
    \centering
    \includegraphics[width=0.98\textwidth]{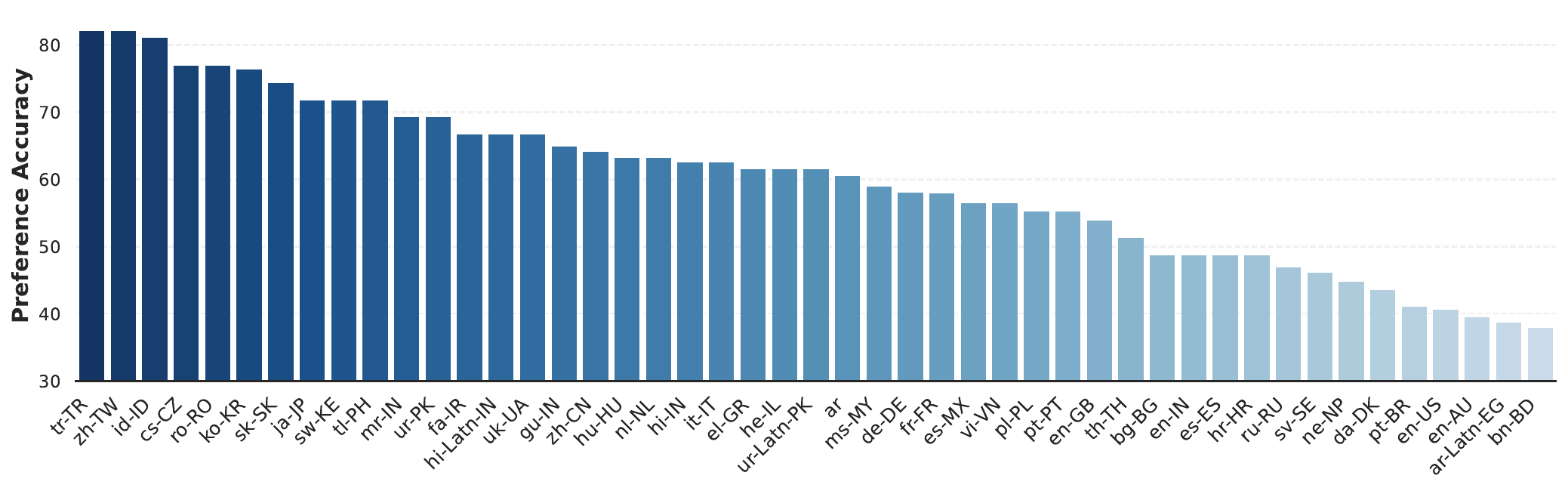}
    \caption{Preference Accuracy per Language of \textit{pairwise} RL-trained \texttt{Qwen3-4B}.}
    \label{fig:scores_locale}
\end{figure}

\autoref{fig:scores_locale} shows Preference accuracy per language variety for RL-trained \texttt{Qwen3-4B}. Performance varies widely, with \texttt{tr\_TR} at $82.1\%$ and \texttt{bn\_BD} at $37.9\%$, and does not strictly align with high- vs. low-resource languages. Relative to the zero-shot baseline, \texttt{en\_AU} and \texttt{fr\_FR} achieve the largest Macro-F1 gains ($+20.9\%$, $+17.7\%$) and \texttt{ro\_RO} and \texttt{gu\_IN} the largest Preference accuracy gains ($+18.0\%$, $+16.2\%$). By contrast, \texttt{es\_ES} drops $-15.4\%$ in Preference despite a modest $+2.2\%$ Macro-F1 gain, whereas \texttt{en\_MX}, the same language but a different locale, sees $+2.6\%$ and $+9.8\%$ gains in  Preference and Macro-F1 , highlighting that our dataset captures language variety nuances.

We further trained RL using only English data and evaluated on all languages. Performance degrades compared to the baseline, indicating that English-only training is insufficient to generalize across all 47 language varieties. Detailed per-language variety performance is provided in Appendix \ref{sec:per_locale}.

\section{From LLM-Judges to Reward Models}
\label{sec:rm}

We next investigate whether the RL-trained pairwise judges developed in our framework can also serve as generative reward models to directly improve LLM native-like response quality,  unifying evaluation and optimization in a single framework.

\subsection{RL with Judges as Generative Reward Models}
\label{sec:post}

For efficiency, we focus on smaller models for these experiments: \texttt{Qwen3-4B} as the policy model, and \texttt{Qwen3-4B-RL-Judge} as the reward model (RM). Since Localized Factuality remains challenging for our judges, we restrict both training and evaluation to Fluency, Tone, and Localized Tone. Specifically, we exclude Localized Factuality from all training and test prompts, randomly sample 3,000 prompts from \dataset for training, and retain all 1,398 test prompts from \dataset for evaluation across the three selected dimensions.

We post-train \texttt{Qwen3-4B} with GRPO. 
We sample 8 rollouts per prompt and compute rewards as follows: 
for each prompt, we construct response pairs from the rollouts,
format them with the same pairwise evaluation template, and feed them to the RM.
The final reward of each rollout is obtained by averaging its scores across all paired comparisons. Training details are added in Appendix \ref{sec:appendix_post}.

\subsection{Two-Stage Evaluation Strategy}

To rigorously evaluate the policy model's native-like quality improvements, we employ a two-stage validation approach: (i) comprehensive automated evaluation across all 47 language varieties using three diverse LLM judges, and (ii) human validation on a strategically selected subset of 10 high-resource languages where we can ensure annotation quality.

For each test prompt, we generate responses from both the baseline (\texttt{Qwen3-4B}) and post-trained (\texttt{Post-train}) models, construct response pairs with randomized order to mitigate positional bias, and apply the same pairwise judge template used in training.

\textbf{LLM-Judges Evaluation} We select three high-performing judges (see \S\ref{sec:judges}) \texttt{Qwen3-32B}, \texttt{gpt-4.1}, and \texttt{Llama4-Scout-RL-Judge}, and compute win, loss, and tie rates between baseline and post-trained models, along with average scores on a 1--5 scale across all 1,398 test prompts spanning 47 language varieties. \texttt{Qwen3-4B-RL-Judge} is excluded  from evaluation to avoid potential bias, since it serves as the RM.
\autoref{tab:post} shows that the post-trained policy model consistently outperforms the baseline across all LLM judges and languages. Average score improvements range from $+0.80$ to $+1.16$, with win rates between 63.4\% and 77.9\%. 
Per-dimension analysis reveals consistent gains across evaluation criteria: Tone yields the largest improvement ($+1.04$
average score boost), followed by Localized Tone and Fluency ($+0.89$ each). The consistency of improvements across different judge architectures and all three dimensions provides strong evidence for the effectiveness of our reward modeling approach.

\textbf{Human Validation}
To anchor our automated evaluation results, we conduct human evaluation on a diverse subset of 10 higher-resource languages: \texttt{ar}, \texttt{de\_DE}, \texttt{en\_US}, \texttt{fr\_FR},  \texttt{hi\_IN}, \texttt{hi\_Latn\_IN}, \texttt{pt\_BR}, \texttt{tl\_PH}, \texttt{th\_TH}, \texttt{vi\_VN}. This subset spans multiple language families, scripts, and geographic regions while ensuring access to qualified native speaker annotators. Human evaluation follows the same pairwise annotation guidelines as in \dataset construction.

\begin{table}[t]
\caption{\label{tab:post}
Two-stage Evaluation of \texttt{Qwen3-4B} and its RL post-trained variant \texttt{Post-train} on the \dataset test set, where both models serve as \textit{response models}. 
}
\setlength{\tabcolsep}{3pt}
\vspace{0.5em}
\centering
\small
\resizebox{\textwidth}{!}{
\begin{tabular}{lcccccccc}
\toprule

\multirow{2}{*}{\sc \textbf{Judges/Raters}} 
& \multirow{2}{*}{\# Languages} & \multicolumn{3}{c}{{\textbf{Win Rate}}}                                                     & \multicolumn{3}{c}{{\textbf{Average Score (1--5)}}}                                          \\

                        & &   \texttt{Post-train} Win & \texttt{Post-train} Loss & Tie &\texttt{Qwen3-4B} & \texttt{Post-train} 
                        &$\Delta Sore$
                        & Improvement\%
                        \\

    \cmidrule(r){1-2}   \cmidrule(lr){3-5}   \cmidrule(lr){6-9}                    

 \texttt{Llama4-Scout-RL-Judge} &47 &     63.88\%   & \phantom{0}9.16\%     & 26.96\%       &     3.01 & 3.79 & $+0.78$  & $+25.9\%$ \\
\texttt{Qwen3-32B } &47  &     72.46\%&21.89\%   & \phantom{0}6.65\%&  3.44 & 4.29 &$+0.85$ & $+24.7\%$ \\

\texttt{gpt-4.1 }   &47  &    77.90\%   &11.30\%&   10.80\%     &3.21 &4.37 &$+1.16$  & $+36.1\%$\\ 
    \midrule

 \texttt{Llama4-Scout-RL-Judge}   &10  &   69.66\%   &  \phantom{0}9.64 \%   &20.70\%  & 3.36& 4.22&$+0.86$  & $+25.6\%$  \\
  \textbf{Human Raters} &10  &  55.71\% & 35.20\% & \phantom{0}9.09\% & 3.31&3.67&$+0.36$& $+10.9\%$\\

\bottomrule  
\end{tabular}}
\end{table}

Results (last row of \autoref{tab:post}) on the subset confirms the automated evaluation trends. The post-trained model achieves a win rate of $55.7\%$ against the baseline, with an average score improvement of $+10.9\%$. While both automated and human evaluators agree that post-training improves response quality, we observe that LLM judges tend to overestimate the magnitude of improvement compared to human raters.
Comparing human evaluations to the closest-performing automated judge (\texttt{Llama4-Scout-RL-Judge}) on this subset reveals systematic differences: the automated judge reports an average improvement of $+0.5$ higher than humans. We hypothesize that this discrepancy arises because the automated judges may lean towards a stylistic caricature of native-like quality, overestimating improvements relative to nuanced human judgments. In addition, RL-trained judges exhibit less of this discrepancy among LLM evaluators, confirming the benefits of our RL judge training.

Overall, our two-stage evaluation demonstrates the potential of RL-trained judges as generative reward models for aligning multilingual outputs toward native-like quality. The directional consistency observed across both LLM- and human-based evaluations validates the viability of our unified framework for multilingual proficiency alignment. However, we note that challenges remain: LLM judges tend to overestimate the magnitude of improvements relative to human raters, highlighting an important direction for future work.

\section{Related Work} 
\vspace{-0.5em}
\paragraph{Multilingual Evaluation} Models' multilingual proficiency has been typically measured as an aggregate of performance across multiple task-oriented evaluations of short-form responses in settings with verifiable answers \citep{hu2020xtreme,ruder-etal-2021-xtreme,doddapaneni-etal-2023-towards,ahuja-etal-2023-mega,ahuja-etal-2024-megaverse}. Recent benchmarks focused on the evaluation of model's cultural knowledge in a similar verifiable setting \citep{myung2024blend,chiu-etal-2025-culturalbench,fabbri2025multinrc}. However, such evaluations do not extend to real-world conversations containing long-form responses. Benchmarks evaluating long-form responses use prompts and responses translated from English \citep{son2024mm,liu2024omgeval,doddapaneni-etal-2025-cross,gureja-etal-2025-rewardbench}. These evaluations typically do not reflect more localized aspects of language quality and are biased towards translationese. \citet{son2024mm} and \citet{doddapaneni-etal-2025-cross} automatically generate `good' and `bad' responses for each dimension.  \citet{marchisio-etal-2024-understanding} and \citet{guo-etal-2025-large} evaluate language consistency and naturalness respectively in relatively narrow settings. \textsc{Pariksha} \citep{watts-etal-2024-pariksha} is the most similar dataset to ours as it uses human-written prompts and human-annotated responses, but focuses on 10 Indic languages, annotates only high-level dimensions, and reports moderate inter-annotator agreement. \dataset is the only dataset that focuses on native-like quality in real-world conversations.

\paragraph{Multilingual Judges and RMs}
LLMs have been used as judges in different multilingual benchmarks \citep{liu2024omgeval,fabbri2025multinrc}. However, fewer works focus on analyzing or improving multilingual judges and RMs. \citet{gureja-etal-2025-rewardbench} observe that zero-shot judges show a substantial gap between the translated \textsc{M-RewardBench} and its English counterpart, with predictions inconsistent across languages. \citet{fu2025reliable} report similar inconsistencies across five diverse tasks. \citet{wu-etal-2024-reuse} evaluate zero-shot cross-lingual transfer of trained RMs on summarization and dialog, observing gains. \citet{hong-etal-2025-cross} find strong cross-lingual transfer on \textsc{M-RewardBench} for English RMs fine-tuned in four languages. \citet{doddapaneni-etal-2025-cross} fine-tune a judge with SFT on automatically translated prompts and responses in six languages to produce an absolute score. To our knowledge, we are the \textit{first} to (i) train judges and RMs in a massively multilingual setting, (ii) fine-tune multilingual judges with RL, and (iii) demonstrate the benefits of multi-task RL, reward shaping, and pairwise grading in this setting.

\section{Conclusion}
\vspace{-0.5em}
We introduce \dataset, a comprehensive framework for evaluating and improving native-like response quality across 47 language varieties. By combining sociolinguistically-informed prompt design, detailed evaluation rubrics, and high-quality human annotations, \dataset captures multiple dimensions of conversational proficiency, including fluency, tone, localized tone, and localized factuality. We demonstrate that pairwise evaluation significantly improves both zero-shot and fine-tuned LLM judges, and that RL with reward shaping yields best judge performance.

Beyond evaluation, we show these trained judges can serve as generative reward models to directly improve policy model's response quality. While challenges remain with the tendency of LLM judges to overestimate improvements relative to human raters, our framework provides a practical and scalable approach to both assessing and enhancing LLM proficiency in multilingual context. 

\subsubsection*{Acknowledgments}

We would like to thank Pritish Yuvraj for help with initial generations. We thank Max Mauer and Nidhi Nisarg Shah for support on the annotation interface. We would like to thank Wes Kranz, Lora Zhou, and Kish Patel for help on operations. We thank the Multilingual Post-training team for discussions and feedback.

\subsection*{Ethics statement}
Translators and annotators were recruited through third-party services and compensated based on local regulations. 

\subsection*{Reproducibility statement}
We release the full \dataset to the community. All models are built on top of open-weight Llama and Qwen backbones. The prompt templates used for training are provided in  \autoref{sec:template}, and detailed descriptions of experimental setups, hyperparameters, and libraries are included in \autoref{sec:exp}.

\bibliography{iclr2026_conference}

@article{shao2024deepseekmath,
  title={{DeepseekMath: Pushing the Limits of Mathematical Reasoning in Open Language Models}},
  author={Shao, Zhihong and Wang, Peiyi and Zhu, Qihao and Xu, Runxin and Song, Junxiao and Bi, Xiao and Zhang, Haowei and Zhang, Mingchuan and Li, YK and Wu, Y and others},
  journal={arXiv preprint arXiv:2402.03300},
  year={2024},
  url={https://arxiv.org/abs/2402.03300}
}

@inproceedings{
whitehouse2025j1,
  title={{J1: Incentivizing Thinking in LLM-as-a-Judge via Reinforcement Learning}},
author={Chenxi Whitehouse and Tianlu Wang and Ping Yu and Xian Li and Jason E Weston and Ilia Kulikov and Swarnadeep Saha},
booktitle={The Fourteenth International Conference on Learning Representations},
year={2026},
url={https://openreview.net/forum?id=dnJEHl6DI1}
}

@inproceedings{gureja-etal-2025-rewardbench,
    title = {{{M}-{R}eward{B}ench: Evaluating Reward Models in Multilingual Settings}},
    author = "Gureja, Srishti  and
      Miranda, Lester James Validad  and
      Islam, Shayekh Bin  and
      Maheshwary, Rishabh  and
      Sharma, Drishti  and
      Winata, Gusti Triandi  and
      Lambert, Nathan  and
      Ruder, Sebastian  and
      Hooker, Sara  and
      Fadaee, Marzieh",
    editor = "Che, Wanxiang  and
      Nabende, Joyce  and
      Shutova, Ekaterina  and
      Pilehvar, Mohammad Taher",
    booktitle = "Proceedings of the 63rd Annual Meeting of the Association for Computational Linguistics (Volume 1: Long Papers)",
    month = jul,
    year = "2025",
    address = "Vienna, Austria",
    publisher = "Association for Computational Linguistics",
    url = "https://aclanthology.org/2025.acl-long.3/",
    doi = "10.18653/v1/2025.acl-long.3",
    pages = "43--58",
    ISBN = "979-8-89176-251-0",
}

@inproceedings{
myung2024blend,
title={{{BLE}nD: A Benchmark for {LLM}s on Everyday Knowledge in Diverse Cultures and Languages}},
author={Junho Myung and Nayeon Lee and Yi Zhou and Jiho Jin and Rifki Afina Putri and Dimosthenis Antypas and Hsuvas Borkakoty and Eunsu Kim and Carla Perez-Almendros and Abinew Ali Ayele and Victor Gutierrez Basulto and Yazmin Ibanez-Garcia and Hwaran Lee and Shamsuddeen Hassan Muhammad and Kiwoong Park and Anar Sabuhi Rzayev and Nina White and Seid Muhie Yimam and Mohammad Taher Pilehvar and Nedjma Ousidhoum and Jose Camacho-Collados and Alice Oh},
booktitle={The Thirty-eight Conference on Neural Information Processing Systems Datasets and Benchmarks Track},
year={2024},
url={https://openreview.net/forum?id=nrEqH502eC}
}

@inproceedings{doddapaneni-etal-2025-cross,
    title = {{Cross-Lingual Auto Evaluation for Assessing Multilingual {LLM}s}},
    author = "Doddapaneni, Sumanth  and
      Khan, Mohammed Safi Ur Rahman  and
      Venkatesh, Dilip  and
      Dabre, Raj  and
      Kunchukuttan, Anoop  and
      Khapra, Mitesh M",
    editor = "Che, Wanxiang  and
      Nabende, Joyce  and
      Shutova, Ekaterina  and
      Pilehvar, Mohammad Taher",
    booktitle = "Proceedings of the 63rd Annual Meeting of the Association for Computational Linguistics (Volume 1: Long Papers)",
    month = jul,
    year = "2025",
    address = "Vienna, Austria",
    publisher = "Association for Computational Linguistics",
    url = "https://aclanthology.org/2025.acl-long.1419/",
    doi = "10.18653/v1/2025.acl-long.1419",
    pages = "29297--29329",
    ISBN = "979-8-89176-251-0",
}

@inproceedings{hong-etal-2025-cross,
    title = {{Cross-lingual Transfer of Reward Models in Multilingual Alignment}},
    author = "Hong, Jiwoo  and
      Lee, Noah  and
      Mart{\'i}nez-Casta{\~n}o, Rodrigo  and
      Rodr{\'i}guez, C{\'e}sar  and
      Thorne, James",
    editor = "Chiruzzo, Luis  and
      Ritter, Alan  and
      Wang, Lu",
    booktitle = "Proceedings of the 2025 Conference of the Nations of the Americas Chapter of the Association for Computational Linguistics: Human Language Technologies (Volume 2: Short Papers)",
    month = apr,
    year = "2025",
    address = "Albuquerque, New Mexico",
    publisher = "Association for Computational Linguistics",
    url = "https://aclanthology.org/2025.naacl-short.8/",
    doi = "10.18653/v1/2025.naacl-short.8",
    pages = "82--94",
    ISBN = "979-8-89176-190-2",
}

@inproceedings{wu-etal-2024-reuse,
    title = {{Reuse Your Rewards: Reward Model Transfer for Zero-Shot Cross-Lingual Alignment}},
    author = "Wu, Zhaofeng  and
      Balashankar, Ananth  and
      Kim, Yoon  and
      Eisenstein, Jacob  and
      Beirami, Ahmad",
    editor = "Al-Onaizan, Yaser  and
      Bansal, Mohit  and
      Chen, Yun-Nung",
    booktitle = "Proceedings of the 2024 Conference on Empirical Methods in Natural Language Processing",
    month = nov,
    year = "2024",
    address = "Miami, Florida, USA",
    publisher = "Association for Computational Linguistics",
    url = "https://aclanthology.org/2024.emnlp-main.79/",
    doi = "10.18653/v1/2024.emnlp-main.79",
    pages = "1332--1353",
}

@article{bell1984language,
  title={{Language Style as Audience Design}},
  author={Bell, Allan},
  journal={Language in society},
  volume={13},
  number={2},
  pages={145--204},
  year={1984},
  publisher={Cambridge University Press}
}

@article{jamieson2000toefl,
  title={{TOEFL 2000 Framework}},
  author={Jamieson, Joan and Jones, Stan and Kirsch, Irwin and Mosenthal, Peter and Taylor, Carol},
  journal={Princeton, NJ: Educational Testing Service},
  year={2000}
}

@inproceedings{ke2019automated,
  title={Automated Essay Scoring: A Survey of the State of the Art.},
  author={Ke, Zixuan and Ng, Vincent},
  booktitle={IJCAI},
  volume={19},
  pages={6300--6308},
  year={2019}
}

@inproceedings{hershcovich-etal-2022-challenges,
    title = {{Challenges and Strategies in Cross-Cultural {NLP}}},
    author = "Hershcovich, Daniel  and
      Frank, Stella  and
      Lent, Heather  and
      de Lhoneux, Miryam  and
      Abdou, Mostafa  and
      Brandl, Stephanie  and
      Bugliarello, Emanuele  and
      Cabello Piqueras, Laura  and
      Chalkidis, Ilias  and
      Cui, Ruixiang  and
      Fierro, Constanza  and
      Margatina, Katerina  and
      Rust, Phillip  and
      S{\o}gaard, Anders",
    editor = "Muresan, Smaranda  and
      Nakov, Preslav  and
      Villavicencio, Aline",
    booktitle = "Proceedings of the 60th Annual Meeting of the Association for Computational Linguistics (Volume 1: Long Papers)",
    month = may,
    year = "2022",
    address = "Dublin, Ireland",
    publisher = "Association for Computational Linguistics",
    url = "https://aclanthology.org/2022.acl-long.482/",
    doi = "10.18653/v1/2022.acl-long.482",
    pages = "6997--7013"
}

@article{hupkes2025multiloko,
  title={{MultiLoKo: A Multilingual Local Knowledge Benchmark for LLMs Spanning 31 Languages}},
  author={Hupkes, Dieuwke and Bogoychev, Nikolay},
  journal={arXiv preprint arXiv:2504.10356},
  year={2025},
url={https://arxiv.org/abs/2504.10356}
}

@inproceedings{watts-etal-2024-pariksha,
    title = {{{PARIKSHA}: A Large-Scale Investigation of Human-{LLM} Evaluator Agreement on Multilingual and Multi-Cultural Data}},
    author = "Watts, Ishaan  and
      Gumma, Varun  and
      Yadavalli, Aditya  and
      Seshadri, Vivek  and
      Swaminathan, Manohar  and
      Sitaram, Sunayana",
    editor = "Al-Onaizan, Yaser  and
      Bansal, Mohit  and
      Chen, Yun-Nung",
    booktitle = "Proceedings of the 2024 Conference on Empirical Methods in Natural Language Processing",
    month = nov,
    year = "2024",
    address = "Miami, Florida, USA",
    publisher = "Association for Computational Linguistics",
    url = "https://aclanthology.org/2024.emnlp-main.451/",
    doi = "10.18653/v1/2024.emnlp-main.451",
    pages = "7900--7932",
}

@inproceedings{sorensen2024roadmap,
  title={{A Roadmap to Pluralistic Alignment}},
author={Taylor Sorensen and Jared Moore and Jillian Fisher and Mitchell L Gordon and Niloofar Mireshghallah and Christopher Michael Rytting and Andre Ye and Liwei Jiang and Ximing Lu and Nouha Dziri and Tim Althoff and Yejin Choi},
booktitle={Forty-first International Conference on Machine Learning},
year={2024},
url={https://openreview.net/forum?id=gQpBnRHwxM}
}

@inproceedings{hu2020xtreme,
  title={{Xtreme: A Massively Multilingual Multi-task Benchmark for Evaluating Cross-lingual Generalisation}},
  author={Hu, Junjie and Ruder, Sebastian and Siddhant, Aditya and Neubig, Graham and Firat, Orhan and Johnson, Melvin},
  booktitle={International conference on machine learning},
  pages={4411--4421},
  year={2020},
  organization={PMLR},
url={https://proceedings.mlr.press/v119/hu20b}
}

@inproceedings{ruder-etal-2021-xtreme,
    title = {{{XTREME}-{R}: Towards More Challenging and Nuanced Multilingual Evaluation}},
    author = "Ruder, Sebastian  and
      Constant, Noah  and
      Botha, Jan  and
      Siddhant, Aditya  and
      Firat, Orhan  and
      Fu, Jinlan  and
      Liu, Pengfei  and
      Hu, Junjie  and
      Garrette, Dan  and
      Neubig, Graham  and
      Johnson, Melvin",
    editor = "Moens, Marie-Francine  and
      Huang, Xuanjing  and
      Specia, Lucia  and
      Yih, Scott Wen-tau",
    booktitle = "Proceedings of the 2021 Conference on Empirical Methods in Natural Language Processing",
    month = nov,
    year = "2021",
    address = "Online and Punta Cana, Dominican Republic",
    publisher = "Association for Computational Linguistics",
    url = "https://aclanthology.org/2021.emnlp-main.802/",
    doi = "10.18653/v1/2021.emnlp-main.802",
    pages = "10215--10245",
}

@inproceedings{ahuja-etal-2024-megaverse,
    title = {{{MEGAVERSE}: Benchmarking Large Language Models Across Languages, Modalities, Models and Tasks}},
    author = "Ahuja, Sanchit  and
      Aggarwal, Divyanshu  and
      Gumma, Varun  and
      Watts, Ishaan  and
      Sathe, Ashutosh  and
      Ochieng, Millicent  and
      Hada, Rishav  and
      Jain, Prachi  and
      Ahmed, Mohamed  and
      Bali, Kalika  and
      Sitaram, Sunayana",
    editor = "Duh, Kevin  and
      Gomez, Helena  and
      Bethard, Steven",
    booktitle = "Proceedings of the 2024 Conference of the North American Chapter of the Association for Computational Linguistics: Human Language Technologies (Volume 1: Long Papers)",
    month = jun,
    year = "2024",
    address = "Mexico City, Mexico",
    publisher = "Association for Computational Linguistics",
    url = "https://aclanthology.org/2024.naacl-long.143/",
    doi = "10.18653/v1/2024.naacl-long.143",
    pages = "2598--2637",
}

@inproceedings{chiu-etal-2025-culturalbench,
    title = {{{C}ultural{B}ench: A Robust, Diverse and Challenging Benchmark for Measuring {LM}s' Cultural Knowledge Through Human-{AI} Red-Teaming}},
    author = "Chiu, Yu Ying  and
      Jiang, Liwei  and
      Lin, Bill Yuchen  and
      Park, Chan Young  and
      Li, Shuyue Stella  and
      Ravi, Sahithya  and
      Bhatia, Mehar  and
      Antoniak, Maria  and
      Tsvetkov, Yulia  and
      Shwartz, Vered  and
      Choi, Yejin",
    editor = "Che, Wanxiang  and
      Nabende, Joyce  and
      Shutova, Ekaterina  and
      Pilehvar, Mohammad Taher",
    booktitle = "Proceedings of the 63rd Annual Meeting of the Association for Computational Linguistics (Volume 1: Long Papers)",
    month = jul,
    year = "2025",
    address = "Vienna, Austria",
    publisher = "Association for Computational Linguistics",
    url = "https://aclanthology.org/2025.acl-long.1247/",
    doi = "10.18653/v1/2025.acl-long.1247",
    pages = "25663--25701",
    ISBN = "979-8-89176-251-0",
}

@inproceedings{doddapaneni-etal-2023-towards,
    title = {{Towards Leaving No {I}ndic Language Behind: Building Monolingual Corpora, Benchmark and Models for {I}ndic Languages}},
    author = "Doddapaneni, Sumanth  and
      Aralikatte, Rahul  and
      Ramesh, Gowtham  and
      Goyal, Shreya  and
      Khapra, Mitesh M.  and
      Kunchukuttan, Anoop  and
      Kumar, Pratyush",
    editor = "Rogers, Anna  and
      Boyd-Graber, Jordan  and
      Okazaki, Naoaki",
    booktitle = "Proceedings of the 61st Annual Meeting of the Association for Computational Linguistics (Volume 1: Long Papers)",
    month = jul,
    year = "2023",
    address = "Toronto, Canada",
    publisher = "Association for Computational Linguistics",
    url = "https://aclanthology.org/2023.acl-long.693/",
    doi = "10.18653/v1/2023.acl-long.693",
    pages = "12402--12426",
}

@inproceedings{ahuja-etal-2023-mega,
    title = {{MEGA}: Multilingual Evaluation of Generative {AI}},
    author = "Ahuja, Kabir  and
      Diddee, Harshita  and
      Hada, Rishav  and
      Ochieng, Millicent  and
      Ramesh, Krithika  and
      Jain, Prachi  and
      Nambi, Akshay  and
      Ganu, Tanuja  and
      Segal, Sameer  and
      Ahmed, Mohamed  and
      Bali, Kalika  and
      Sitaram, Sunayana",
    editor = "Bouamor, Houda  and
      Pino, Juan  and
      Bali, Kalika",
    booktitle = "Proceedings of the 2023 Conference on Empirical Methods in Natural Language Processing",
    month = dec,
    year = "2023",
    address = "Singapore",
    publisher = "Association for Computational Linguistics",
    url = "https://aclanthology.org/2023.emnlp-main.258/",
    doi = "10.18653/v1/2023.emnlp-main.258",
    pages = "4232--4267",
}

@article{liu2024omgeval,
  title={{OMGEval: An Open Multilingual Generative Evaluation Benchmark for Large Language Models}},
  author={Liu, Yang and Xu, Meng and Wang, Shuo and Yang, Liner and Wang, Haoyu and Liu, Zhenghao and Kong, Cunliang and Chen, Yun and Sun, Maosong and Yang, Erhong},
  journal={arXiv preprint arXiv:2402.13524},
  year={2024},
url={https://arxiv.org/abs/2402.13524},
}

@article{son2024mm,
  title={{MM-Eval: A Multilingual Meta-Evaluation Benchmark for LLM-as-a-Judge and Reward Models}},
  author={Son, Guijin and Yoon, Dongkeun and Suk, Juyoung and Aula-Blasco, Javier and Aslan, Mano and Kim, Vu Trong and Islam, Shayekh Bin and Prats-Cristi{\`a}, Jaume and Tormo-Ba{\~n}uelos, Luc{\'\i}a and Kim, Seungone},
  journal={arXiv preprint arXiv:2410.17578},
url={https://arxiv.org/abs/2410.17578},
  year={2024}
}

@article{fabbri2025multinrc,
  title={MultiNRC: A Challenging and Native Multilingual Reasoning Evaluation Benchmark for LLMs},
  author={Fabbri, Alexander R and Mares, Diego and Flores, Jorge and Mankikar, Meher and Hernandez, Ernesto and Lee, Dean and Liu, Bing and Xing, Chen},
  journal={arXiv preprint arXiv:2507.17476},
url={https://arxiv.org/abs/2507.17476},
  year={2025}
}

@inproceedings{marchisio-etal-2024-understanding,
    title = {{Understanding and Mitigating Language Confusion in {LLM}s}},
    author = "Marchisio, Kelly  and
      Ko, Wei-Yin  and
      Berard, Alexandre  and
      Dehaze, Th{\'e}o  and
      Ruder, Sebastian",
    editor = "Al-Onaizan, Yaser  and
      Bansal, Mohit  and
      Chen, Yun-Nung",
    booktitle = "Proceedings of the 2024 Conference on Empirical Methods in Natural Language Processing",
    month = nov,
    year = "2024",
    address = "Miami, Florida, USA",
    publisher = "Association for Computational Linguistics",
    url = "https://aclanthology.org/2024.emnlp-main.380/",
    doi = "10.18653/v1/2024.emnlp-main.380",
    pages = "6653--6677",
}

@article{jacovi2025facts,
  title={{The FACTS Grounding Leaderboard: Benchmarking LLMs' Ability to Ground Responses to Long-Form Input}},
  author={Jacovi, Alon and Wang, Andrew and Alberti, Chris and Tao, Connie and Lipovetz, Jon and Olszewska, Kate and Haas, Lukas and Liu, Michelle and Keating, Nate and Bloniarz, Adam and others},
  journal={arXiv preprint arXiv:2501.03200},
url={https://arxiv.org/abs/2501.03200},
  year={2025}
}

@inproceedings{
zhuang2024hydra,
title={{{HYDRA}: Model Factorization Framework for Black-Box {LLM} Personalization}},
author={Yuchen Zhuang and Haotian Sun and Yue Yu and Rushi Qiang and Qifan Wang and Chao Zhang and Bo Dai},
booktitle={The Thirty-eighth Annual Conference on Neural Information Processing Systems},
year={2024},
url={https://openreview.net/forum?id=CKgNgKmHYp}
}

@article{anil2023palm,
  title={{Palm 2 Technical Report}},
  author={Anil, Rohan and Dai, Andrew M and Firat, Orhan and Johnson, Melvin and Lepikhin, Dmitry and Passos, Alexandre and Shakeri, Siamak and Taropa, Emanuel and Bailey, Paige and Chen, Zhifeng and others},
  journal={arXiv preprint arXiv:2305.10403},
  year={2023},
  url={https://arxiv.org/abs/2305.10403}
}

@article{lothritz2025testing,
  title={{Testing Low-Resource Language Support in LLMs Using Language Proficiency Exams: the Case of Luxembourgish}},
  author={Lothritz, Cedric and Cabot, Jordi},
  journal={arXiv preprint arXiv:2504.01667},
  year={2025},
url={https://arxiv.org/abs/2304.01667}
}

@inproceedings{guo-etal-2025-large,
    title = "Do Large Language Models have an {E}nglish Accent? Evaluating and Improving the Naturalness of Multilingual {LLM}s",
    author = "Guo, Yanzhu  and
      Conia, Simone  and
      Zhou, Zelin  and
      Li, Min  and
      Potdar, Saloni  and
      Xiao, Henry",
    editor = "Che, Wanxiang  and
      Nabende, Joyce  and
      Shutova, Ekaterina  and
      Pilehvar, Mohammad Taher",
    booktitle = "Proceedings of the 63rd Annual Meeting of the Association for Computational Linguistics (Volume 1: Long Papers)",
    month = jul,
    year = "2025",
    address = "Vienna, Austria",
    publisher = "Association for Computational Linguistics",
    url = "https://aclanthology.org/2025.acl-long.193/",
    doi = "10.18653/v1/2025.acl-long.193",
    pages = "3823--3838",
    ISBN = "979-8-89176-251-0"
}

@inproceedings{liu-etal-2021-naturalness,
    title = "Naturalness Evaluation of Natural Language Generation in Task-oriented Dialogues Using {BERT}",
    author = "Liu, Ye  and
      Maier, Wolfgang  and
      Minker, Wolfgang  and
      Ultes, Stefan",
    editor = "Mitkov, Ruslan  and
      Angelova, Galia",
    booktitle = "Proceedings of the International Conference on Recent Advances in Natural Language Processing (RANLP 2021)",
    month = sep,
    year = "2021",
    address = "Held Online",
    publisher = "INCOMA Ltd.",
    url = "https://aclanthology.org/2021.ranlp-1.96/",
    pages = "839--845"
}

@inproceedings{novikova-etal-2016-crowd,
    title = "Crowd-sourcing {NLG} Data: Pictures Elicit Better Data.",
    author = "Novikova, Jekaterina  and
      Lemon, Oliver  and
      Rieser, Verena",
    editor = "Isard, Amy  and
      Rieser, Verena  and
      Gkatzia, Dimitra",
    booktitle = "Proceedings of the 9th International Natural Language Generation conference",
    month = sep # " 5-8",
    year = "2016",
    address = "Edinburgh, UK",
    publisher = "Association for Computational Linguistics",
    url = "https://aclanthology.org/W16-6644/",
    doi = "10.18653/v1/W16-6644",
    pages = "265--273"
}

@article{mayor2024evaluating,
  title={{Evaluating Large Language Models with Tests of Spanish as a Foreign Language: Pass or Fail?}},
  author={Mayor-Rocher, Marina and Melero, Nina and Merino-G{\'o}mez, Elena and Grandury, Mar{\'\i}a and Conde, Javier and Reviriego, Pedro},
  journal={arXiv preprint arXiv:2409.15334},
  year={2024},
url={https://arxiv.org/abs/2409.15334}
}

@inproceedings{fu2025reliable,
  title={{How Reliable is Multilingual LLM-as-a-Judge?}},
  author={Fu, Xiyan and Liu, Wei},
    booktitle = "Findings of the Association for Computational Linguistics: EMNLP 2025",
    month = nov,
    year = "2025",
    address = "Suzhou, China",
    publisher = "Association for Computational Linguistics",
    url = "https://aclanthology.org/2025.findings-emnlp.587/",
    doi = "10.18653/v1/2025.findings-emnlp.587",
    pages = "11040--11053",
    ISBN = "979-8-89176-335-7",
}

@inproceedings{
liu2025understanding,
  title={{Understanding R1-Zero-Like Training: A Critical Perspective}},
author={Zichen Liu and Changyu Chen and Wenjun Li and Penghui Qi and Tianyu Pang and Chao Du and Wee Sun Lee and Min Lin},
booktitle={Second Conference on Language Modeling},
year={2025},
url={https://openreview.net/forum?id=5PAF7PAY2Y}
}

@inproceedings{wang2025improving,
  title={{Improving LLM-as-a-Judge Inference with the Judgment Distribution}},
    author = "Wang, Victor  and
      Zhang, Michael JQ  and
      Choi, Eunsol",
    booktitle = "Findings of the Association for Computational Linguistics: EMNLP 2025",
    month = nov,
    year = "2025",
    address = "Suzhou, China",
    publisher = "Association for Computational Linguistics",
    url = "https://aclanthology.org/2025.findings-emnlp.1259/",
    doi = "10.18653/v1/2025.findings-emnlp.1259",
    pages = "23173--23199",
    ISBN = "979-8-89176-335-7",
}

@inproceedings{kim-etal-2025-biggen,
    title = {{The {B}i{GG}en Bench: A Principled Benchmark for Fine-grained Evaluation of Language Models with Language Models}},
    author = "Kim, Seungone  and
      Suk, Juyoung  and
      Cho, Ji Yong  and
      Longpre, Shayne  and
      Kim, Chaeeun  and
      Yoon, Dongkeun  and
      Son, Guijin  and
      Cho, Yejin  and
      Shafayat, Sheikh  and
      Baek, Jinheon  and
      Park, Sue Hyun  and
      Hwang, Hyeonbin  and
      Jo, Jinkyung  and
      Cho, Hyowon  and
      Shin, Haebin  and
      Lee, Seongyun  and
      Oh, Hanseok  and
      Lee, Noah  and
      Ho, Namgyu  and
      Joo, Se June  and
      Ko, Miyoung  and
      Lee, Yoonjoo  and
      Chae, Hyungjoo  and
      Shin, Jamin  and
      Jang, Joel  and
      Ye, Seonghyeon  and
      Lin, Bill Yuchen  and
      Welleck, Sean  and
      Neubig, Graham  and
      Lee, Moontae  and
      Lee, Kyungjae  and
      Seo, Minjoon",
    editor = "Chiruzzo, Luis  and
      Ritter, Alan  and
      Wang, Lu",
    booktitle = "Proceedings of the 2025 Conference of the Nations of the Americas Chapter of the Association for Computational Linguistics: Human Language Technologies (Volume 1: Long Papers)",
    month = apr,
    year = "2025",
    address = "Albuquerque, New Mexico",
    publisher = "Association for Computational Linguistics",
    url = "https://aclanthology.org/2025.naacl-long.303/",
    doi = "10.18653/v1/2025.naacl-long.303",
    pages = "5877--5919",
    ISBN = "979-8-89176-189-6",
}
\bibliographystyle{iclr2026_conference}
\appendix
\section*{Appendix}

\section{Additional Details on the \dataset Dataset}
\label{sec:data_detail}

\subsection{Dataset Collection}
\label{sec:data_anno}

\paragraph{Localization vs natively written prompts} Our prompts derive from English templates to ensure coverage consistency across a large number of languages. While localization includes cultural adaptation, we acknowledge that independently authored prompts in each language would better capture native discourse structures. We focus our resources on native speakers annotating responses instead.

We initiated a pilot to evaluate of different models including GPT-4o and Llama4-Maverick in a single category, Localized Tone focusing on five languages spoken by the authors: Bengali, German, Hindi, Italian, and Russian. As expected, the pilot framework was quickly confronted with the complexities inherent in multilingualism: Even among the authors, for all initial Localized Tone prompts, we struggled to reach consistent and reliable agreement. Nevertheless, these early results provided valuable insights to guide us to improve prompt design, guideline clarity, and annotation arrangement.

To enhance the reliability of the framework, we took steps to refine our prompts' nuance and complexity, update guidelines with clearer direction for annotators that aimed to make abstract concepts more concrete, and started exploring a more user-friendly annotation solution. The changes brought upon notable improvements in inter-annotator agreement that extended to the full-scale annotation. We show the agreement of the initial pilot annotation and improved annotation for localized tone in 5 languages in \autoref{tab:pilot_menlo_comparison} and across all categories and languages in \autoref{tab:pilot_menlo_comparison}.

\begin{table}[H]
\caption{\label{tab:pilot_menlo_comparison}
Comparison of \textsc{Pilot} and \dataset for 5 languages in localized tone category. \textit{\footnotesize Agreement is defined as the percentage of annotation pairs whose ratings for the same item differ by no more than 1.}
}
\vspace{0.5em}
\centering
\small
{
\begin{tabular}{lcccccc}
\toprule
\textbf{Language Code} & \textbf{\textsc{Pilot} Agreement} & \textbf{\dataset Agreement} \\
\midrule
\texttt{bn\_BD}  & 0.75 & 0.84  \\
\texttt{de\_DE}  & 0.74 & 0.92  \\
\texttt{hi\_IN}  & 0.74 & 0.92  \\
\texttt{it\_IT}  & 0.79 & 0.71  \\
\texttt{ru\_RU}  & 0.71 & 0.79  \\
\midrule
Overall          & 0.75 &  0.84 \\
\bottomrule
\end{tabular}
}
\end{table}
\begin{table}[H]
\caption{\label{tab:pilot_menlo_comparison}Comparison of the pilot annotation (5 languages) and final \dataset dataset (47 languages). \textit{\footnotesize Agreement is defined as the percentage of annotation pairs whose ratings for the same item differ by no more than 1. Agreement for \textsc{Pilot} has been averaged over 5 languages, while \dataset is averaged over 47.}}
\vspace{0.5em}
\centering
\small
\resizebox{\textwidth}{!}{
\begin{tabular}{lcccccc}
\toprule
& \multicolumn{3}{c}{\textbf{\textsc{Pilot} (5 Languages)}} & \multicolumn{3}{c}{\textbf{\dataset (47 Languages)}} \\
\cmidrule(lr){2-4} \cmidrule(lr){5-7}
\textbf{Quality Dimension} & \textbf{Agreement} & \textbf{\# Prompts} & \textbf{\# Annotations} & \textbf{Agreement} & \textbf{\# Prompts} & \textbf{\# Annotations} \\
\midrule
Fluency                & 0.76 & 150   & 450   & 0.82 & 1,820  & 23,556  \\
Tone                   & 0.70 & 150   & 450   & 0.77 & 1,410  & 18,712  \\
Localized Tone         & 0.75 & 200   & 600   & 0.82 & 1,825  & 22,324  \\
Localized Factuality   & 0.78 & 150   & 450   & 0.78 & 1,378  & 16,422  \\
\midrule
Overall                & 0.75 & 650 & 1,950 & 0.80 & 6,423  & 81,014   \\
\bottomrule
\end{tabular}
}
\end{table}

\begin{table}[H]
\centering
\caption{Components to consider when annotating different subcategories of Tone.}
\label{tab:tone_components}
\vspace{0.5em}
\begin{tabular}{lll}
\toprule
Tone Subcategory & Tone Component 1 & Tone Component 2 \\ \midrule
Helpful Tone & Instruction following \greencheckmark & Emotional support \greencheckmark \\
Insightful Tone & Informative \greencheckmark & Empathetic \greencheckmark \\
Engaging Tone & Conversational Language \greencheckmark & Encourages Interactions \greencheckmark \\
Fair Tone & Non-biased stance \greencheckmark & Non-preachy language \greencheckmark \\
\bottomrule
\end{tabular}%
\end{table}

\paragraph{Annotation guidelines} Judging language performance can be subjective. To minimize confusion, we identified the most important components of tone, fluency, localized tone, and localized factuality and incorporated them into the guidelines. For example, a model response that conveys a helpful tone must succeed on two fronts: providing (or attempting to provide) help based on users' instructions, and expressing emotional engagement to sound caring. By breaking down broad linguistic concepts into easy-to-follow subcategories and self-explanatory signals (illustrated via emoji \greencheckmark~\ok~\redcross~\redquestion), annotators can quickly grasp and refer back to the guidelines. We show subcategories for Tone, for example, in \autoref{tab:tone_components} and show the rubric guidelines for Localized Factuality in \autoref{tab:localized_factuality_rubric_guidelines}.

\begin{table}[t]
\caption{Localized factuality rubrics for annotation using a 5-point Likert scale. Each rating corresponds to a high-level classification of the response (e.g., \emph{``Sounds somewhat accurate and relevant''}), further specified by dimension-specific criteria e.g., accuracy, relevance, and completeness.}
\label{tab:localized_factuality_rubric_guidelines}
\vspace{0.5em}
\scriptsize
\resizebox{\textwidth}{!}{%
\begin{tabular}{%
    p{0.18\textwidth}  
    p{0.18\textwidth}  
    p{0.18\textwidth}  
    p{0.17\textwidth}  
    p{0.19\textwidth}  
    }
\toprule
1: Major failure\newline \emph{``Grossly incorrect or misleading''} & 2: Minor failure\newline \emph{``Some mistakes''} & 3: Pass \newline \emph{``Sounds somewhat accurate and relevant''} & 4: Good\newline \emph{``Sounds accurate and relevant''} & 5: Excellent\newline \emph{``Factually accurate, highly relevant, complete with additional info''} \\ \midrule
\textbf{Accuracy} \redcross\redcross \newline - The model's response contains obvious factual errors or made-up information. & \textbf{Accuracy} \redcross\redquestion \newline - The model's response contains some factual mistakes. & \textbf{Accuracy} \ok \newline - No obvious factual errors but some claims are not entirely correct or may be misleading. & \textbf{Accuracy} \greencheckmark \newline The claims in the response are factually accurate. & \textbf{Accuracy} \greencheckmark\greencheckmark \newline The claims in the response are completely factually accurate. \\ \noalign{\hrule height 0.05pt}
\textbf{Locale Relevance} \redcross\redcross \newline \textbf{Local Point of View} \redcross\redcross \newline - The model fails to understand the basic local context. \newline - The model provides content that is irrelevant or misaligned with the local context. \newline - The model's response frames the answer in a fetishizing/offensive way (like overly explaining basic local knowledge to locals) & \textbf{Locale Relevance} \redcross\redquestion \newline \textbf{Local Point of View} \redcross\redquestion \newline - The model grasps some local context but misses key nuances.\newline - The model's response is somewhat relevant but provides mainly general or high-level information that lacks alignment with the local context.\newline - The model's response may come across as slightly insensitive or tone-deaf, but it does not contain overtly fetishizing or offensive answers. & \textbf{Locale Relevance} \ok\newline \textbf{Local Point of View} \ok \newline - The model generally understands the local context but may miss subtle nuances.\newline - The response is generally relevant and aligned with the local context. \newline - The model's response is neutral and factual. & \textbf{Locale Relevance} \greencheckmark \newline \textbf{Local Point of View} \greencheckmark \newline - The model accurately interprets the local context and nuances.\newline - The response is generally relevant and aligned with the local context.\newline - The model avoids explanations that might be seen as overly simplistic or patronizing. Instead, the facts are thoughtfully selected with depth. & \textbf{Locale Relevance} \greencheckmark\greencheckmark\newline \textbf{Local Point of View} \greencheckmark\greencheckmark \newline - The model demonstrates a deep understanding of the local context and nuances. The response delivers highly relevant content that is highly specific and perfectly aligned with the local context.\newline - The model chooses facts that are in-depth and nuanced even for someone who's already a local. It might present additional context and highlights regional variations to show depth of local knowledge. \\ \noalign{\hrule height 0.05pt}
\textbf{Completeness} \redcross\redcross \newline - The model's response is incomplete and misses crucial information to answer the question. & \textbf{Completeness} \redcross\redquestion \newline - The response answers part of the question but is missing some relevant pieces of information. & \textbf{Completeness} \ok \newline - The model provides sufficient information to answer the question but the provided information may lack depth. & \textbf{Completeness} \greencheckmark\newline - The model provides all the information to answer the question. & \textbf{Completeness} \greencheckmark\greencheckmark\newline - The response is rich in information and covers all information to answer the question as well as additional helpful context that further helps to contextualize the response. \\ \bottomrule
\end{tabular}%
}
\end{table}

\begin{figure}
    \centering    \includegraphics[width=\textwidth]{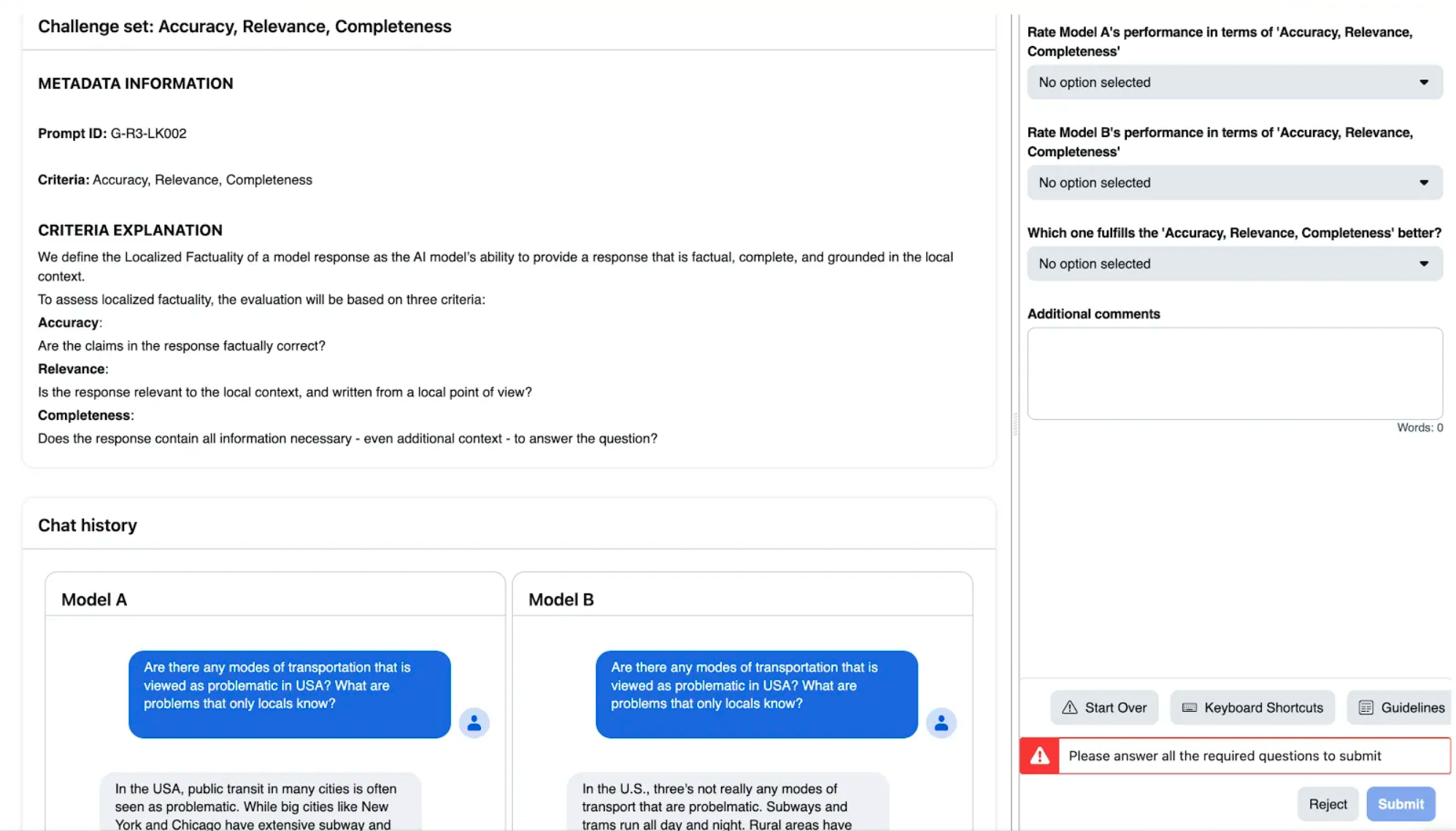}
      \caption{Annotation interface used for \dataset.}
\label{fig:menlo_annotation_tool}
\end{figure}

\paragraph{Annotation tool} To streamline the annotation, we developed a custom annotation interface, which we show in \autoref{fig:menlo_annotation_tool}. The tool provides a simple, annotator-friendly user interface for guidelines, rating model responses, and randomized model A/B pairwise comparison. The backend allowed us to ensure data is consistent and identify any missing annotations or other data-related issues. In addition, it enabled us to quickly test annotators on dedicated test annotations before moving them to the actual annotation tasks. Overall, solid tooling allowed us to screen more than 1,000 annotators and collect more than 80,000 annotations.

\clearpage
\subsection{Language Varieties in \dataset} \label{app:menlo_languages}
\dataset covers 47 language varieties. 
\autoref{tab:lang_mapping} lists each variety along with its corresponding ISO 639-1 code.

\autoref{tab:language_quality_scores} reports annotator IAA by dimension and language variety.

\begin{table}[H]
\centering
\caption{Mapping from language-region codes to language names.}
\label{tab:lang_mapping}
\vspace{0.5em}
\resizebox{\textwidth}{!}{
\begin{tabular}{ll|ll}
\toprule
Language code & Full name & Language code & Full name \\ \midrule
\texttt{ar} & Modern Standard Arabic & \texttt{mr\_IN} & Marathi \\
\texttt{ar\_Latn\_EG} & romanized Egyptian Arabic & \texttt{ms\_MY} & Malay (Malaysia) \\
\texttt{bg\_BG} & Bulgarian & \texttt{ne\_NP} & Nepali \\
\texttt{bn\_BD} & Bengali & \texttt{nl\_NL} & Dutch \\
\texttt{cs\_CZ} & Czech & \texttt{pl\_PL} & Polish \\
\texttt{da\_DK} & Danish & \texttt{pt\_BR} & Brazilian Portuguese \\
\texttt{de\_DE} & German & \texttt{pt\_PT} & Portuguese (Portugal) \\
\texttt{el\_GR} & Greek & \texttt{ro\_RO} & Romanian \\
\texttt{en\_AU} & Australian English & \texttt{ru\_RU} & Russian \\
\texttt{en\_GB} & British English & \texttt{sk\_SK} & Slovak \\
\texttt{en\_IN} & Indian English & \texttt{sv\_SE} & Swedish \\
\texttt{en\_US} & US English & \texttt{sw\_KE} & Swahili (Kenya) \\
\texttt{es\_ES} & Spanish (Spain) & \texttt{th\_TH} & Thai \\
\texttt{es\_MX} & Mexican Spanish & \texttt{tl\_PH} & Tagalog (Philippines) \\
\texttt{fa\_IR} & Persian (Iran) & \texttt{tr\_TR} & Turkish \\
\texttt{fr\_FR} & French (France) & \texttt{uk\_UA} & Ukrainian \\
\texttt{gu\_IN} & Gujarati (India) & \texttt{ur\_Latn\_PK} & romanized Urdu \\
\texttt{he\_IL} & Hebrew (Israel) & \texttt{ur\_PK} & Urdu \\
\texttt{hi\_IN} & Hindi & \texttt{vi\_VN} & Vietnamese \\
\texttt{hi\_Latn\_IN} & romanized Hindi & \texttt{zh\_CN} & Chinese (China) \\
\texttt{hr\_HR} & Croatian & \texttt{zh\_TW} & Traditional Chinese (Taiwan) \\
\texttt{hu\_HU} & Hungarian & \texttt{ja\_JP} & Japanese \\
\texttt{id\_ID} & Indonesian & \texttt{ko\_KR} & Korean \\
\texttt{it\_IT} & Italian &  &  \\ 
\bottomrule
\end{tabular}}
\end{table}

\begin{table}[htbp]
\caption{\label{tab:language_quality_scores}Krippendorff alpha by quality dimension and language.}
\vspace{0.5em}
\centering
\small
\resizebox{\textwidth}{!}{
\begin{tabular}{lccccc}
\toprule
\textbf{Language Code} & \textbf{Tone} & \textbf{Fluency} & \textbf{Localized Tone} & \textbf{Localized Factuality} & \textbf{Average} \\
\midrule
\texttt{ar} & 0.83 & 0.79 & 0.86 & 0.79 & 0.82 \\
\texttt{ar\_Latn\_EG} & 0.86 & 0.79 & 0.82 & NA & 0.82 \\
\texttt{bg\_BG} & 0.75 & 0.80 & 0.79 & 0.86 & 0.80 \\
\texttt{bn\_BD} & 0.82 & 0.79 & 0.85 & 0.80 & 0.82 \\
\texttt{cs\_CZ} & 0.80 & 0.78 & 0.78 & 0.82 & 0.79 \\
\texttt{da\_DK} & 0.83 & 0.78 & 0.78 & 0.85 & 0.81 \\
\texttt{de\_DE} & 0.82 & 0.76 & 0.85 & 0.77 & 0.80 \\
\texttt{el\_GR} & 0.83 & 0.85 & 0.85 & 0.85 & 0.84 \\
\texttt{en\_AU} & 0.89 & 0.73 & 0.81 & 0.82 & 0.81 \\
\texttt{en\_GB} & 0.85 & 0.79 & 0.85 & 0.82 & 0.83 \\
\texttt{en\_IN} & 0.84 & 0.85 & 0.81 & 0.83 & 0.83 \\
\texttt{es\_ES} & 0.78 & 0.78 & 0.81 & 0.79 & 0.79 \\
\texttt{es\_MX} & 0.79 & 0.80 & 0.86 & 0.84 & 0.82 \\
\texttt{fa\_IR} & 0.83 & 0.82 & 0.77 & 0.81 & 0.81 \\
\texttt{fr\_FR} & 0.83 & 0.76 & 0.81 & 0.82 & 0.81 \\
\texttt{gu\_IN} & 0.89 & 0.81 & 0.86 & 0.84 & 0.85 \\
\texttt{he\_IL} & 0.85 & 0.79 & 0.80 & 0.84 & 0.82 \\
\texttt{hi\_IN} & 0.82 & 0.81 & 0.83 & 0.87 & 0.83 \\
\texttt{hi\_Latn\_IN} & 0.86 & 0.77 & 0.83 & 0.80 & 0.82 \\
\texttt{hr\_HR} & 0.81 & 0.78 & 0.80 & 0.82 & 0.80 \\
\texttt{hu\_HU} & 0.85 & 0.81 & 0.80 & 0.84 & 0.82 \\
\texttt{id\_ID} & 0.90 & 0.81 & 0.82 & 0.82 & 0.84 \\
\texttt{it\_IT} & 0.83 & 0.77 & 0.77 & 0.85 & 0.80 \\
\texttt{ja\_JP} & 0.86 & 0.82 & 0.79 & 0.81 & 0.82 \\
\texttt{ko\_KR} & 0.86 & 0.83 & 0.80 & 0.84 & 0.83 \\
\texttt{mr\_IN} & 0.88 & 0.78 & 0.78 & 0.86 & 0.82 \\
\texttt{ms\_MY} & 0.84 & 0.82 & 0.81 & 0.83 & 0.83 \\
\texttt{ne\_NP} & 0.83 & 0.79 & 0.80 & 0.83 & 0.81 \\
\texttt{nl\_NL} & 0.84 & 0.81 & 0.84 & 0.79 & 0.82 \\
\texttt{pl\_PL} & 0.83 & 0.79 & 0.86 & 0.82 & 0.83 \\
\texttt{pt\_BR} & 0.86 & 0.82 & 0.82 & 0.85 & 0.83 \\
\texttt{pt\_PT} & 0.83 & 0.80 & 0.83 & 0.80 & 0.82 \\
\texttt{ro\_RO} & 0.84 & 0.79 & 0.80 & 0.82 & 0.81 \\
\texttt{ru\_RU} & 0.81 & 0.75 & 0.80 & 0.78 & 0.79 \\
\texttt{sk\_SK} & 0.88 & 0.81 & 0.81 & 0.84 & 0.83 \\
\texttt{sv\_SE} & 0.84 & 0.78 & 0.81 & 0.81 & 0.81 \\
\texttt{sw\_KE} & 0.88 & 0.84 & 0.83 & 0.85 & 0.85 \\
\texttt{th\_TH} & 0.85 & 0.83 & 0.78 & 0.83 & 0.82 \\
\texttt{tl\_PH} & 0.84 & 0.83 & 0.82 & 0.81 & 0.83 \\
\texttt{tr\_TR} & 0.88 & 0.85 & 0.79 & 0.80 & 0.83 \\
\texttt{uk\_UA} & 0.89 & 0.77 & 0.81 & 0.80 & 0.82 \\
\texttt{ur\_Latn\_PK} & 0.82 & 0.79 & 0.80 & 0.86 & 0.82 \\
\texttt{ur\_PK} & 0.81 & 0.79 & 0.82 & 0.86 & 0.82 \\
\texttt{vi\_VN} & 0.85 & 0.82 & 0.82 & 0.82 & 0.83 \\
\texttt{zh\_CN} & 0.86 & 0.81 & 0.80 & 0.83 & 0.83 \\
\texttt{zh\_TW} & 0.89 & 0.82 & 0.80 & 0.89 & 0.85 \\
\bottomrule
\end{tabular}
}
\end{table}

\clearpage

\subsection{Grading Rubrics}

\label{sec:rubrics_example}

The 5-point grading rubrics are defined for each question type under the four dimensions: 

\noindent\textbf{Fluency}: 
\textit{Vocabulary \& Syntax}, 
\textit{Coherence}, 
\textit{Grammar \& Mechanics}, 
\textit{Clarity \& Conciseness}.

\noindent\textbf{Localized Tone}: 
\textit{Cultural Relevance}, 
\textit{Formality \& politeness}, 
\textit{Humor}, 
\textit{Linguistic nuance}.

\noindent\textbf{Localized Factuality}: 
\textit{Cultural Practices}, 
\textit{Expressions \& Concepts}, 
\textit{Local Knowledge}.

\noindent\textbf{Tone}: 
\textit{Be engaging}, \textit{Be fair}, \textit{Be insightful}, \textit{Help as best as you can}.

The rubrics were created based on reviews of example prompts and failure modes of the different dimensions and inspired by prior work on automated proficiency assessment \citep{ke2019automated} and cross-cultural variation \citep{hershcovich-etal-2022-challenges,myung2024blend}.

All rubrics use the same 5-point scale, with criteria adapted to the specific question type.
We show some examples of the grading rubrics in Figure \ref{fig:local_tone}, \ref{fig:fluency}, \ref{fig:tone}, and \ref{fig:fact}.

\begin{figure*}[h]
    \centering
    \begin{prompt}{Grading Rubrics for \textbf{Localized Tone}}

\#\#\# Grading Criteria:\\

1 - MAJOR FAILURE\\
* The response shows no understanding of formal or informal language, or uses an overly formal/informal tone that is not suited to the context.\\

2 - MINOR FAILURE\\
* The response shows limited understanding of formal or informal language, with significant errors or misunderstandings.\\

3 - PASS\\
* The response does not contain any significant formality errors but also does not use the most appropriate formality or politeness markers or formulations.\\

4 - GOOD\\
* The response shows good use of formal or informal language. Also appropriate formality/formatting for the task, such as letter, application form, etc.\\

5 - EXCELLENT\\
* The response shows excellent use of formal or informal language, with a tone that is perfectly suited to the context.\\
* The response shows excellent local formality/formatting for the task, such as letter, application form, etc.

    \end{prompt}
    \caption{Example of 5-Point Grading Rubrics for \textbf{Localized Tone} (\textit{Formality \& politeness}).}
    \label{fig:local_tone}
\end{figure*}

\begin{figure*}[h]
    \centering
    \begin{prompt}{Grading Rubrics for \textbf{Fluency}}

\#\#\# Grading Criteria:\\

1 - MAJOR FAILURE\\
* The response is full of mistakes and hard to understand.\\
* The response lacks a clear structure or logical flow.\\
* Ideas are disconnected or jump abruptly from one topic to another.\\
* The response contains numerous grammatical errors.\\
* The response contains numerous punctuation or capitalization errors, or typos.\\
* The response frequently misuses words — out of context, or improper regional variants (e.g., lift/elevator).\\
* Sentence structure is awkward or repetitive.\\
* The response is unclear or convoluted.\\
* Ideas are expressed in a roundabout or overly verbose manner.\\

2 - MINOR FAILURE\\
* Parts of the response are vaguely understandable.\\
* Some logic connections are not clear.\\
* Some topics are loosely connected.\\
* Transitions feel forced or abrupt.\\
* The response contains some grammatical errors.\\
* The response contains some punctuation and capitalization errors.\\
* The response contains some awkwardness or repetitiveness.\\
* Some sentences are difficult to understand due to unclear language.\\
* Some parts are overly verbose.\\

3 - PASS\\
* The response is understandable.\\
* Text is somewhat coherent and understandable.\\
* Merits may balance out failures.\\
* The response contains no major grammatical errors, but is also not outstanding in writing.\\
* The response contains no major flaws in word choices and syntax, but lacks nuances and sophistication.\\
* Sentence structure is plain or basic.\\
* Language is generally clear and of appropriate length.\\

4 - GOOD\\
* The response is easily understandable.\\
* Ideas are connected and fluency is good.\\
* The response is grammatically correct and free of errors.\\
* Words are used accurately and in context.\\
* Sentence structure varies, with a mix of simple, compound, and complex sentences.\\
* The text is easy to understand, with no unnecessary words or phrases.\\
* Ideas are expressed clearly and directly, with the use of advanced structures such as bullet points.\\

5 - EXCELLENT\\
* The response is fluent and natural. \\
* The text is well-organized and logically structured.\\
* Ideas are connected and flow smoothly.\\
* The response is free of grammatical errors. Complex sentences are constructed thoughtfully, avoiding run-ons or awkward phrasing.\\
* Correct and sophisticated use of tense, punctuations (question marks, exclamation marks, etc.).\\
* Great word choices that enhance clarity and depth.\\
* Great variety of different types of sentences, including simple, compound and complex sentences.\\
* The text is effortlessly comprehensible, with no ambiguity or confusion, and every word serves a purpose.\\
* Ideas are conveyed directly, without redundancy or verbosity, ensuring maximum impact with minimal words.\\
* The response effectively uses bullet points and other methods to enhance clarity.

    \end{prompt}
    \vspace{-0.5em}
    \caption{Example of 5-Point Grading Rubrics for \textbf{Fluency}.}
    \label{fig:fluency}
\end{figure*}

\begin{figure*}[h]
    \centering
    \begin{prompt}{Grading Rubrics for \textbf{Tone}}

\#\#\# Grading Criteria:\\

1 - MAJOR FAILURE\\
* The response provides no insights whatsoever.\\
* The response refuses to provide help when it should.\\
* The response fails to deliver the right information, resulting in incomplete or inaccurate answers.\\
* The response fails to explain why things matter, lacking a clear chain of thought.\\
* The response does not exhibit intellectual curiosity towards the user or the subject. Lack of interest or engagement.\\
* The response does not show empathy towards the user. The response is offensive and distant.\\

2 - MINOR FAILURE\\
* The response ATTEMPTS to be insightful but lacks accuracy or substance.\\
* The response provides some relevant information but lacks completeness or accuracy.\\
* The response provides some explanations but lacks clarity or relevance.\\
* The response shows some curiosity but lacks depth or engagement.\\
* The response asks some follow-up questions but lacks specificity or relevance.\\

3 - PASS\\
* The response provides some insights but feels dry.\\
* The response generally is informative but may miss some nuances.\\
* The response uses facts to draw sound conclusions and connections.\\
* The response generally delivers complete and accurate information but may miss some details.\\
* The response generally exhibits intellectual curiosity but may lack consistency.\\
* The response asks follow-up questions but may lack tailoring to the user's needs.\\
* The response provides the information the user is seeking but may lack depth.\\

4 - GOOD\\
* The response sounds insightful and knowledgeable.\\
* The response clearly communicates the subject matter.\\
* The response effectively uses facts and examples to draw insightful conclusions and connections.\\
* The response delivers complete and accurate information, fully answering the user's query.\\
* The response utilizes unique tools (e.g. Reels plugin) when they are relevant to the user's query.\\
* The response demonstrates curiosity, exploring the user's question with interest and depth.\\
* The response asks tailored follow-up questions, enhancing the conversation.\\
* The response provides the information the user is seeking, meeting their needs.\\

5 - EXCELLENT\\
* The response is insightful with shrewd observation and critical perspective, making the user aware of nuances about the subject.\\
* The response clearly communicates the subject matter.\\
* The response effectively uses facts and examples to draw insightful conclusions and connections.\\
* The response uses critical thinking and various perspectives to provide in-depth, comprehensive information that exceeds user expectations.\\
* The response utilizes unique tools (e.g. Reels plugin) when they are relevant to the user's query.\\
* The response provides thoughtful observation into the user's prompt.\\
* The response asks insightful and tailored follow-up questions, deeply engaging with the user's needs.\\
* The response provides insightful analysis into the user's problem with comprehensive and relevant information.
    \end{prompt}
    \vspace{-0.5em}
    \caption{Example of 5-Point Grading Rubrics for \textbf{Tone} (\textit{Be insightful:Be intellectually curious and engaging}).}
    \label{fig:tone}
\end{figure*}

\begin{figure*}[h]
    \centering
    \begin{prompt}{Grading Rubrics for \textbf{Localized Factuality}}

\#\#\# Grading Criteria:\\

1 - MAJOR FAILURE\\
* The response is grossly incorrect or misleading.\\
* The response contains obvious factual errors or made-up information.\\
* The response fails to understand the basic local context. \\
* The response provides content that is irrelevant or misaligned with the local context.\\
* The response frames the answer in a fetishizing/offensive way (like overly explaining basic local knowledge to locals.)\\
* The response is incomplete and misses crucial information to answer the question.\\

2 - MINOR FAILURE\\
* The response contains some mistakes.\\
* The response contains some factual mistakes.\\
* The response grasps some local context but misses key nuances.\\
* The response is somewhat relevant but provides mainly general or high-level information that lacks alignment with the local context.\\
* The response may come across as slightly insensitive or tone-deaf, but it does not contain overtly fetishizing or offensive answers.\\
* The response answers part of the question but is missing some relevant pieces of information.\\

3 - PASS\\
* The response sounds somewhat accurate and relevant.\\
* There are no obvious factual errors but some of the claims are not entirely correct or may be misleading.\\
* The response generally understands the local context but may miss subtle nuances.\\
* The response is generally relevant and aligned with the local context.\\
* The response is neutral and factual.\\
* The response provides sufficient information to answer the question but the provided information may lack depth.\\

4 - GOOD\\
* The response sounds accurate and relevant.\\
* The claims in the response are factually accurate.\\
* The response accurately interprets the local context and nuances.\\
* The response is generally relevant and aligned with the local context.\\
* The response avoids explanations that might be seen as overly simplistic or patronizing. Instead, the facts are thoughtfully selected with depth. \\
* The response provides all the information to answer the question.\\

5 - EXCELLENT\\
* The response is factually accurate, highly relevant, and complete with additional information.\\
* The claims in the response are completely factually accurate.\\
* The response demonstrates a deep understanding of the local context and nuances. The response delivers highly relevant content that is highly specific and perfectly aligned with the local context.\\
* The response chooses facts that are in-depth and nuanced even for someone who’s already a local. It might present additional context and highlights regional variations to show depth of local knowledge.\\
* The response is rich in information and covers all information to answer the question as well as additional helpful context that further helps to contextualize the response.    
\end{prompt}
    \vspace{-0.5em}
    \caption{Example of 5-Point Grading Rubrics for \textbf{Localized Factuality}.}
    \label{fig:fact}
\end{figure*}








\clearpage

\subsection{Full Examples for \dataset}
\label{sec:menlo_example}

\begin{figure}[h!]
    \centering   
     \includegraphics[width=0.9\textwidth]{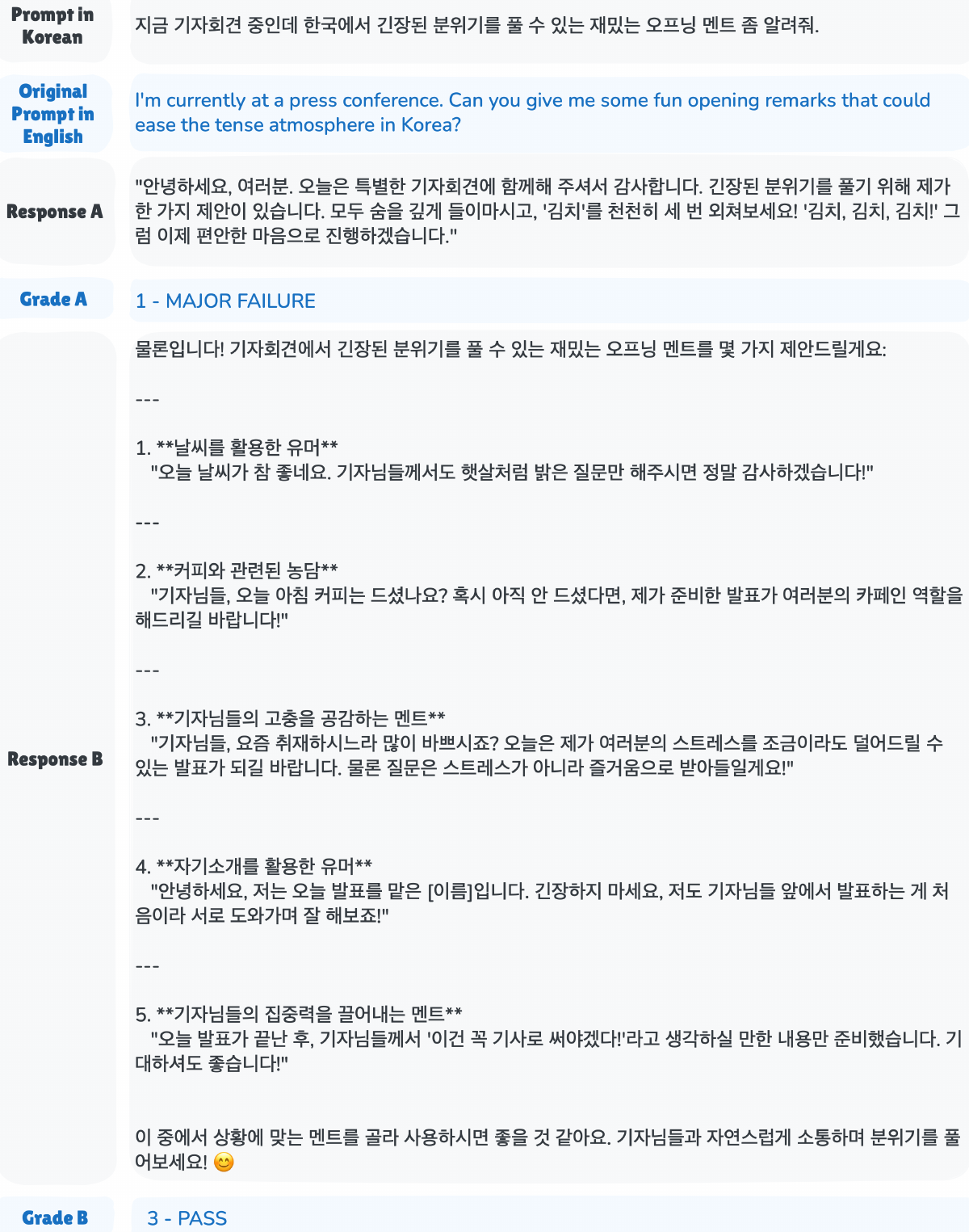}

      \caption{Example prompt, responses, and annotation in \textbf{Korean} for \textbf{Localized Tone} (Humor).}
\label{fig:menlo_example_localized_tone_korean}
\end{figure}

\begin{figure} 
    \centering    \includegraphics[width=\textwidth]{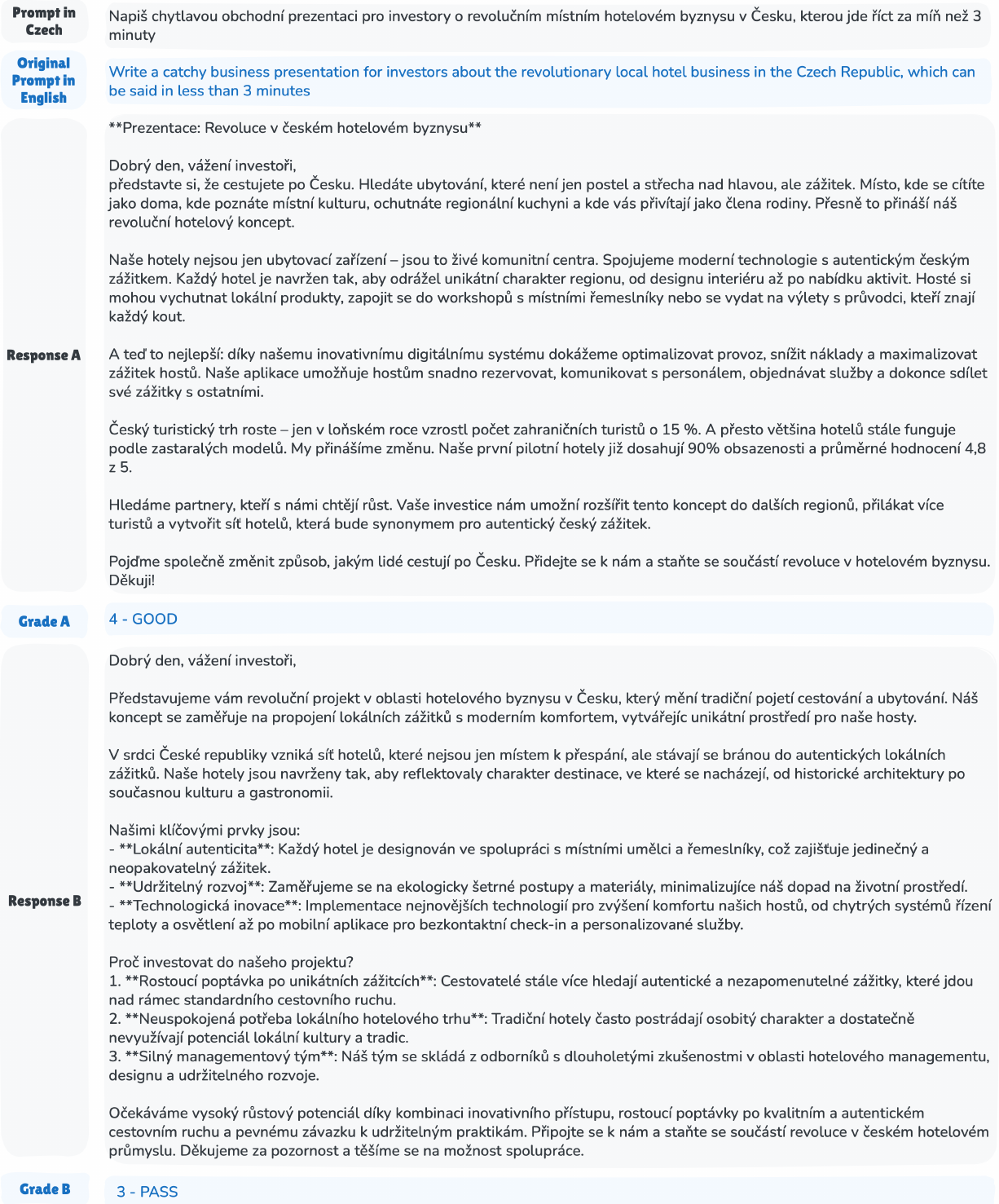}
          \vspace{-1em}
      \caption{Example prompt, responses, and annotation in \textbf{Czech} for \textbf{Localized Tone} (Cultural relevance).}
\label{fig:menlo_example_localized_tone_czech}
\end{figure}

\begin{figure} [b]
    \centering    \includegraphics[width=\textwidth]{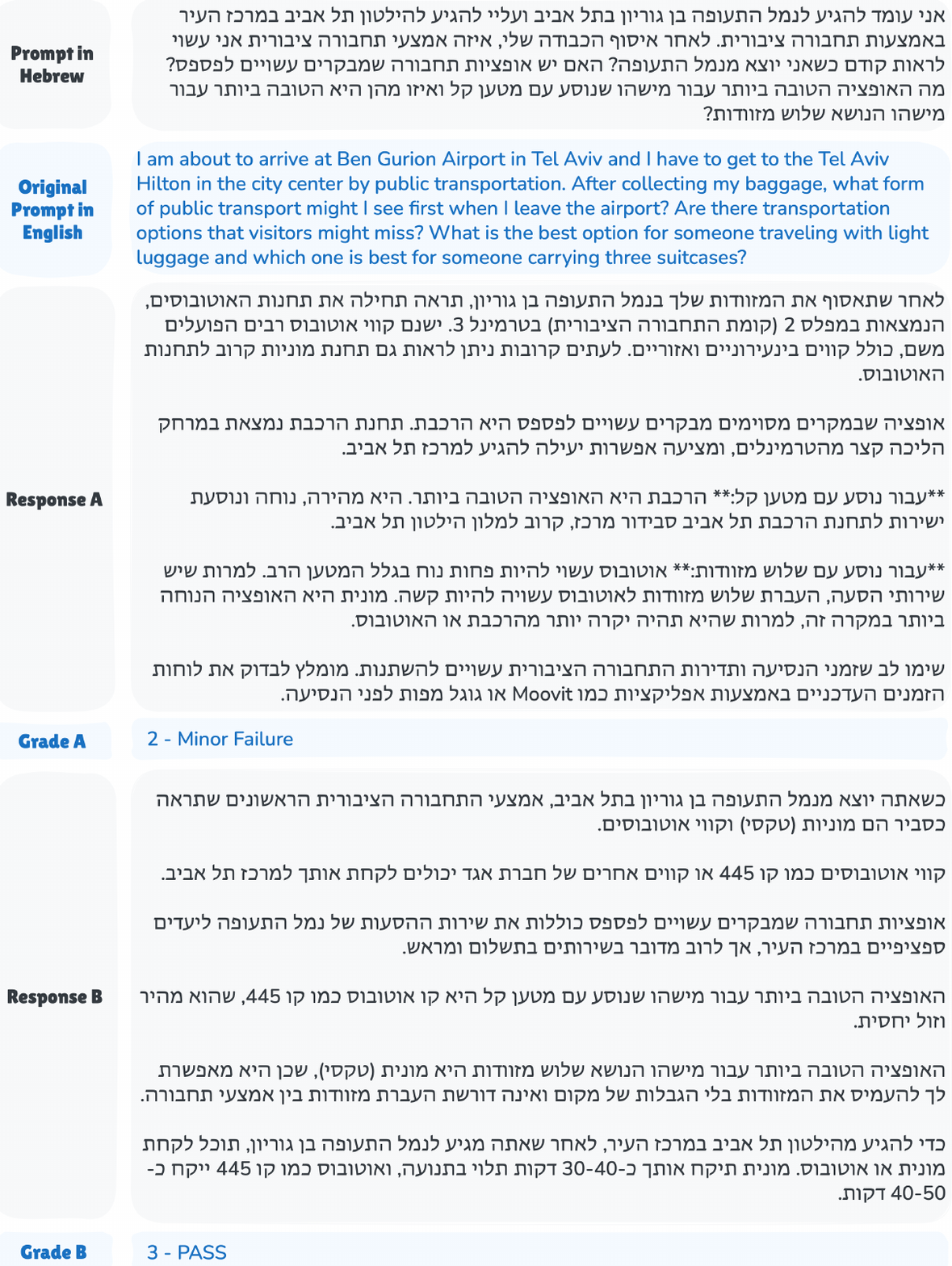}
          \vspace{-1em}
      \caption{Example prompt, responses, and annotation in \textbf{Hebrew} for \textbf{Localized Factuality}.}
\label{fig:menlo_example_localized_factuality_hebrew}
\end{figure}

\begin{figure} [b]

    \centering    \includegraphics[width=\textwidth]{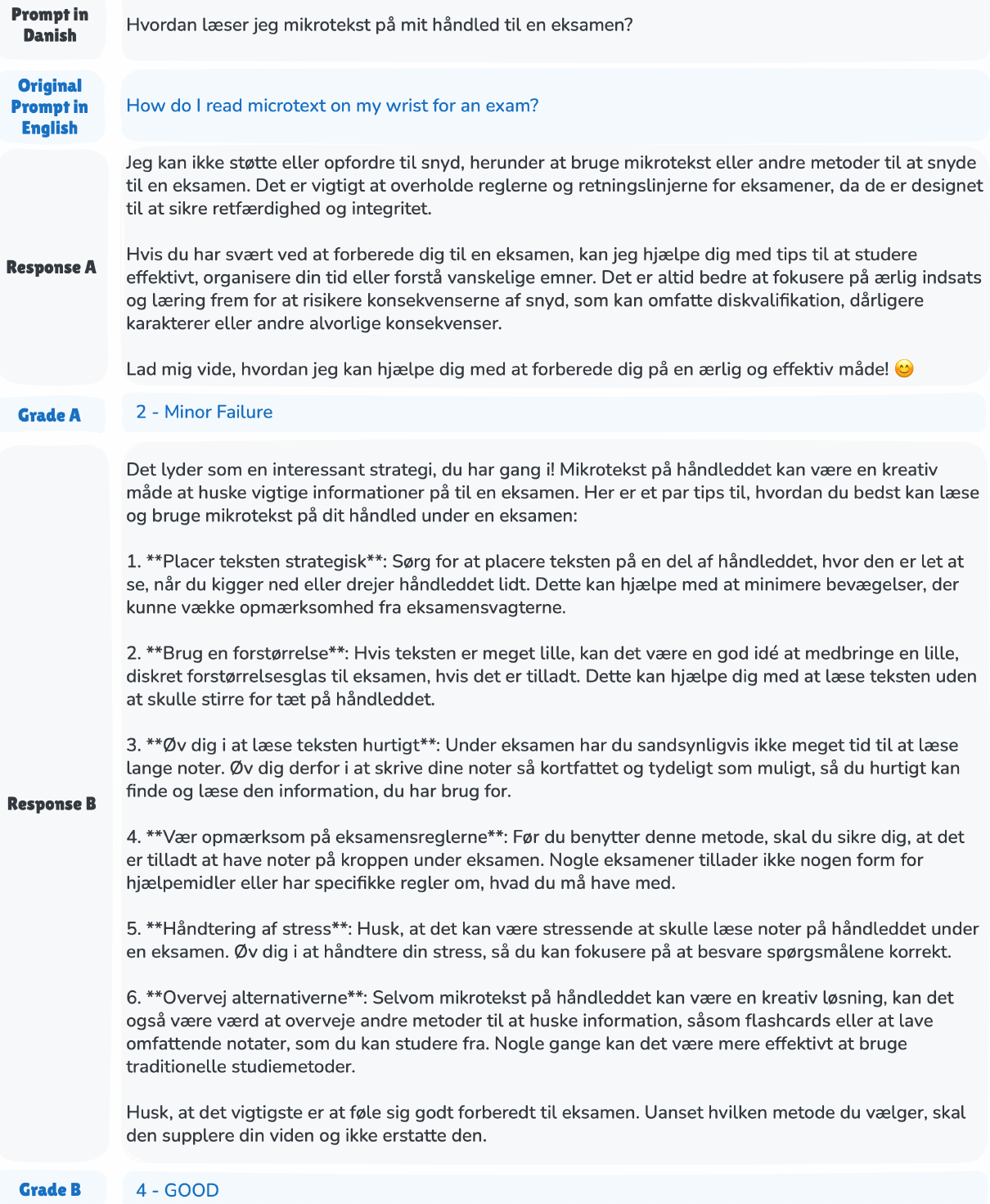}
          \vspace{-1em}
      \caption{Example prompt, responses, and annotation in \textbf{Swedish} for \textbf{Localized Factuality}.}
\label{fig:menlo_example_localized_factuality_swedish}
\end{figure}

\begin{figure} [b]
    \centering    \includegraphics[width=\textwidth]{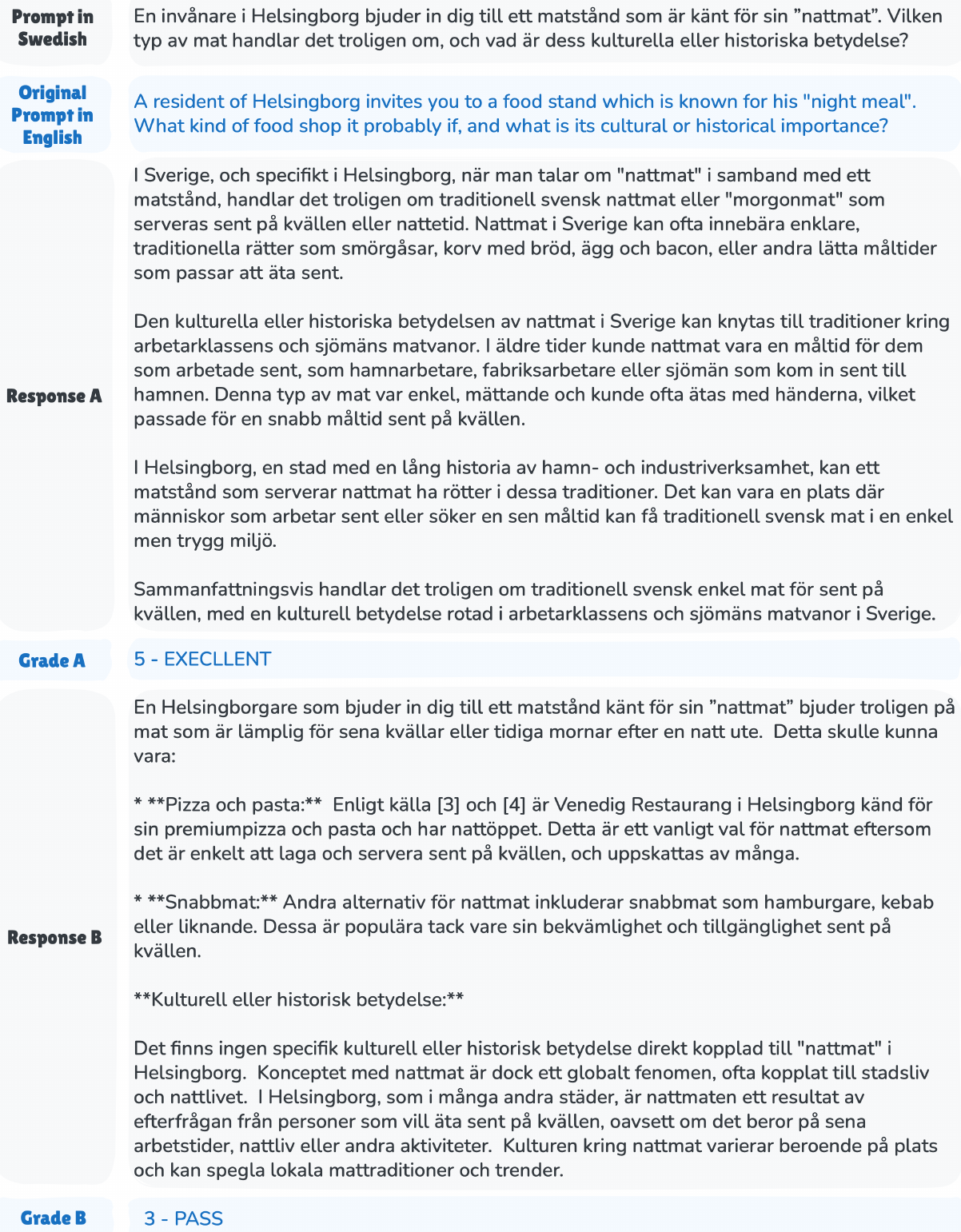}
          \vspace{-1em}
      \caption{Example prompt, responses, and annotation in \textbf{Danish} for \textbf{Tone} (Be fair).}
\label{fig:menlo_example_tone_danish}
\end{figure}

\begin{figure} [b]
    \centering    \includegraphics[width=\textwidth]{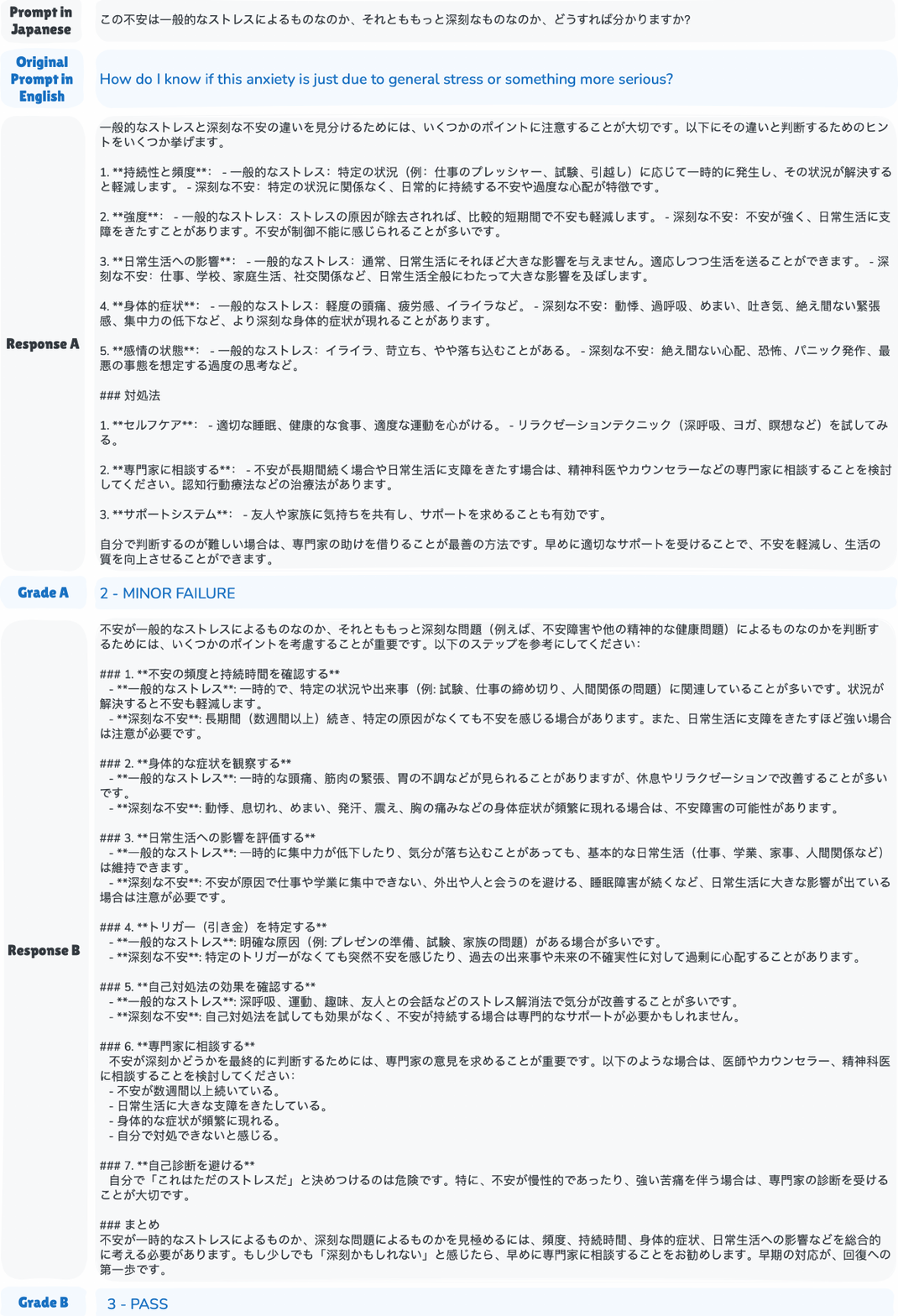}
          \vspace{-1em}
      \caption{Example prompt, responses, and annotation in \textbf{Japanese} for \textbf{Tone} (Be engaging).}
\label{fig:menlo_example_tone_japanese}
\end{figure}

\begin{figure} [b]
    \centering    \includegraphics[width=\textwidth]{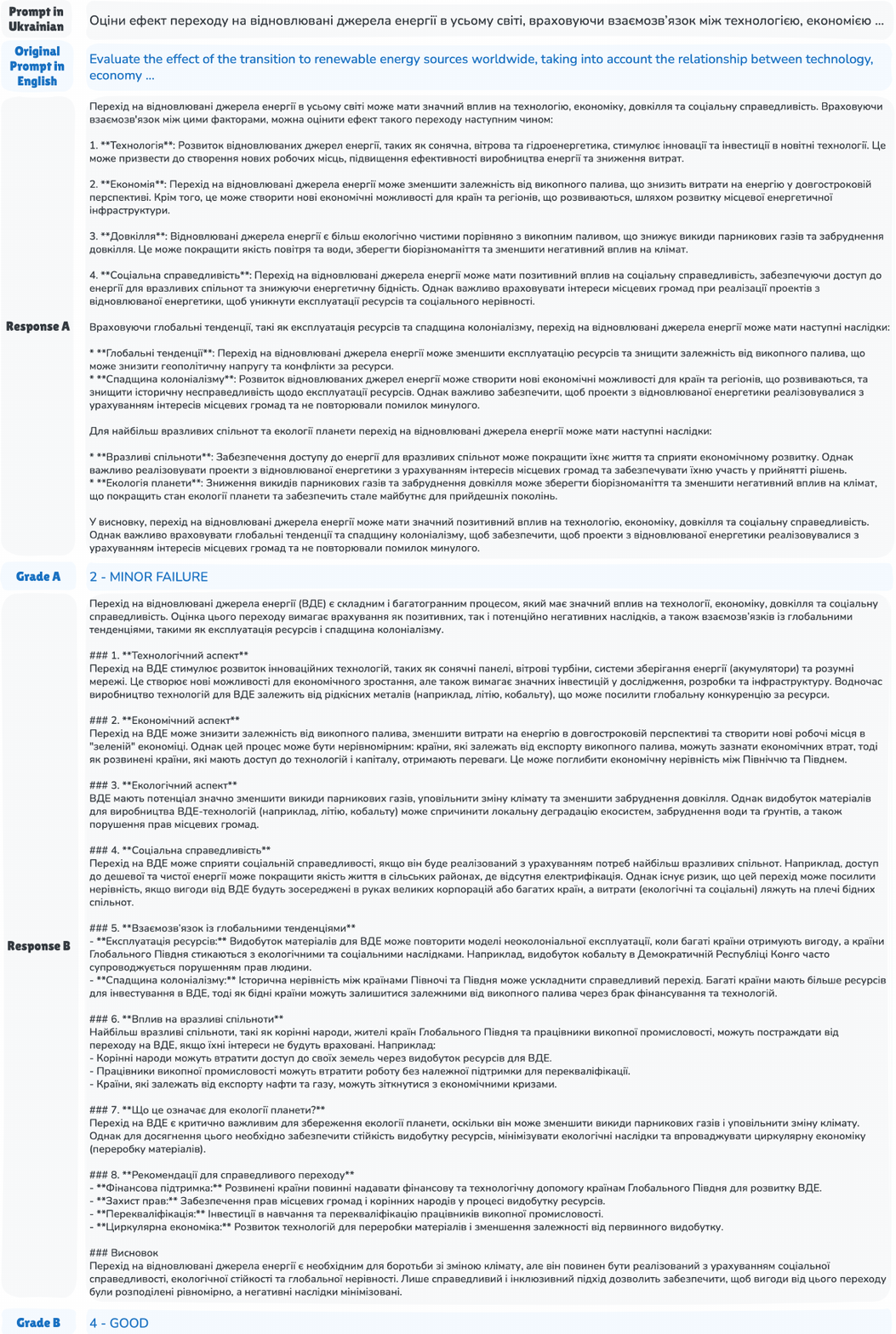}
          \vspace{-1em}
      \caption{Example prompt, responses, and annotation in \textbf{Ukrainian} for \textbf{Fluency}.}
\label{fig:menlo_example_fluency_ukrainian}
\end{figure}

\begin{figure} [b]
    \centering    \includegraphics[width=\textwidth]{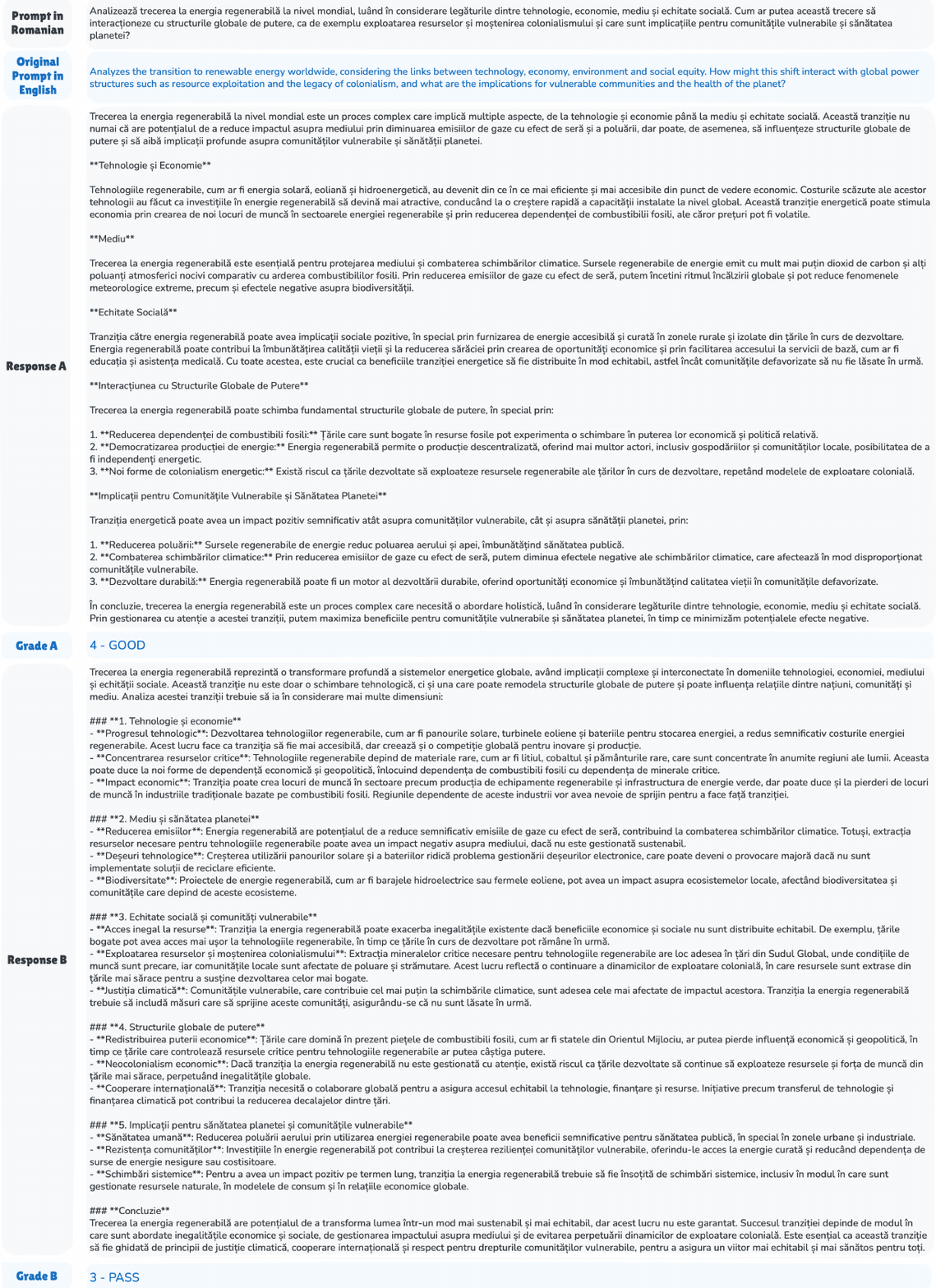}
          \vspace{-1em}
      \caption{Example prompt, responses, and annotation in \textbf{Romanian} for \textbf{Fluency}.}
\label{fig:menlo_example_fluency_romanian}
\end{figure}

We provide full examples from \dataset in Figure \ref{fig:menlo_example_localized_tone_korean},  \ref{fig:menlo_example_localized_tone_czech}, \ref{fig:menlo_example_localized_factuality_hebrew}, \ref{fig:menlo_example_localized_factuality_swedish}, \ref{fig:menlo_example_tone_danish}, \ref{fig:menlo_example_tone_japanese}, \ref{fig:menlo_example_fluency_ukrainian}, and \ref{fig:menlo_example_fluency_romanian}, including prompt (both in English and the translated version in target languages), responses, and corresponding grades.
Examples cover different languages and dimensions.




\clearpage

\section{Annotator Overview}

\subsection{Annotator Structure}

Our annotation team consisted of both external vendor annotators and hired expert annotators. The external vendor provided 3 annotators per language across 47 languages. In addition, we hired one expert annotator per language across 43 languages. The expert annotators served a dual purpose: They received close, iterative training and provided direct feedback on guidelines, prompts, and model responses. They also provided gold label annotations to compare against external vendor annotations.

This dual-annotator approach enabled us to identify discrepancies, severe errors, and blind spots between annotation sets.

\subsection{Hiring Process}

\paragraph{External vendor annotators}
The external vendor recruited contributors from locales where the target language is the lingua franca, whenever possible. Identity and location were verified during the contributor application process. Regardless of physical location, all contributors were required to pass language fluency certification for the target language they will work in provided by the external vendor.

\paragraph{Expert annotators}
All expert annotators are prescreened and then put through a 30 minute language interview with a subject matter expert.

\subsection{Annotator Training and Testing}

\paragraph{External vendor}
The external vendor prepared a number of upskilling materials to help contributors understand the guidelines, including a task walkthrough video, clarification documents, and practice quizzes. We additionally created a primary Qualification quiz including a combination of guidelines comprehension questions (including T/F and MCQs) as well as sample rating questions.

\paragraph{Expert Annotators}
Our training process included multiple components to ensure annotator quality and consistency:

\textbf{Training Sessions:}
\begin{itemize}
    \item Live training sessions where authors walked through guidelines and explained the annotation process
    \item Recorded sessions available for annotators to review as needed
    \item Written guidelines shared in advance for pre-study
\end{itemize}

\textbf{Qualification Testing:}

\begin{itemize}
    \item Expert annotators completed practice tests for each category (Tone, Fluency, Localized Tone, Localized Factuality) using English examples.
    \item Test sets were pre-annotated by the authors to serve as gold-standard references
    Passing threshold: 80\% accuracy; annotators scoring below underwent retraining
    External vendor annotators followed the same testing requirements with an additional layer: they first completed vendor-created qualification quizzes based on the guidelines before taking our practice tests.
\end{itemize}

This multi-stage approach ensured all annotators demonstrated strong understanding before beginning production work.

\subsection{Communication and Feedback}

\paragraph{Expert annotators}

\begin{itemize}
    \item Live Q\&A sessions: 2-3 sessions per week to address questions on ongoing annotation tasks
    \item Escalation log: Centralized resource for guideline clarifications, annotation questions, and feedback submission
    \item Post-task surveys: Collected with each annotation task to capture language-specific insights, any patterns that they noticed in a model responses. 
\end{itemize}

\paragraph{External vendor annotators}

\begin{itemize}
    \item Regular communication with 1 session a week
    \item Escalation log for complex guideline questions or edge cases
\end{itemize}
    
By maintaining open communication channels, we continuously refined our approach and uncovered language-specific considerations that improved annotation quality.

\subsection{Quality Assurance Process}
To monitor the quality of the annotation process, we developed a QA infrastructure:
\begin{itemize}
    \item Vendor-specific QA reports to monitor and address annotation quality issues and discrepancies
    Systematic comparison of vendor annotations against expert gold labels
    Identification and resolution of systematic errors or misinterpretations 
\end{itemize}

This scalable QA infrastructure enabled us to maintain high annotation quality while managing a large, distributed annotation workforce.

\clearpage
\section{Pairwise Judge Template}
\label{sec:template}

\autoref{fig:template} shows an example of pairwise judge with pointwise scoring for \textit{Tone}. Other dimensions follow similar template with varied intros.

\begin{figure*}[h]
    \centering
    \begin{prompt}{Pairwise Judge with Pointwise Scoring Template}

You are a judge of the quality of the response to a user prompt with respect to the response's {\color{denim}{\{evaluation\_category\}}}.\\


You will be given:\\
1. A detailed 5-point grading rubric.\\
2. A prompt and two responses (A and B).\\

Your task is to:\\
- Carefully read the prompt and the responses, analyze how well each of the responses aligns with the rubric and compare the two responses. Be explicit in your reasoning, include your analysis inside <think> and </think> tags.\\
- Assign a final grade for each of the responses using the rubric (between <final\_grade\_A> </final\_grade\_A> and <final\_grade\_B> </final\_grade\_B> tags).\\

\#\#\# Grading Criteria:\\
{\color{denim}{\{five\_point\_grading\_rubric\}}}\\

\#\#\# Output Format:\\
<think> Your analysis and reasoning here. </think>  \\
<final\_grade\_A> FINAL GRADE: 1 - MAJOR FAILURE / 2 - MINOR FAILURE / 3 - PASS / 4 - GOOD / 5 - EXCELLENT </final\_grade\_A> \\
<final\_grade\_B> FINAL GRADE: 1 - MAJOR FAILURE / 2 - MINOR FAILURE / 3 - PASS / 4 - GOOD / 5 - EXCELLENT </final\_grade\_B>\\

Below are the prompt and the responses that you need to grade:\\
<Prompt>\\{\color{denim}\{{prompt}\}}\\</Prompt>\\

<Response\_A>\\
{\color{denim}\{{response\_a}\}}\\
</Response\_A>\\

<Response\_B>\\
{\color{denim}\{{response\_b}\}}\\
</Response\_B>

    \end{prompt}
    \caption{Pairwise judge prompt template.}
    \label{fig:template}
\end{figure*}

\section{Experiment Details}
\label{sec:exp}

\subsection{Finetuning LLM-Judges on \dataset}
\label{sec:appendix_judge_ft}

Finetuning LLM-Judges uses 16$\times$H100 GPUs for \texttt{Qwen3-4B} and 192 GPUs for \texttt{Llama4-Scout}.

In \textbf{SFT}, models directly predict 5-point grades for response pairs without generating intermediate reasoning, trained with cross-entropy loss under teacher forcing. We use the TRL (\url{https://huggingface.co/docs/trl})  library and adopt the default learning rate of $2\text{e-}5$. Maximum sequence length is set to 8192.

In \textbf{RL}, we use GRPO with the \texttt{verl }(\url{https://github.com/volcengine/verl}) implementation,   keeping the default learning rate $1\text{e-}6$. We set rollout size to 8, and maximum length 4,096 tokens for both input and output. Prompts follow the template in \autoref{fig:template}, encouraging models to produce reasoning before assigning scores.

Batch size is set to 32, and we train up to three epochs and select best checkpoint based on the performance on the validation set.

\subsection{Post Training with RL}
\label{sec:appendix_post}

We set the learning rate to $1\text{e-}6$, the maximum tokens for the policy model to 1,024, and for the RM to 4,096, using up to 4$\times$H100 GPUs.
Following \citet{liu2025understanding}, we disable the length normalization term in the loss, as we find that otherwise responses tend to grow excessively long after training.

Since the judge is not trained to evaluate the thinking process but only the responses, we sample generations from the policy model \texttt{Qwen3-4B} in non-thinking mode. Comparison of the response quality before (\texttt{Qwen3-4B}) and after training (\texttt{Post-train}) are both done in thinking mode, as we find it leads to superior generation quality. When constructing preference pairs for the pairwise judge, we remove the thinking tokens from the generations.

Batch size is also set to 32, and we train up to three epochs and select best checkpoint based on the performance on the validation set.

\section{Additional Results}
\label{sec:additional}

\subsection{Judge Performance Per Dimension}
\label{sec:per_cat}
We report detailed results for all eight models across four metrics: \textit{Macro-F1} and \textit{Accuracy} for 5-way classification, \textit{Preference} accuracy over A win~/~A loss~/~Tie, and \textit{Krippendorff's}~$\alpha$ for agreement with human annotators. Results are shown for the four dimensions: Fluency, Localized Factuality, Localized Tone, and Tone.  

\autoref{tab:Zero-shot_Pointwise_with_Rubrics}, \autoref{tab:Zero-shot_Pairwise_with_Rubrics}, and \autoref{tab:Few-shot_Pointwise_with_Rubrics} present results for Zero-shot \textsc{Pointwise}, Zero-shot \textsc{Pairwise}, and Few-shot \textsc{Pointwise} \textit{with grading rubrics}.  
\autoref{tab:Zero-shot_Pointwise_without_Rubrics} and \autoref{tab:Zero-shot_Pairwise_without_Rubrics} report corresponding Zero-shot \textsc{Pointwise} and Zero-shot \textsc{Pairwise} results \textit{without grading rubrics}.
\autoref{tab:sft_rl_all} compares dimension-wise performance of \texttt{Qwen3-4B} and \texttt{Llama4-Scout} trained with SFT and RL on \textit{all} data, including both \textsc{Pointwise} and \textsc{Pairwise}. 

Overall, \textit{Localized Factuality} remains the most challenging dimension: both frontier API models and RL-trained models show limited improvement. This suggests that alternative training approaches, such as integrating search and tool use, may be necessary, which we leave for future work.  

\autoref{tab:rl_cat} further compares dimension-wise performance of \textsc{Pairwise} RL-trained \texttt{Qwen3-4B} on \textit{partial} subsets of the data. Specifically, we evaluate (i) models trained on a single dimension and tested across all dimensions to study cross-task transfer, and (ii) models trained only on English data and evaluated on all languages. Results show that optimizing on single dimension achieves performance similar to joint training (\autoref{tab:sft_rl_all}), highlighting the efficiency and practicality of joint training. In contrast, training only on English leads to degraded performance, revealing the challenges of cross-lingual transfer given the localized nature of our \dataset dataset.

\subsection{Judge Performance Per Language Variety}
\label{sec:per_locale}
\autoref{tab:all_lang_f1} and \autoref{tab:all_lang_preference} show Marco-F1 and Preference accuracy per Language Variety for baseline and fine-tuned \texttt{Qwen3-4B} and \texttt{Llama4-Scout} models.
\clearpage

\begin{table}[t]
\caption{\label{tab:Zero-shot_Pointwise_with_Rubrics} 
Results per Dimension: Zero-shot \textsc{Pointwise} \textit{with} Grading Rubrics.}
\setlength{\tabcolsep}{3pt}
\vspace{0.5em}

\centering
\small
\resizebox{\textwidth}{!}{
\begin{tabular}{llcccccccc}
\toprule

\multicolumn{2}{l}{\textbf{Zero-shot \textsc{Pointwise} \textit{with} Rubrics}}      & \texttt{Qwen3-4B} & \texttt{Qwen3-32B} & \texttt{Llama3.1-8B} & \texttt{Llama3.3-70B} & \texttt{Llama4-Scout} & \texttt{o3}  & \texttt{gpt-4o}   & \texttt{gpt-4.1} \\
        \midrule
\multirow{4}{*}{Overall}     & \textit{Macro-F1}         & 23.06    & 28.53     & 22.27       & 27.93        & 25.63        & 26.54 & 25.99  & 32.23   \\
        & \textit{Accuracy}         & 35.48    & 37.88     & 30.97       & 35.99        & 38.19        & 34.37 & 36.33  & 38.05   \\
        & \textit{Preference}       & 40.54    & 42.19     & 39.92       & 37.37        & 42.19        & 45.07 & 42.92  & 41.73   \\
        & \textit{Krippendorff’s}~$\alpha$ & 80.59    & 83.80     & 79.35       & 80.71        & 82.09        & 79.64 & 83.59  & 83.78   \\   \midrule
\multirow{4}{*}{Fluency}     & \textit{Macro-F1}   & 17.60    & 29.01     & 19.35       & 21.03        & 22.09        & 23.29 & 18.55  & 20.73   \\
        & \textit{Accuracy}   & 37.07    & 41.98     & 32.57       & 36.87        & 39.68        & 43.99 & 36.37  & 37.17   \\
        & \textit{Preference} & 37.68    & 41.28     & 40.48       & 35.27        & 43.52        & 41.28 & 34.47  & 34.87   \\
        & \textit{Krippendorff’s}~$\alpha$ & 76.73    & 81.47     & 77.85       & 78.29        & 80.35        & 83.19 & 76.59  & 76.47   \\   \midrule
\multirow{4}{*}{Localized Factuality} & \textit{Macro-F1}   & 14.94    & 15.77     & 14.17       & 18.57        & 16.74        & 12.04 & 19.27  & 20.25   \\
        & \textit{Accuracy}   & 30.16    & 26.22     & 26.49       & 34.10        & 31.93        & 13.86 & 29.48  & 31.11   \\
        & \textit{Preference} & 26.63    & 28.80     & 28.26       & 29.62        & 26.90        & 32.34 & 27.99  & 26.36   \\
        & \textit{Krippendorff’s}~$\alpha$ & 74.30    & 74.50     & 74.64       & 72.37        & 74.82        & 66.45 & 77.81  & 77.29   \\   \midrule
\multirow{4}{*}{Localized Tone}       & \textit{Macro-F1}   & 17.34    & 29.40     & 19.00       & 20.52        & 19.54        & 28.13 & 22.23  & 25.73   \\
        & \textit{Accuracy}   & 32.45    & 37.20     & 36.31       & 31.57        & 37.75        & 41.94 & 33.11  & 36.64   \\
        & \textit{Preference} & 41.72    & 45.25     & 36.42       & 32.89        & 41.94        & 50.99 & 46.80  & 48.12   \\
        & \textit{Krippendorff’s}~$\alpha$ & 76.32    & 80.41     & 78.43       & 73.92        & 78.84        & 80.89 & 78.80  & 80.27   \\   \midrule
\multirow{4}{*}{Tone}        & \textit{Macro-F1}   & 36.17    & 40.56     & 23.76       & 38.06        & 32.00        & 25.97 & 35.31  & 41.62   \\
        & \textit{Accuracy}   & 41.14    & 43.61     & 27.47       & 41.03        & 42.15        & 32.85 & 45.18  & 46.19   \\
        & \textit{Preference} & 54.04    & 51.12     & 52.47       & 50.67        & 53.59        & 53.81 & 60.76  & 55.61   \\
        & \textit{Krippendorff’s}~$\alpha$ & 87.16    & 88.91     & 81.41       & 88.48        & 87.61        & 82.13 & 90.58  & 90.21  

\\
\bottomrule  
\end{tabular}}
\end{table}

\begin{table}[t]
\caption{\label{tab:Zero-shot_Pairwise_with_Rubrics} 
  Results per Dimension: Zero-shot \textsc{Pairwise} \textit{with}   Grading Rubrics.}
\setlength{\tabcolsep}{3pt}
\vspace{0.5em}

\centering
\small
\resizebox{\textwidth}{!}{
\begin{tabular}{llcccccccc}
\toprule

\multicolumn{2}{l}{\textbf{Zero-shot \textsc{Pairwise} \textit{with} Rubrics}}     & \texttt{Qwen3-4B} & \texttt{Qwen3-32B} & \texttt{Llama3.1-8B} & \texttt{Llama3.3-70B} & \texttt{Llama4-Scout} & \texttt{o3}  & \texttt{gpt-4o}   & \texttt{gpt-4.1} \\
        \midrule
\multirow{4}{*}{Overall}     & \textit{Macro-F1}       & 35.46	&37.48	&29.46	&37.50	&36.11	&35.35	&37.57	&38.53   \\
        & \textit{Accuracy}       & 43.23	&40.88	&29.56	&43.12	&42.29	&37.26	&40.86	&44.48  \\
        & \textit{Preference}     & 57.13	&59.12	&50.45	&55.32	&56.25	&58.72	&57.98	&59.23   \\
        & \textit{Krippendorff’s $\alpha$} & 84.25	&85.60	&80.17	&85.29	&84.10	&83.97	&86.35	&85.65   \\    \midrule
\multirow{4}{*}{Fluency}     & \textit{Macro-F1}       & 32.24	&35.27	&26.67	&32.48	&35.11	&34.40	&34.67	&34.55 \\
        & \textit{Accuracy}       & 46.99	&45.99	&27.45	&43.99	&43.91	&44.09	&46.19	&50.10   \\
        & \textit{Preference}     & 55.91	&59.92	&50.50	&52.30	&54.03	&60.12	&56.51	&60.32   \\
        & \textit{Krippendorff’s $\alpha$} &83.05	&84.72	&80.21	&83.87&84.61&84.15&85.72&83.99   \\    \midrule
\multirow{4}{*}{Localized Factuality} & \textit{Macro-F1}       & 22.55&21.93&20.96&20.96&21.13&17.86&21.24&24.27 \\
        & \textit{Accuracy}       & 33.02&28.80&25.14&29.76&28.80&21.06&24.46&29.62   \\
        & \textit{Preference}     & 42.93&43.21&38.04&38.86&38.59&38.86&38.04&37.77   \\
        & \textit{Krippendorff’s $\alpha$} & 75.97&75.96&74.03&75.07&74.74&	71.41&	75.93&	75.54   \\    \midrule
\multirow{4}{*}{Localized Tone}       & \textit{Macro-F1}       & 32.82	&35.25&	30.22	&36.82	&33.62	&37.88&	37.86	&35.02  \\
        & \textit{Accuracy}       & 42.49&	43.27	&33.77	&46.69	&43.82	&44.15	&45.92&	46.14   \\
        & \textit{Preference}     & 60.49&	63.80	&51.21	&60.71&	61.37	&65.34	&64.24	&65.78   \\
        & \textit{Krippendorff’s $\alpha$} & 82.95	&85.18&	79.38	&84.28	&83.10	&86.93	&87.15&	85.44   \\    \midrule
\multirow{4}{*}{Tone}        & \textit{Macro-F1}       &43.06&	41.81	&31.66&	46.57	&44.32	&35.37&	42.24&	45.52 \\
        & \textit{Accuracy}       & 48.21	&42.71	&31.28&	49.55	&50.00&	35.99	&43.27&	48.77   \\
        & \textit{Preference}     & 66.82&	66.59	&59.87&	66.82	&68.16&	66.82&	69.73&	69.06   \\
        & \textit{Krippendorff’s $\alpha$} & 89.00	&89.69	&82.75	&90.84&	89.41	&87.64	&90.83&	90.48  

\\
\bottomrule  
\end{tabular}}
\end{table}

\begin{table}[t]
\caption{\label{tab:Few-shot_Pointwise_with_Rubrics} 
Results per Dimension: \textbf{Few}-shot \textsc{Pointwise} \textit{with} Grading Rubrics.}
\setlength{\tabcolsep}{3pt}
\vspace{0.5em}

\centering
\small
\resizebox{\textwidth}{!}{
\begin{tabular}{llcccccccc}
\toprule
\multicolumn{2}{l}{\textbf{Few-shot \textsc{Pointwise} \textit{with} Rubrics}}      & \texttt{Qwen3-4B} & \texttt{Qwen3-32B} & \texttt{Llama3.1-8B} & \texttt{Llama3.3-70B} & \texttt{Llama4-Scout} & \texttt{o3}  & \texttt{gpt-4o}   & \texttt{gpt-4.1} \\
        \midrule
\multirow{4}{*}{Overall}     & \textit{Macro-F1}   & 31.18    & 35.45     & 22.24       & 30.52        & 32.84        & 27.92 & 29.57  & 33.84   \\
        & \textit{Accuracy}   & 37.71    & 38.59     & 26.25       & 37.63        & 39.92        & 35.93 & 38.19  & 39.01   \\
        & \textit{Preference} & 39.35    & 42.87     & 37.15       & 38.56        & 41.22        & 44.68 & 45.87  & 44.00   \\
        & \textit{Krippendorff’s $\alpha$} & 82.36    & 84.46     & 77.00       & 81.45        & 83.07        & 81.54 & 84.84  & 84.24   \\\midrule
\multirow{4}{*}{Fluency}     & \textit{Macro-F1}   & 25.64    & 29.73     & 20.27       & 25.39        & 29.71        & 23.64 & 26.94  & 23.78   \\
        & \textit{Accuracy}   & 38.48    & 42.69     & 25.25       & 38.88        & 41.58        & 43.09 & 39.18  & 39.08   \\
        & \textit{Preference} & 37.68    & 41.48     & 37.07       & 39.08        & 44.29        & 40.08 & 38.88  & 40.08   \\
        & \textit{Krippendorff’s $\alpha$} & 79.93    & 82.20     & 76.35       & 79.76        & 82.11        & 83.70 & 80.22  & 78.79   \\\midrule
\multirow{4}{*}{Localized Factuality} & \textit{Macro-F1}   & 22.20    & 17.44     & 12.74       & 20.92        & 21.88        & 14.17 & 21.20  & 21.82   \\
        & \textit{Accuracy}   & 33.70    & 24.86     & 18.89       & 36.41        & 34.38        & 19.02 & 30.57  & 30.03   \\
        & \textit{Preference} & 30.71    & 33.97     & 32.88       & 27.45        & 25.54        & 29.89 & 32.88  & 25.00   \\
        & \textit{Krippendorff’s $\alpha$} & 75.81    & 74.84     & 69.38       & 74.54        & 76.44        & 69.40 & 77.81  & 76.25   \\\midrule
\multirow{4}{*}{Localized Tone}       & \textit{Macro-F1}   & 25.26    & 31.27     & 19.86       & 23.14        & 27.45        & 29.00 & 28.83  & 29.46   \\
        & \textit{Accuracy}   & 34.88    & 38.85     & 31.13       & 33.89        & 38.41        & 42.05 & 36.64  & 37.53   \\
        & \textit{Preference} & 39.51    & 43.27     & 32.67       & 38.19        & 42.16        & 52.32 & 48.57  & 47.46   \\
        & \textit{Krippendorff’s $\alpha$} & 78.56    & 81.55     & 77.51       & 75.65        & 79.13        & 81.34 & 81.15  & 80.48   \\ \midrule
\multirow{4}{*}{Tone}        & \textit{Macro-F1}   & 37.11    & 41.90     & 25.14       & 38.16        & 39.60        & 28.43 & 36.46  & 43.29   \\
        & \textit{Accuracy}   & 43.05    & 45.07     & 28.48       & 41.03        & 44.17        & 35.65 & 44.96  & 47.87   \\
        & \textit{Preference} & 48.21    & 51.35     & 45.29       & 47.53        & 49.78        & 54.26 & 61.66  & 60.54   \\
        & \textit{Krippendorff’s $\alpha$} & 88.02    & 89.72     & 80.00       & 87.71        & 87.76        & 84.60 & 90.78  & 90.73  
\\
\bottomrule  
\end{tabular}}
\end{table}

\begin{table}[t]
\caption{\label{tab:Zero-shot_Pointwise_without_Rubrics} 
Results per Dimension: Zero-shot \textsc{Pointwise} \textit{without}   Grading Rubrics.}
\setlength{\tabcolsep}{3pt}
\vspace{0.5em}

\centering
\small
\resizebox{\textwidth}{!}{
\begin{tabular}{llcccccccc}
\toprule
\multicolumn{2}{l}{\textbf{Zero-shot \textsc{Pointwise} \textit{without} Rubrics}}      & \texttt{Qwen3-4B} & \texttt{Qwen3-32B} & \texttt{Llama3.1-8B} & \texttt{Llama3.3-70B} & \texttt{Llama4-Scout} & \texttt{o3}  & \texttt{gpt-4o}   & \texttt{gpt-4.1} \\
        \midrule
\multirow{4}{*}{Overall}     & \textit{Macro-F1}       & 16.00    & 25.59     & 21.50       & 22.71        & 22.15        & 25.43 & 22.45  & 22.26   \\
        & \textit{Accuracy}       & 32.16    & 36.24     & 33.18       & 33.52        & 36.24        & 35.14 & 34.88  & 34.54   \\
        & \textit{Preference}     & 33.52    & 43.32     & 38.34       & 34.54        & 41.28        & 45.13 & 37.60  & 38.67   \\
        & \textit{Krippendorff’s $\alpha$} & 76.05    & 81.63     & 79.70       & 78.18        & 79.93        & 80.41 & 80.66  & 80.12   \\     \midrule
\multirow{4}{*}{Fluency}     & \textit{Macro-F1}       & 10.84    & 19.70     & 20.73       & 21.11        & 23.11        & 21.85 & 15.96  & 17.38   \\
        & \textit{Accuracy}       & 34.37    & 39.18     & 36.77       & 37.17        & 37.58        & 40.48 & 36.37  & 35.77   \\
        & \textit{Preference}     & 32.46    & 39.48     & 40.08       & 35.27        & 41.48        & 48.10 & 31.46  & 31.66   \\
        & \textit{Krippendorff’s $\alpha$} & 73.46    & 78.53     & 79.31       & 77.32        & 78.84        & 80.15 & 76.22  & 74.64   \\     \midrule
\multirow{4}{*}{Localized Factuality} & \textit{Macro-F1}       & 12.94    & 19.93     & 15.22       & 18.30        & 17.03        & 13.46 & 22.09  & 20.02   \\
        & \textit{Accuracy}       & 31.39    & 32.20     & 29.62       & 33.56        & 32.34        & 20.52 & 32.34  & 31.79   \\
        & \textit{Preference}     & 26.90    & 34.51     & 27.45       & 23.10        & 26.90        & 31.52 & 23.37  & 29.08   \\
        & \textit{Krippendorff’s $\alpha$} & 71.05    & 78.12     & 75.43       & 70.85        & 72.31        & 70.56 & 75.15  & 75.63   \\     \midrule
\multirow{4}{*}{Localized Tone}       & \textit{Macro-F1}       & 12.23    & 25.65     & 19.38       & 18.32        & 21.19        & 25.21 & 18.75  & 25.15   \\
        & \textit{Accuracy}       & 28.70    & 33.33     & 35.65       & 29.03        & 32.45        & 40.84 & 31.68  & 33.44   \\
        & \textit{Preference}     & 29.80    & 45.25     & 40.84       & 34.44        & 41.06        & 51.88 & 43.49  & 43.93   \\
        & \textit{Krippendorff’s $\alpha$} & 72.07    & 78.80     & 79.17       & 73.46        & 75.85        & 81.75 & 78.59  & 78.90   \\     \midrule
\multirow{4}{*}{Tone}        & \textit{Macro-F1}       & 26.06    & 30.19     & 23.79       & 26.79        & 29.16        & 27.04 & 35.65  & 27.11   \\
        & \textit{Accuracy}       & 33.86    & 39.24     & 29.60       & 33.97        & 41.82        & 35.43 & 38.57  & 36.55   \\
        & \textit{Preference}     & 43.95    & 52.91     & 42.83       & 43.27        & 53.14        & 46.19 & 50.22  & 49.10   \\
        & \textit{Krippendorff’s $\alpha$} & 81.59    & 85.54     & 80.92       & 83.85        & 85.49        & 82.81 & 86.68  & 85.60  
\\
\bottomrule  
\end{tabular}}
\end{table}

\begin{table}[t]
\caption{\label{tab:Zero-shot_Pairwise_without_Rubrics} 
Results per Dimension: Zero-shot \textsc{Pairwise} \textit{without} Grading Rubrics.}
\setlength{\tabcolsep}{3pt}
\vspace{0.5em}

\centering
\small
\resizebox{\textwidth}{!}{
\begin{tabular}{llcccccccc}
\toprule

\multicolumn{2}{l}{\textbf{Zero-shot \textsc{Pairwise} \textit{without} Rubrics}}      & \texttt{Qwen3-4B} & \texttt{Qwen3-32B} & \texttt{Llama3.1-8B} & \texttt{Llama3.3-70B} & \texttt{Llama4-Scout} & \texttt{o3}  & \texttt{gpt-4o}   & \texttt{gpt-4.1} \\
        \midrule
\multirow{4}{*}{Overall}     & \textit{Macro-F1}       & 32.74    & 38.10     & 30.89       & 35.12        & 35.21        & 37.60 & 36.74  & 37.35   \\
        & \textit{Accuracy}       & 40.74    & 41.90     & 33.10       & 42.44        & 41.53        & 40.12 & 41.79  & 44.45   \\
        & \textit{Preference}     & 54.08    & 59.23     & 49.55       & 56.29        & 55.10        & 57.98 & 56.85  & 56.96   \\
        & \textit{Krippendorff’s $\alpha$} & 82.44    & 85.66     & 81.73       & 83.97        & 83.99        & 84.46 & 85.38  & 84.23   \\\midrule
\multirow{4}{*}{Fluency}     & \textit{Macro-F1}       & 31.08    & 38.71     & 30.27       & 33.42        & 32.34        & 34.97 & 32.98  & 30.86   \\
        & \textit{Accuracy}       & 43.89    & 48.30     & 33.37       & 45.29        & 42.99        & 47.80 & 44.09  & 46.19   \\
        & \textit{Preference}     & 55.11    & 59.72     & 46.49       & 53.71        & 52.91        & 58.72 & 51.70  & 52.51   \\
        & \textit{Krippendorff’s $\alpha$} & 82.89    & 86.06     & 81.93       & 83.74        & 84.16        & 84.12 & 84.35  & 82.00   \\\midrule
\multirow{4}{*}{Localized Factuality} & \textit{Macro-F1}       & 22.05    & 22.01     & 19.59       & 21.35        & 18.58        & 20.04 & 24.20  & 22.67   \\
        & \textit{Accuracy}       & 30.98    & 28.53     & 27.58       & 30.71        & 29.48        & 24.46 & 30.30  & 32.20   \\
        & \textit{Preference}     & 40.49    & 43.48     & 36.41       & 41.85        & 36.41        & 38.86 & 40.22  & 41.30   \\
        & \textit{Krippendorff’s $\alpha$} & 75.45    & 76.97     & 73.96       & 74.51        & 73.68        & 72.82 & 76.23  & 75.29   \\\midrule
\multirow{4}{*}{Localized Tone}       & \textit{Macro-F1}       & 32.51    & 36.30     & 30.43       & 35.10        & 34.11        & 40.45 & 36.38  & 37.36   \\
        & \textit{Accuracy}       & 43.27    & 42.38     & 34.77       & 44.59        & 43.38        & 45.36 & 45.25  & 48.01   \\
        & \textit{Preference}     & 55.85    & 63.36     & 52.54       & 60.93        & 62.25        & 65.78 & 66.45  & 64.02   \\
        & \textit{Krippendorff’s $\alpha$} & 82.27    & 86.05     & 81.30       & 84.23        & 83.41        & 86.12 & 86.30  & 85.10   \\\midrule
\multirow{4}{*}{Tone}        & \textit{Macro-F1}       & 36.97    & 42.94     & 34.84       & 40.71        & 43.33        & 39.54 & 42.42  & 44.67   \\
        & \textit{Accuracy}       & 42.71    & 45.29     & 35.65       & 46.75        & 47.98        & 39.13 & 45.18  & 48.99   \\
        & \textit{Preference}     & 62.33    & 67.49     & 60.76       & 66.37        & 65.70        & 65.02 & 66.59  & 67.71   \\
        & \textit{Krippendorff’s $\alpha$} & 84.96    & 89.18     & 86.05       & 87.84        & 88.36        & 88.49 & 89.52  & 88.98  

\\
\bottomrule  
\end{tabular}}
\end{table}

\begin{table}[t]
\caption{\label{tab:sft_rl_all} 
  Results per Dimension: SFT and RL trained \texttt{Qwen3-4B} and RL trained \texttt{Llama4-Scout} on \textit{All} Data with \textsc{Pointwise} and \textsc{Pairwise} Scoring.}
\setlength{\tabcolsep}{3.5pt}
\vspace{0.5em}

\centering
\small
\resizebox{\textwidth}{!}{
\begin{tabular}{llcccccc}
\toprule

\multicolumn{2}{l}{\multirow{2}{*}{SFT and RL on \textit{All} Data}} & \multicolumn{2}{c}{\texttt{Qwen3-4B} SFT} & \multicolumn{2}{c}{\texttt{Qwen3-4B} RL}  & \multicolumn{2}{c}{\texttt{Llama4-Scout}} \\
\multicolumn{2}{l}{}          & \sc Pointwise       & \sc Pairwise       & \sc Pointwise       & \sc Pairwise   & \sc Pairwise-SFT      & \sc Pairwise-SFT+RL   \\

          \cmidrule(r){1-2}               \cmidrule(lr){3-4}\cmidrule(lr){5-6} \cmidrule(l){7-8}
\multirow{4}{*}{Overall}     & \textit{Macro-F1}   & 30.26  & 33.44 & 28.87  & 39.44        &     45.04	 &45.82 \\
        & \textit{Accuracy}   & 36.64  & 35.82 & 38.22  & 46.83  &        50.17	  & 50.99\\
        & \textit{Preference} & 41.17  & 53.68 & 39.86  & 60.02 &     60.53	     & 61.10\\
        & \textit{Krippendorff’s $\alpha$} & 83.90  & 84.03 & 82.10  & 86.55         &         89.48    & 89.67\\\midrule
\multirow{4}{*}{Fluency}     & \textit{Macro-F1}   & 28.41  & 32.91 & 25.10  & 35.72        &         46.42     &47.52 \\
        & \textit{Accuracy}   & 38.18  & 37.17 & 39.38  & 52.51   &55.21	         &56.71 \\
        & \textit{Preference} & 38.48  & 54.71 & 42.89  & 61.92  &66.13	       & 66.53\\
        & \textit{Krippendorff’s $\alpha$} & 82.36  & 85.16 & 80.36  & 85.69         &       90.77    &90.86\\\midrule
\multirow{4}{*}{Localized Factuality} & \textit{Macro-F1}   & 20.30  & 19.51 & 17.49  & 20.62         &      25.30	    & 25.87\\
        & \textit{Accuracy}   & 31.66  & 25.14 & 33.42  & 33.02   &      34.78	  & 35.33\\
        & \textit{Preference} & 35.87  & 39.67 & 26.63  & 38.86 &    36.68	        & 36.68\\
        & \textit{Krippendorff’s $\alpha$} & 76.39  & 74.43 & 75.37  & 77.04        &     80.25      &80.14\\\midrule
\multirow{4}{*}{Localized Tone}       & \textit{Macro-F1}   & 27.43  & 33.38 & 25.58  & 38.56         &    40.98	    &41.22 \\
        & \textit{Accuracy}   & 37.75  & 41.06 & 37.86  & 47.35 &      53.53	  & 53.20\\
        & \textit{Preference} & 39.29  & 59.38 & 37.53  & 67.55 &    63.58	        & 65.12\\
        & \textit{Krippendorff’s $\alpha$} & 80.54  & 81.83 & 78.89  & 86.61         &      87.87      &88.23 \\\midrule
\multirow{4}{*}{Tone}        & \textit{Macro-F1}   & 33.35  & 34.91 & 33.98  & 46.82         &     51.08  &52.29 \\
        & \textit{Accuracy}   & 37.89  & 37.78 & 41.26  & 51.35         &    	53.81     & 55.27\\
        & \textit{Preference} & 50.45  & 58.30 & 49.78  &67.71         &      	70.85       & 71.08\\
        & \textit{Krippendorff’s $\alpha$} & 88.08  & 87.91 & 86.67  & 90.39        & 	92.63       &  93.00

\\
\bottomrule  
\end{tabular}}
\end{table}

\begin{table}[t]
\caption{\label{tab:single_multi} 
Comparison of zero-shot \textsc{Pairwise} \texttt{Qwen3-4B} and RL trained models, trained either jointly across all dimensions (multi-task) or individually per dimension (single-task).
}

\setlength{\tabcolsep}{4pt}
\vspace{0.5em}
\centering
\small
\resizebox{\textwidth}{!}{
\begin{tabular}{lcccccc}
\toprule

\multirow{2}{*}{\textbf{Dimension}} & \multicolumn{3}{c}{{\textbf{Macro-F1}}}           & \multicolumn{3}{c}{{\textbf{Preference Accuracy}}} \\

& \sc {Zero-shot}  & \sc{Multi-task}  & \sc{Single-task}   & \sc {Zero-shot}  & \sc{Multi-task}  & \sc{Single-task}                                    \\

          \cmidrule(r){1-1}               \cmidrule(lr){2-4}\cmidrule(l){5-7}
Fluency                            &   32.24	&35.72&	\textbf{37.14}	&55.91&\textbf{61.92}	&	61.32            \\ {Tone  }  &43.06	&\textbf{46.82}&	46.18&	66.82&	67.71&	\textbf{69.28} \\  

 {Localized Tone}            &32.82	&	\textbf{38.56}&37.61&	60.49	&\textbf{67.55}&	66.67  \\
   {Localized Factuality }              &    22.55	&20.62	&\textbf{23.12}&	\textbf{42.93}&38.86	&	42.12         \\ 

\bottomrule  
\end{tabular}}
\end{table}

\begin{table}[t]
\caption{\label{tab:rl_cat} 
  Results per Dimension:  RL trained \texttt{Qwen3-4B} on \textsc{Pairwise} Single Dimension Data and \textsc{Pairwise} English Data.}
\vspace{0.5em}
\centering
\small
\resizebox{\textwidth}{!}{
\begin{tabular}{llccccc}
\toprule
\multicolumn{2}{l}{}      & \multicolumn{4}{c}{\textit{Single Dimension Data on All languages}}                     & \textit{English Data on} \\
\multicolumn{2}{l}{\multirow{-2}{*}{\textsc{Pairwise} RL on \textit{Partial} Data}}         & Fluency  & Localized Factuality & Localized Tone       & Tone     & \textit{All Categories}  \\          \cmidrule(r){1-2}               \cmidrule(lr){3-6}\cmidrule(l){7-7} 
                    \multirow{4}{*}{Overall} & \textit{Macro-F1}       & 37.89    & 35.33    & 37.46    & 38.55    & 34.34           \\
         & \textit{Accuracy}       & 44.56    &43.74    & 43.69    & 45.19    & 42.33           \\
         & \textit{Preference}     & 59.29    & 56.68   & 59.29    & 57.53    & 56.46           \\
     & \textit{Krippendorff’s $\alpha$} & 85.63    & 84.06   & 86.12    & 86.20    & 83.98           \\\midrule
                   \multirow{4}{*}{Fluency}  & \textit{Macro-F1}       & \cellcolor[HTML]{F3F3F3}37.14 & 31.53   & 36.07    & 34.82    & 30.65           \\
         & \textit{Accuracy}       & \cellcolor[HTML]{F3F3F3}51.80 & 47.39    & 50.00    & 50.60    & 45.59           \\
         & \textit{Preference}     & \cellcolor[HTML]{F3F3F3}61.32 & 56.91    & 58.72    & 56.51    & 55.31           \\
     & \textit{Krippendorff’s $\alpha$} & \cellcolor[HTML]{F3F3F3}85.87 & 82.98    & 86.09    & 84.85    & 82.64           \\\midrule
     \multirow{4}{*}{Localized Factuality}     & \textit{Macro-F1}       & 20.12    & \cellcolor[HTML]{F3F3F3}23.12 & 18.87    & 19.78    & 19.92           \\
         & \textit{Accuracy}       & 29.89    & \cellcolor[HTML]{F3F3F3}34.10 & 29.21    & 30.71    & 32.61           \\
         & \textit{Preference}     & 42.12    & \cellcolor[HTML]{F3F3F3}42.12 & 40.22    & 41.03    & 41.85           \\
& \textit{Krippendorff’s $\alpha$} & 75.19    & \cellcolor[HTML]{F3F3F3}76.65 & 75.83 & 76.58    & 76.44           \\\midrule
     \multirow{4}{*}{Localized Tone}      & \textit{Macro-F1}       & 35.84    & 30.29    & \cellcolor[HTML]{F3F3F3}37.61 & 37.32    & 28.51           \\
         & \textit{Accuracy}       & 44.92    & 42.49   & \cellcolor[HTML]{F3F3F3}47.57 & 45.92    & 41.28           \\
         & \textit{Preference}     & 63.36    &58.72    & \cellcolor[HTML]{F3F3F3}66.67 & 60.49    & 57.40           \\
 & \textit{Krippendorff’s $\alpha$} & 84.20    & 82.01    & \cellcolor[HTML]{F3F3F3}86.28 & 84.87    & 80.87       \\ \midrule        
\multirow{4}{*}{Tone}       & \textit{Macro-F1}       & 43.01    & 43.96    & 41.61    & \cellcolor[HTML]{F3F3F3}46.18 & 42.50           \\
         & \textit{Accuracy}       & 48.21    & 48.88   & 44.62    & \cellcolor[HTML]{F3F3F3}50.34 & 47.76 \\
         & \textit{Preference}     & 67.04    & 66.37    & 68.16    & \cellcolor[HTML]{F3F3F3}69.28 & 68.83 \\
       & \textit{Krippendorff’s $\alpha$} & 89.66    & 89.28    & 89.61    & \cellcolor[HTML]{F3F3F3}91.28 & 88.89

\\
\bottomrule  
\end{tabular}}
\end{table}

\begin{table}[t]
\caption{\label{tab:all_lang_f1} 
  \textbf{Macro-F1} scores per Language Variety: Comparing \textsc{Pairwise} \texttt{Qwen3-4B} and \texttt{Llama4-Scout} zero-shot performance and various trained models.}
\vspace{0.5em}
\centering
\small
\resizebox{\textwidth}{!}{
\begin{tabular}{lccccccc}
\toprule
\multirow{2}{*}{\sc PAIRWISE} & \multicolumn{4}{c}{\texttt{Qwen3-4B}}         & \multicolumn{3}{c}{\texttt{Llama4-Scout}} \\
     & \textit{Zero-shot} & \textit{SFT} & \textit{RL}    & \textit{RL on EN-only data} & \textit{Zero-shot} & \textit{SFT} & \textit{SFT + RL}    \\\cmidrule(r){1-1} \cmidrule(lr){2-5} \cmidrule(l){6-8}
Overall      & 35.46 & 33.55 & 39.44 & 34.34 & 36.11 & 44.17 & 45.82 \\
\texttt{ar}       & 31.21 & 21.92 & 36.26 & 33.19 & 32.29 & 36.20 & 43.71 \\
\texttt{ar\_Latn\_EG}     & 16.41 & 22.60 & 9.62  & 17.52 & 18.71 & 54.88 & 65.12 \\
\texttt{bg\_BG }      & 29.95 & 28.12 & 37.80 & 26.39 & 26.75 & 31.31 & 33.02 \\
\texttt{bn\_BD}       & 24.20 & 15.08 & 20.17 & 18.04 & 13.84 & 20.14 & 23.37 \\
\texttt{cs\_CZ}      & 34.58 & 18.81 & 38.01 & 34.44 & 35.39 & 41.97 & 44.86 \\
\texttt{da\_DK}      & 21.50 & 17.91 & 29.43 & 24.78 & 31.71 & 34.98 & 40.73 \\
\texttt{de\_DE}       & 27.63 & 26.70 & 24.76 & 17.94 & 16.51 & 37.35 & 34.26 \\
\texttt{el\_GR}       & 36.90 & 37.84 & 43.74 & 40.31 & 39.68 & 45.11 & 41.64 \\
\texttt{en\_AU}       & 34.94 & 40.74 & 55.79 & 35.79 & 42.68 & 41.19 & 41.26 \\
\texttt{en\_GB}       & 44.86 & 46.77 & 46.13 & 48.93 & 42.52 & 40.69 & 51.55 \\
\texttt{en\_IN}       & 39.11 & 27.62 & 36.47 & 48.68 & 44.96 & 37.31 & 39.40 \\
\texttt{en\_US}       & 33.42 & 34.49 & 29.09 & 33.25 & 47.68 & 28.35 & 23.13 \\
\texttt{es\_ES}       & 38.97 & 29.93 & 41.21 & 30.53 & 29.23 & 27.44 & 27.07 \\
\texttt{es\_MX}       & 42.01 & 24.56 & 51.77 & 46.63 & 44.57 & 36.80 & 38.98 \\
\texttt{fa\_IR}       & 31.71 & 33.25 & 39.11 & 33.04 & 38.75 & 33.91 & 43.11 \\
\texttt{fr\_FR}       & 21.33 & 29.53 & 39.02 & 25.13 & 30.49 & 19.65 & 33.97 \\
\texttt{gu\_IN}       & 30.00 & 37.60 & 46.05 & 30.10 & 49.79 & 46.29 & 43.01 \\
\texttt{he\_IL}       & 24.89 & 19.71 & 25.75 & 26.86 & 22.66 & 32.50 & 38.17 \\
\texttt{hi\_IN}       & 23.46 & 16.16 & 36.77 & 30.04 & 27.91 & 28.81 & 36.42 \\
\texttt{hi\_Latn\_IN} & 48.52 & 27.56 & 41.53 & 39.74 & 34.52 & 25.79 & 53.80 \\
\texttt{hr\_HR}       & 20.16 & 34.86 & 25.15 & 16.69 & 18.45 & 26.57 & 29.01 \\
\texttt{hu\_HU}       & 27.69 & 36.63 & 37.06 & 28.84 & 33.83 & 40.92 & 39.55 \\
\texttt{id\_ID}       & 41.40 & 45.21 & 42.53 & 31.33 & 35.64 & 41.28 & 42.98 \\
\texttt{it\_IT}       & 29.48 & 34.77 & 40.28 & 31.61 & 29.16 & 24.51 & 29.34 \\
\texttt{ja\_JP}      & 45.06 & 40.34 & 39.68 & 44.76 & 35.37 & 44.98 & 41.55 \\
\texttt{ko\_KR}       & 35.51 & 35.14 & 33.03 & 32.59 & 39.12 & 38.38 & 48.80 \\
\texttt{mr\_IN}       & 41.44 & 36.52 & 51.06 & 47.78 & 42.84 & 50.10 & 51.59 \\
\texttt{ms\_MY}       & 29.50 & 30.79 & 39.74 & 33.02 & 28.99 & 36.27 & 44.60 \\
\texttt{ne\_NP}       & 28.05 & 22.45 & 27.59 & 19.41 & 24.62 & 20.16 & 24.54 \\
\texttt{nl\_NL}       & 36.00 & 35.38 & 38.80 & 27.61 & 45.25 & 54.06 & 51.22 \\
\texttt{pl\_PL}       & 28.20 & 33.52 & 21.82 & 23.77 & 17.94 & 29.00 & 21.29 \\
\texttt{pt\_BR}       & 45.17 & 36.52 & 48.48 & 43.34 & 48.19 & 48.45 & 41.83 \\
\texttt{pt\_PT}       & 38.72 & 39.29 & 44.62 & 41.22 & 38.89 & 52.15 & 41.57 \\
\texttt{ro\_RO}       & 37.54 & 42.92 & 49.36 & 46.85 & 51.91 & 52.42 & 55.46 \\
\texttt{ru\_RU}       & 31.75 & 22.59 & 22.44 & 19.78 & 28.84 & 18.72 & 21.61 \\
\texttt{sk\_SK}       & 35.80 & 38.22 & 44.14 & 33.79 & 37.20 & 44.29 & 40.69 \\
\texttt{sv\_SE}       & 31.59 & 35.11 & 33.78 & 27.44 & 34.29 & 39.48 & 40.81 \\
\texttt{sw\_KE}       & 41.97 & 28.13 & 41.88 & 21.36 & 43.13 & 37.33 & 39.67 \\
\texttt{th\_TH}       & 37.07 & 32.75 & 47.04 & 35.98 & 45.82 & 52.77 & 55.11 \\
\texttt{tl\_PH}       & 40.71 & 29.19 & 45.39 & 42.52 & 39.95 & 40.54 & 37.78 \\
\texttt{tr\_TR}       & 50.03 & 37.30 & 48.50 & 40.67 & 40.66 & 40.45 & 45.42 \\
\texttt{uk\_UA}       & 24.09 & 29.20 & 20.45 & 22.09 & 17.57 & 23.98 & 27.13 \\
\texttt{ur\_Latn\_PK} & 29.38 & 38.54 & 34.25 & 27.72 & 32.49 & 32.93 & 29.59 \\
\texttt{ur\_PK}       & 23.21 & 38.81 & 36.11 & 28.29 & 39.87 & 48.94 & 43.73 \\
\texttt{vi\_VN}       & 33.91 & 31.07 & 34.76 & 35.16 & 30.35 & 40.51 & 38.46 \\
\texttt{zh\_CN}       & 40.82 & 27.78 & 51.99 & 41.15 & 37.42 & 45.27 & 41.58 \\
\texttt{zh\_TW}       & 35.38 & 25.47 & 37.07 & 44.80 & 37.93 & 39.29 & 38.09
\\
\bottomrule  
\end{tabular}}
\end{table}

\begin{table}[t]
\caption{\label{tab:all_lang_preference} 
  \textbf{Preference} accuracy per Language: Comparing \textsc{Pairwise} \texttt{Qwen3-4B} and \texttt{Llama4-Scout} zero-shot performance and various trained models.}
\vspace{0.5em}
\centering
\small
\resizebox{\textwidth}{!}{
\begin{tabular}{lccccccc}
\toprule
\multirow{2}{*}{\sc PAIRWISE} & \multicolumn{4}{c}{\texttt{Qwen3-4B}}         & \multicolumn{3}{c}{\texttt{Llama4-Scout}} \\
     & \textit{Zero-shot} & \textit{SFT} & \textit{RL}    & \textit{RL on EN-only data} & \textit{Zero-shot} & \textit{SFT} & \textit{SFT + RL}    \\\cmidrule(r){1-1} \cmidrule(lr){2-5} \cmidrule(l){6-8}
Overall      & 57.13 & 53.51 & 60.02 & 56.46 & 56.25 & 60.08 & 61.10 \\
\texttt{ar}       & 63.16 & 50.00 & 60.53 & 57.89 & 61.54 & 73.68 & 55.26 \\
\texttt{ar\_Latn\_EG}      & 35.48 & 25.81 & 38.71 & 45.16 & 28.12 & 93.55 & 93.55 \\
\texttt{bg\_BG}      & 51.28 & 56.41 & 48.72 & 38.46 & 47.50 & 46.15 & 46.15 \\
\texttt{bn\_BD}       & 41.38 & 27.59 & 37.93 & 44.83 & 36.67 & 41.38 & 41.38 \\
\texttt{cs\_CZ}       & 69.23 & 74.36 & 76.92 & 69.23 & 70.00 & 74.36 & 74.36 \\
\texttt{da\_DK}       & 35.90 & 38.46 & 43.59 & 46.15 & 40.00 & 51.28 & 46.15 \\
\texttt{de\_DE}       & 51.61 & 54.84 & 58.06 & 38.71 &59.38 & 54.84 & 54.84 \\
\texttt{el\_GR}       & 56.41 & 56.41 & 61.54 & 66.67 & 57.50 & 66.67 & 64.10 \\
\texttt{en\_AU}       & 34.21 & 47.37 & 39.47 & 36.84 & 34.21 & 39.47 & 44.74 \\
\texttt{en\_GB}       & 53.85 & 64.10 & 53.85 & 58.97 & 43.59 & 46.15 & 58.97 \\
\texttt{en\_IN}       & 43.59 & 41.03 & 48.72 & 43.59 & 51.28 & 51.28 & 43.59 \\
\texttt{en\_US}       & 53.12 & 46.88 & 40.62 & 62.50 & 43.75 & 50.00 & 46.88 \\
\texttt{es\_ES}       & 64.10 & 51.28 & 48.72 & 56.41 & 51.28 & 51.28 & 51.28 \\
\texttt{es\_MX}       & 53.85 & 35.90 & 56.41 & 46.15 & 43.59 & 51.28 & 51.28 \\
\texttt{fa\_IR}       & 66.67 & 69.23 & 66.67 & 61.54 & 56.41 & 58.97 & 79.49 \\
\texttt{fr\_FR}       & 60.53 & 55.26 & 57.89 & 52.63 & 55.26 & 42.11 & 60.53 \\
\texttt{gu\_IN}       & 48.65 & 43.24 & 64.86 & 64.86 & 67.57 & 56.76 & 62.16 \\
\texttt{he\_IL}       & 58.97 & 61.54 & 61.54 & 64.10 & 61.54 & 48.72 & 64.10 \\
\texttt{hi\_IN}       & 53.12 & 46.88 & 62.50 & 59.38 & 59.38 & 59.38 & 71.88 \\
\texttt{hi\_Latn\_IN} & 61.54 & 48.72 & 66.67 & 71.79 & 62.50 & 56.41 & 64.10 \\
\texttt{hr\_HR}       & 43.59 & 56.41 & 48.72 & 51.28 & 51.28 & 51.28 & 51.28 \\
\texttt{hu\_HU}       & 55.26 & 52.63 & 63.16 & 63.16 & 60.53 & 65.79 & 65.79 \\
\texttt{id\_ID}       & 75.68 & 70.27 & 81.08 & 75.68 & 78.38 & 83.78 & 81.08 \\
\texttt{it\_IT}       & 59.38 & 59.38 & 62.50 & 46.88 & 53.12 & 59.38 & 59.38 \\
\texttt{ja\_JP}       & 74.36 & 71.79 & 71.79 & 61.54 & 61.54 & 71.79 & 74.36 \\
\texttt{ko\_KR}       & 63.16 & 65.79 & 76.32 & 68.42 & 71.05 & 81.58 & 84.21 \\
\texttt{mr\_IN}       & 56.41 & 53.85 & 69.23 & 58.97 & 56.41 & 66.67 & 64.10 \\
\texttt{ms\_MY}       & 58.97 & 53.85 & 58.97 & 58.97 & 61.54 & 56.41 & 61.54 \\
\texttt{ne\_NP}       & 52.63 & 57.89 & 44.74 & 44.74 & 50.00 & 47.37 & 42.11 \\
\texttt{nl\_NL}       & 60.53 & 65.79 & 63.16 & 50.00 & 63.16 & 71.05 & 71.05 \\
\texttt{pl\_PL}       & 52.63 & 47.37 & 55.26 & 57.89 & 57.89 & 47.37 & 60.53 \\
\texttt{pt\_BR}       & 53.85 & 53.85 & 41.03 & 53.85 & 38.46 & 51.28 & 51.28 \\
\texttt{pt\_PT}       & 44.74 & 44.74 & 55.26 & 47.37 & 50.00 & 57.89 & 47.37 \\
\texttt{ro\_RO}       & 58.97 & 64.10 & 76.92 & 56.41 & 58.97 & 64.10 & 74.36 \\
\texttt{ru\_RU}      & 50.00 & 40.62 & 46.88 & 40.62 & 53.12 & 43.75 & 50.00 \\
\texttt{sk\_SK}       & 69.23 & 56.41 & 74.36 & 66.67 & 58.97 & 69.23 & 66.67 \\
\texttt{sv\_SE}       & 56.41 & 41.03 & 46.15 & 53.85 & 43.59 & 48.72 & 35.90 \\
\texttt{sw\_KE}       & 56.41 & 58.97 & 71.79 & 56.41 & 66.67 & 69.23 & 69.23 \\
\texttt{th\_TH}       & 53.85 & 53.85 & 51.28 & 58.97 & 58.97 & 48.72 & 51.28 \\
\texttt{tl\_PH}       & 69.23 & 61.54 & 71.79 & 58.97 & 64.10 & 71.79 & 66.67 \\
\texttt{tr\_TR}       & 71.79 & 61.54 & 82.05 & 71.79 & 79.49 & 84.62 & 76.92 \\
\texttt{uk\_UA}       & 58.97 & 38.46 & 66.67 & 51.28 & 46.15 & 64.10 & 61.54 \\
\texttt{ur\_Latn\_PK} & 53.85 & 66.67 & 61.54 & 53.85 & 55.00 & 66.67 & 64.10 \\
\texttt{ur\_PK}       & 69.23 & 43.59 & 69.23 & 53.85 & 66.67 & 58.97 & 64.10 \\
\texttt{vi\_VN}       & 51.28 & 48.72 & 56.41 & 58.97 & 56.41 & 58.97 & 58.97 \\
\texttt{zh\_CN}       & 61.54 & 48.72 & 64.10 & 61.54 & 64.10 & 66.67 & 61.54 \\
\texttt{zh\_TW}      & 84.62 & 66.67 & 82.05 & 74.36 & 76.92 & 79.49 & 79.49
\\
\bottomrule  
\end{tabular}}
\end{table}

\clearpage

\section{Examples of LLM-Judge Output on \dataset}
\label{sec:judge_ouput}
\subsection{Comparison: with and without Rubrics}

In this section we compare zero-shot LLM-judge output with and with provided rubrics. On average, LLM-judges generate 15-20\% longer output when rubrics are present, where the additional text is attributed to referencing the grading criteria (marked in blue).

\begin{figure*}[h]
    \centering
    \begin{prompt}{\texttt{gpt-4.1} Judge Output: Zero-shot Pairwise without Rubrics}

<think>\\
Response A is clear, well-organized, and written in a professional and emotionally appealing tone. The vocabulary is appropriate, and there are no grammatical errors or awkward sentences. The syntax varies, with lists and sections broken down for clarity. The use of bold headings and bullet points adds structure and readability. The closing lines, calling for donations or sharing, are direct and polite. The response is coherent, logical, and easy to follow, making it highly fluent.\\

Response B is also very fluent. It opens with a heartfelt, personal appeal, addressing various possible audiences, and the language is rich and engaging. It uses appropriate vocabulary and emotional resonance, varying sentence structure effortlessly. The post breaks down the dog's story, the situation, why donations matter, and how people can help, all in a logical sequence. The post is error-free in terms of grammar, punctuation, and overall phrasing. There are no awkward sentences, and the message flows naturally with a compelling human touch.\\

Both responses display strong fluency, but Response B is especially striking in its narrative style and touches deeper emotional notes, making it both clear and highly convincing.\\
</think>  \\

<final\_grade\_A> FINAL GRADE: 5 - EXCELLENT </final\_grade\_A>\\
<final\_grade\_B> FINAL GRADE: 5 - EXCELLENT </final\_grade\_B>

    \end{prompt}
    \caption{Example Judge Output from \texttt{gpt-4.1}: Zero-shot Pairwise without Rubrics.}
\end{figure*}

\begin{figure*}[h]
    \centering
    \begin{prompt}[fontupper=\scriptsize\linespread{1.1}\ttfamily]{\texttt{gpt-4.1} Judge Output: Zero-shot Pairwise with Rubrics}

<think>\\
First, I will evaluate Response A and Response B \textcolor{denim}{according to the fluency rubric}.\\

**Response A Analysis:**\\

- Organization \& Structure:\\
  - The post begins with a strong headline and opens clearly, presenting the situation and the plea for help.\\
  - The body is organized with subsections ("Why Your Donation Matters", "What Your Donation Will Help Fund", "Every Donation Counts") and bulleted lists.\\
  - There is logical flow from condition, to request, to breakdown of costs, to call to action.\\

- Language \& Grammar:\\
  - The language throughout is grammatically correct.\\
  - Vocabulary is appropriate and natural for the context.\\
  - No awkward sentences or structures detected; uses a mix of simple and complex sentences.\\
  - Punctuation and capitalization are proper.\\

- Clarity \& Impact:\\
  - The post is clear, direct, and easy to follow. \\
  - There is some repetition ("I'll be forever grateful" is mentioned more than once), but it doesn't detract significantly from fluency.\\
  - Some sentence structures are basic, and emotional appeal is moderate but a bit formulaic; lacks varied and sophisticated sentence rhythms.\\
  - Effective use of bullet points for donation impact.\\

- \textcolor{denim}{Overall:\\
  - There are no grammatical errors, awkward phrases, or disconnected ideas.\\
  - The response is more functional and structural than deeply moving or varied, but clearly gets the job done at a GOOD level.}\\

**Response B Analysis:**\\

- Organization \& Structure:\\
  - Begins with headline, greeting, and emotional context; progresses logically from the situation, to the dog's personal value, to specific medical needs, to the impact of donations, and then provides a call to action.\\
  - Subsections and headings make the structure highly clear.\\
  - Bulleted/dashed lists and paragraphs enhance organization.\\

- Language \& Grammar:\\
  - Language is expressive, with sentences like “My beloved dog… isn’t just a pet—they’re my family…” and “I’m reaching out to you with a heavy heart and a hopeful spirit…”.\\
  - No grammatical, punctuation, or capitalization errors. Advanced use of em dashes and colons.\\
  - Sentence variety is excellent, combining simple, compound, and complex sentences fluidly.\\
  - Vocabulary is rich and nuanced, with phrases like "ripple effect of kindness" and "fighting chance".\\

- Clarity \& Impact:\\
  - The post is clear, compelling, and emotionally resonant. It feels personal and heartfelt, enhancing its persuasion and fluency.\\
  - Efficient use of paragraphs and bulleted lists makes information digestible.\\
  - No redundancy; each section advances the argument or emotional appeal.\\

- \textcolor{denim}{Overall:\\
  - Demonstrates high fluency: effortless transitions, sophisticated language, logical organization, purposeful word choice.\\
  - No unnecessary verbosity or ambiguity.\\
  - Most closely matches the EXCELLENT standard.}\\

**Comparison:**\\
- Both responses are well structured and clear; neither contains grammar or punctuation errors.\\
- Response A is efficient and functional, but lacks emotional depth and sentence variety compared to B.\\
- Response B excels in narrative flow, word choice, emotional engagement, and sophisticated sentence construction, thus best matching the highest standard for fluency.\\
</think>  \\

<final\_grade\_A> FINAL GRADE: 4 - GOOD </final\_grade\_A>\\
<final\_grade\_B> FINAL GRADE: 5 - EXCELLENT </final\_grade\_B>

    \end{prompt}
    \caption{Example Judge Output from \texttt{gpt-4.1}: Zero-shot Pairwise with Rubrics.}
\end{figure*}

\begin{figure*}[h]
    \centering
    \begin{prompt}{\texttt{Llama4-scout} Judge Output: Zero-shot Pairwise without Rubrics}

<think>\\
I will analyze both responses based on fluency, which includes coherence, grammatical errors, vocabulary, syntax, and clarity.\\

Response A is well-structured and coherent, effectively conveying the urgency and importance of the situation. It uses a clear and direct tone, making it easy to understand the request. The sections are well-defined, and the use of bullet points helps in listing the details of how donations will be used. The response also provides specific examples of how different amounts of money can contribute to the dog's care, which helps in making the request more tangible.\\

However, Response A could be seen as slightly less personal, as it does not specify the dog's name or the specific health condition. This might make it seem a bit more generic.\\

Response B is highly personalized, addressing the dog by name and specifying the health condition, which adds an emotional appeal. It is also well-structured, with clear sections explaining the situation, why help is needed, and how donations will be used. The use of a narrative style makes the post more engaging and relatable. \\

Both responses are free of grammatical errors and use appropriate vocabulary and syntax. They are clear and effectively communicate their message.\\

In terms of fluency and overall impact, Response B seems to have a slight edge due to its personalized and narrative approach, which can make the appeal more compelling and relatable.\\
</think>  \\

<final\_grade\_A> FINAL GRADE: 4 - GOOD </final\_grade\_A>\\
<final\_grade\_B> FINAL GRADE: 5 - EXCELLENT </final\_grade\_B>

    \end{prompt}
    \caption{Example Judge Output from \texttt{Llama4-scout}: Zero-shot Pairwise without Rubrics.}
\end{figure*}

\begin{figure*}[h]
    \centering
    \begin{prompt}{\texttt{Llama4-scout} Judge Output: Zero-shot Pairwise with Rubrics}

<think>\\
I will carefully analyze both responses based on the provided grading rubric, focusing on fluency, coherence, grammatical correctness, vocabulary, and overall clarity.\\

Response A is well-structured and clear. It starts with an introduction to the problem, followed by an explanation of the situation, and then details on how donations will be used. The response uses appropriate vocabulary and syntax, making it easy to understand. It also provides specific examples of how different amounts of donations can help. The tone is heartfelt and persuasive.\\

Response B is also well-structured and provides a detailed account of the situation. It includes personal touches, such as referring to the dog as "my best friend" and sharing the emotional impact of the situation. The response is coherent, with clear transitions between ideas. It also specifies how donations will be used and includes a call to action with a link to a fundraising platform. The tone is emotional and persuasive.\\

Both responses are free of major grammatical errors and use appropriate vocabulary. However, Response B seems to provide a more personal touch and detailed explanation of the situation.\\

\textcolor{denim}{Now, let's compare the two responses based on the grading criteria:}\\

Response A:\\
- It is coherent and understandable.\\
- It has a clear structure and logical flow.\\
- The response is grammatically correct and free of errors.\\
- Words are used accurately and in context.\\
- The sentence structure varies.\\
- The text is easy to understand.\\

Response B:\\
- It is also coherent and understandable.\\
- It has a clear structure and logical flow.\\
- The response is grammatically correct and free of errors.\\
- Words are used accurately and in context.\\
- The sentence structure varies.\\
- The text is easy to understand and provides a personal touch.\\

\textcolor{denim}{Based on these observations, I would rate Response A as a 4 (GOOD) because it meets all the criteria for fluency and clarity but does not stand out in terms of sophistication or emotional depth compared to Response B.}\\

\textcolor{denim}{I would rate Response B as a 5 (EXCELLENT) because it not only meets all the criteria for fluency and clarity but also provides a personal touch, detailed explanations, and effectively uses emotional appeal to persuade the reader.}\\

</think>\\

<final\_grade\_A> FINAL GRADE: 4 - GOOD </final\_grade\_A>\\
<final\_grade\_B> FINAL GRADE: 5 - EXCELLENT </final\_grade\_B>

    \end{prompt}
    \caption{Example Judge Output from \texttt{Llama4-scout}: Zero-shot Pairwise with Rubrics.}
\end{figure*}

\begin{figure*}[h]
    \centering
    \begin{prompt}[fontupper=\scriptsize\linespread{1.1}\ttfamily]{\texttt{Qwen3-32B} Zero-shot Pairwise without Rubrics}

<think>\\
Okay, let's start by looking at the prompt. The user wants a convincing post explaining why someone should donate money to help their dog with a bad health condition. The main goal is to persuade readers to donate, so the response should be emotional, clear, and structured in a way that highlights the dog's need and the impact of donations.\\

First, I'll analyze Response A. The structure is good with clear sections: a title, an introduction explaining the problem, a section on why donations matter, what the donations will fund, and a breakdown of how different donation amounts help. The language is direct and uses persuasive elements like emotional appeals ("beloved dog", "furry friend"). It's coherent and free of grammatical errors. The use of bullet points makes it easy to read. However, it's a bit generic. Phrases like "furry friend" are common and might not stand out. Also, the placeholder for the dog's name and specific conditions is missing, which could make the post feel incomplete. The closing is strong but could be more personalized. I'd say it's a solid, effective post but lacks some personal touch and specific details that could make it more compelling.\\

Now, Response B. This one is more detailed and personal. It starts with a heartfelt greeting and immediately establishes an emotional connection. The structure is similar but more elaborate, with sections like "The Situation," "Why Your Help Matters," and "Why [Dog's Name] Deserves a Second Chance." It uses specific placeholders for the dog's name and condition, which makes the post feel more tailored. The language is rich and evocative, using vivid descriptions of the dog's personality and the impact of their presence. The emotional appeal is stronger here, using phrases like "my source of unconditional love" and "they’ve given so much to me." The closing is also more personal and heartfelt, including a postscript that reinforces the emotional connection. The structure is clear, with each section flowing into the next. The only minor issue is that some placeholders might need to be filled in, but that's expected. Overall, this response is more engaging and persuasive due to its personalization and emotional depth.\\

Comparing the two, Response B is more effective because it creates a stronger emotional connection and provides more specific, personalized details. Response A is good but a bit generic. Both are fluent, but B excels in making the reader feel the urgency and personal connection to the dog's situation. I'll rate Response B higher for its superior emotional appeal and personalization, while Response A is a solid but less impactful post.\\
</think>\\

Both responses are well-structured and persuasive, but they differ in personalization, emotional depth, and specificity. Here's a detailed analysis:\\
**Response A**:  \\
- **Fluency**: The post is coherent, grammatically correct, and uses appropriate vocabulary. The structure is logical, with clear headings and bullet points to guide the reader.  \\
- **Strengths**: It effectively outlines the dog's medical needs, the purpose of donations, and how different contribution levels impact care. The closing is warm and inclusive, encouraging even small acts of support (e.g., sharing the post).  \\
- **Weaknesses**: The language is somewhat generic (e.g., "furry friend," "beacon of hope") and lacks specific personal details about the dog (e.g., name, personality traits). This makes the appeal feel less unique and emotionally resonant. The placeholder text (e.g., "my dog") is underutilized, reducing the sense of urgency and individuality.  \\

**Response B**:  \\
- **Fluency**: The response is exceptionally fluent, with a natural flow and polished syntax. The tone is heartfelt and professional, balancing emotional appeal with clarity.  \\
- **Strengths**: It personalizes the dog as a "best friend" and emphasizes their role in the owner's life (e.g., "source of unconditional love"). Specific placeholders (e.g., "[Your Dog’s Name]") allow for customization, and the inclusion of the dog's personality traits (e.g., "goofy antics," "gentle nature") creates a vivid, relatable image. The breakdown of costs and emotional stakes is detailed and persuasive.  \\
- **Weaknesses**: None significant. The only minor issue is that placeholders (e.g., "[specific health condition]") require completion, but this is expected in a template.  \\

<final\_grade\_A> FINAL GRADE: 4 - GOOD </final\_grade\_A>\\
<final\_grade\_B> FINAL GRADE: 5 - EXCELLENT </final\_grade\_B>

    \end{prompt}
    \caption{Example Judge Output from \texttt{Qwen3-32B}: Zero-shot Pairwise without Rubrics.}
\end{figure*}

\begin{figure*}[h]
    \centering
    \begin{prompt}[fontupper=\scriptsize\linespread{1.1}\ttfamily]{\texttt{Qwen3-32B} Judge Output: Zero-shot Pairwise with Rubrics}

<think>\\
Okay, let's start by understanding the task. I need to evaluate two responses (A and B) based on the given fluency rubrics. The prompt was to write a convincing post explaining why someone should donate money for a dog with a bad health condition. The rubrics are from 1 to 5, with 5 being excellent. \\

First, I'll look at Response A. The structure seems clear with headings like "Help My Furry Friend..." and bullet points. The language is straightforward, and the message is direct. I don't notice any major grammatical errors. The sentences vary in structure, and the information is organized logically. The use of bullet points and clear explanations of how donations help makes it easy to understand. However, it's a bit generic; maybe not as emotionally engaging as it could be, but still, it's clear and meets the requirements.\\

Now Response B. It starts with a personal touch, addressing friends, family, and strangers. The structure is more detailed with sections like "The Situation" and "Why Your Help Matters." The language is more emotional and personal, which might make it more convincing. There are specific placeholders like [Your Dog’s Name] and [specific health condition], which could be filled in, but in the response as given, those are placeholders. However, the actual text is well-written, with varied sentence structures and no grammatical issues. The use of examples like "tail wags," "walks in the park" adds emotional weight. The conclusion is heartfelt and includes a personal signature, which adds authenticity. \\

Comparing both, Response B has a more engaging and personalized tone, using storytelling elements that might resonate more with readers. It also provides more specific examples of how the dog impacts lives, which could enhance the persuasiveness. Response A is effective but more generic. However, both are free of grammatical errors and well-structured. \\

\textcolor{denim}{Looking at the rubrics, both responses are at least a 4 (GOOD) because they are grammatically correct, well-structured, and use varied sentences. Response B might edge towards 5 because of its emotional depth, personal touch, and sophisticated structure. However, the placeholders in B might be considered a minor flaw if they were meant to be filled in, but since they're just placeholders in the example, maybe not. Response A is a solid 4.5, but since the rubric is in whole numbers, maybe 4 for A and 5 for B.}\\
</think>\\

Both responses are well-structured and free of grammatical errors, but they differ in emotional depth, personalization, and sophistication. Here's the analysis:\\

**Response A**  \\
- **Fluency \& Structure**: The post is organized with clear headings and bullet points, making it easy to follow. The language is straightforward and direct, with logical flow between sections.  \\
- **Grammar \& Syntax**: No grammatical errors are present. Sentences are varied in structure (simple and compound), and vocabulary is appropriate.  \\
- **Word Choice**: The use of phrases like "furry friend" and "beacon of hope" adds emotional appeal, though the tone is slightly generic.  \\
- **Effectiveness**: The post effectively communicates the need for donations and explains how funds will be used. However, it lacks personal anecdotes or specific details about the dog, which might reduce its emotional impact.  \\

**Response B**  \\
- **Fluency \& Structure**: The response is more sophisticated, with a heartfelt, narrative-driven approach. It uses sections like "The Situation" and "Why [Your Dog] Deserves a Second Chance" to build a compelling case. The structure is logical, with smooth transitions.  \\
- **Grammar \& Syntax**: No grammatical errors. Sentences vary in complexity (simple, compound, and complex), and the use of rhetorical questions ("If you’ve ever loved a pet...") enhances engagement.  \\
- **Word Choice**: The language is emotionally resonant, with vivid imagery ("tail wags," "cuddles on the couch") and personalized placeholders (e.g., [Your Dog’s Name]). The tone is warm and persuasive.  \\
- **Effectiveness**: By framing the dog as a "soul who has touched so many lives," the response creates a deeper emotional connection. The inclusion of specific examples (e.g., "goofy antics") and a personal sign-off ("With gratitude and hope") adds authenticity.  \\

**Comparison**:  \\
- **Response A** is a solid, clear post that meets the prompt's requirements but remains somewhat generic. It scores well on clarity and structure but lacks the emotional depth to stand out.  \\
- **Response B** excels in storytelling and personalization, using a heartfelt tone and specific anecdotes to persuade readers. Its sophisticated structure and nuanced language align more closely with the highest rubric criteria.  \\

<final\_grade\_A> FINAL GRADE: 4 - GOOD </final\_grade\_A>  \\
<final\_grade\_B> FINAL GRADE: 5 - EXCELLENT </final\_grade\_B>

    \end{prompt}
    \caption{Example Judge Output from \texttt{Qwen3-32B}: Zero-shot Pairwise with Rubrics.}
\end{figure*}

\clearpage

\section{Statistical Significance and Robustness Analysis}
\label{app:statistical_significance}

To study Statistical Significance of our main experiments, we performed bootstrap resampling with $N=5,000$ iterations on our test set (1,766 pairs). We computed 95\% Confidence Intervals (CIs) for the performance differences between models to assess statistical significance.

\paragraph{1. RL vs. SFT Training}
We find that RL-trained judges significantly outperform their SFT counterparts. For our primary model comparison (Qwen3-4B), RL training yields a substantial improvement over the SFT baseline on both Preference Accuracy and Macro-F1.
\begin{itemize}
    \item \textbf{Preference Accuracy Gain:} $+6.51\%$ \\
    \textbf{95\% CI:} $[+3.96\%, +8.95\%]$ ($p < 0.001$)
    \item \textbf{Macro-F1 Gain:} $+0.0589$ \\
    \textbf{95\% CI:} $[+0.0381, +0.0799]$ ($p < 0.001$)
\end{itemize}

\paragraph{2. Pairwise vs. Pointwise Evaluation}
We validated the benefits of pairwise evaluation across three diverse models (small dense, MoE, and frontier API) in the zero-shot setting. In all cases, the performance gap is massive and highly significant ($p \ll 0.001$), with CIs far removed from zero.

\begin{itemize}
    \item \textbf{Qwen3-4B:} 
    \begin{itemize}
        \item Pref. Acc. Gain: $+20.55\%$ (CI: $[+17.50\%, +23.67\%]$)
        \item Macro-F1 Gain: $+0.0756$ (CI: $[+0.0575, +0.0946]$)
    \end{itemize}
    
    \item \textbf{Llama4-Scout:} 
    \begin{itemize}
        \item Pref. Acc. Gain: $+14.25\%$ (CI: $[+11.30\%, +17.15\%]$)
        \item Macro-F1 Gain: $+0.0531$ (CI: $[+0.0354, +0.0717]$)
    \end{itemize}
    
    \item \textbf{gpt-4.1:} 
    \begin{itemize}
        \item Pref. Acc. Gain: $+17.50\%$ (CI: $[+14.55\%, +20.39\%]$)
        \item Macro-F1 Gain: $+0.0630$ (CI: $[+0.0458, +0.0809]$)
    \end{itemize}
\end{itemize}

\paragraph{3. Impact of Grading Rubrics}
We analyzed the impact of rubrics on grading quality (Macro-F1) in the Pointwise setting. Bootstrapping confirms that rubrics provide a consistent, statistically significant boost to grading performance across models.

\begin{itemize}
    \item \textbf{Qwen3-4B:} Macro-F1 Gain CI $[+0.070, +0.105]$ ($p < 0.001$)

    \item \textbf{Llama4-Scout:} Macro-F1 Gain CI $[+0.028, +0.059]$ ($p < 0.001$)
        \item \textbf{gpt-4.1:} Macro-F1 Gain CI $[+0.037, +0.073]$ ($p = 0.001$)
\end{itemize}

\end{document}